\documentclass{article}

\usepackage{arxiv}

\usepackage[utf8]{inputenc} % allow utf-8 input
\usepackage[T1]{fontenc}    % use 8-bit T1 fonts
\usepackage{hyperref}       % hyperlinks
\usepackage{url}            % simple URL typesetting
\usepackage{booktabs}       % professional-quality tables
\usepackage{amsfonts}       % blackboard math symbols
\usepackage{nicefrac}       % compact symbols for 1/2, etc.
\usepackage{microtype}      % microtypography
\usepackage{lipsum}		% Can be removed after putting your text content
\usepackage{graphicx}
\usepackage{natbib}
\usepackage{doi}
\usepackage[linesnumbered,ruled,vlined]{algorithm2e}
\usepackage{geometry}                		% See geometry.pdf to learn the layout options. There are lots.
\usepackage{subfigure}
\usepackage{amssymb}
\usepackage{tabularx}
\usepackage{amsmath}
\usepackage[english]{babel}
\usepackage{fancyhdr}
\usepackage{multirow}

\newcommand{\Neib}{\mathcal{N}}

\newcommand{\R}{\mathbb{R}}

\newcommand{\ep}{\epsilon}

\newcommand{\norm}[1]{\|#1\|}

\title{the effect of training parameters and mechanisms on decentralized federated learning based on MNIST dataset}
\author{{\hspace{1mm}Zhuofan Zhang} \\
	School of Electrical and Computer Engineering\\
	Georgia Institute of Technology\\
	Atlanta, GA 30332 \\
	\texttt{zzhang433@gatech.edu} \\
	\And
	{\hspace{1mm}Mi Zhou} \\
	School of Electrical and Computer Engineering\\
	Georgia Institute of Technology\\
	Atlanta, GA 30332  \\
	\texttt{mizhou2018@gatech.edu} \\
	\And
{\hspace{1mm}Kaicheng Niu} \\
	School of Electrical and Computer Engineering\\
	Georgia Institute of Technology\\
	Atlanta, GA 30332 \\
	\texttt{kniu9@gatech.edu} \\
	\And
	{\hspace{1mm}Chaouki Abdallah} \\
	School of Electrical and Computer Engineering\\
	Georgia Institute of Technology\\
	Atlanta, GA 30332  \\
	\texttt{ctabdallah@gatech.edu} \\	
}
\renewcommand{\shorttitle}{\textit{arXiv} Template}

\begin{document}
\maketitle
\begin{abstract}
Federated Learning is an algorithm suited for training models on decentralized data, but the requirement of a central "server" node is a bottleneck.
In this document, we first introduce the notion of Decentralized Federated Learning (DFL).
We then perform various experiments on different  setups, such as changing model aggregation frequency, switching from independent and identically distributed (IID) dataset partitioning to non-IID partitioning with partial global sharing, using different optimization methods across clients, and breaking models into segments with partial sharing. 
All experiments are run on the MNIST handwritten digits dataset. 
We observe that those altered training procedures are generally robust, albeit non-optimal. 
We also observe failures in training when the variance between model weights is too large. 
The open-source experiment code is accessible through GitHub\footnote{Code was uploaded at \url{https://github.com/zhzhang2018/DecentralizedFL}}.
\end{abstract}

\keywords{Federated Learning \and Distributed Learning \and Segmented Federated Learning \and small-world phenomenon}

\section{Introduction}
Federated learning is an approach to learn on decentralized data, first proposed by \cite{FLMcM}. 
In aims to provide privacy and safety benefits, federated learning has a distinct advantage over traditional data center-based training algorithm, thus has a huge potential in the Internet of Things (IoT) networks.
The idea is to have a central server send a model to a batch of randomly selected devices (i.e., clients) with local data, then have each client perform a local update independently, and finally have the server aggregate all the updated client-side model parameters, and take their average as the model parameters in the next iteration. 
This procedure is named the Federated Averaging (\texttt{FedAvg}) algorithm. 

Federated learning has been a rapidly growing research field ever since its introduction. 
The main challenges of federated learning are heterogeneous data distribution in real applications, the difficulty of collecting data, and theoretic analysis of convergence.
Many recent research works have been adjusting the \texttt{FedAvg} algorithm and its variants for different applications and for reducing overall communication costs~\citep{FLMcM,prox,wang2020federated,scaffold}.
Many parts of the \texttt{FedAvg} algorithm could be adjusted for different situations, such as testing different aggregation methods (other than averaging), reduction of communication cost during aggregation, application considerations in real working environments, and so on.

Above federated learning algorithm requires a centralized model to maintain a global model over the network after aggregating information from other nodes.
Considering of this drawback, distributed federated learning is proposed by some works~\citep{lalitha2019peer,savazzi2020federated,taya2021decentralized}.
\cite{lalitha2019peer} proposes a peer-to-peer distributed federated learning on graphs by performing a Bayesian update on the private belief vector thus form the public belief vector.  
\cite{savazzi2020federated} proposes a novel class of distributed federated learning based on the iteratively exchange of both model updates and gradients. Moreover, large scale massive networks are used to validate the algorithm.
Recently, \cite{taya2021decentralized} presents a decentralized federated learning scheme called CMFD, the core idea of which is to aggregate the local prediction functions instead of parameters.
These works show that switching from a centralized sharing scheme in \texttt{FedAvg} to a decentralized one can allow models to reach a consensus at the global minimum under some conditions, while avoiding communication bottleneck due to the server node.

This paper is organized as follows: Section \ref{sec:background} is the background of our problem. 
Section \ref{sec:baseline} presents the baseline we use for compare.
Section \ref{sec:datadis} shows the effect of different dataset distribution.
Section \ref{sec:optimethods} presents the effect of different optimization methods.
Section \ref{sec:segfl} is the analysis and simulation results of segmented federated learning with combination of small-world communication topology.
Finally, Section \ref{sec:conclusion} is the conclusion of our work and some future works.

The main contributions are stated as follows:
\begin{enumerate}
    \item The effects of model aggregation frequency without initial weights is explored.
    \item The effects of different dataset distribution, such as different dataset sizes, partially overlapping training data, skewed training data and differently normalized training data are presented.
    \item The effects of different optimization methods are also tested on MNIST dataset.
    \item Segmented Federated learning and small-world topology is combined and implemented.
    \item Several unsolved problems and future works are proposed.
\end{enumerate}
%%%%%%%%%%%%%%%%%%%%%%%%%%%%%%%%%%%%%
\section{Background} \label{sec:background}
In this section, we will first introduce the \texttt{FedAvg} algorithm, and the decentralized federated learning (DFL) adaptation. 
Then, we describe the neural network structure and computation tools used throughout the paper.

\subsection{Decentralized Federated Learning} 
We attempt to solve the MNIST classification problem \citep{lecun1998gradient} with a general decentralized federated learning scheme, outlined in this section. 
For the following, we assume that $G$ is an undirected graph that describes client connectivity, and maintains strong connectivity over time. $\Neib_i(t)$ represents the nodes connected to client $i$ at time $t$. 
We use $K$ to represent the total number of clients, and $w^i(t)$ to represent the $i$-th client's parameters at time $t$. 
The pseudocode is as follows: 

\begin{algorithm}[H]
\SetAlgoLined
 Initialize all clients with the same parameters $w(0)$\;
 \For{$t = 1,2,\cdots,T$}{
  \For{$i = 1,\cdots,K $ \text{in parallel}}{
   \For{$j = 1,\cdots, |D_i|E/B$}{
    $w^i(t) \leftarrow w^i(t) - \eta \nabla l(w^i(t); b_j)$\;
   }
   \For{$k = 1,\cdots, s_\ep(t)$ in parallel}{
    \For{$i = 1,\cdots,K$}{
     $w^i(t) \leftarrow w^i(t) + \sum_{j \in \Neib_i} \alpha_{ij} (w^j(t)-w^i(t))$ \;
     $w(t+1) \leftarrow w(t)$\;
    }
   }
  }
 }
\end{algorithm}

where $\eta$, $\alpha_{ij}$ are constants that determine update rates, $D_i$ is the local dataset for client $i$, $B$ is the batch size for each client's local updates, $b_j$ is the samples drawn in the $j$-th batch, and $l$ is the loss function. 
The value of $\alpha_{ij}$ can be different according to $i$ and $j$ in the implementation. 

Notice that step 9 follows the consensus protocol, but might not be really necessary. 
If each client is able to receive the full weight parameters from all neighbors in this step, then we might as well just get to the consensus position as fast as possible instead: \begin{align}
     w^i(t) \leftarrow \frac{1}{|\Neib_i|}\sum_{j\in\Neib_i\text{ or }j=i} w^j(t).
\end{align}

This algorithm is not a generalization to the centralized \texttt{FedAvg} algorithm. 
The \texttt{FedAvg} algorithm effectively uses a star graph $G_S$ with $K+1$ nodes, where the server node is the center. 
Lines 1-5 would still apply because the server node does not have its own local data. 
Lines 6-8 would be slightly different, in that the server does not put its own model in the averaging, and no client participates in the averaging, either. The \texttt{FedAvg} update rules would be: 
\begin{align*}
    & 6.  w^S(t) \leftarrow \frac{1}{K*C}\sum_{j \in C(t)} w^j(t)\\
    & 7. \text{Select a new set of clients }C(t+1)\\
    & 8.  \text{for } i \in C(t+1) \text{ do}:\\
    & 9. ~ ~ ~ ~ w^i(t+1) \leftarrow w^S(t).
\end{align*} 
Organizing the training procedure is much easier in this case. 
In addition, in \texttt{FedAvg}, the server only selects a subset of all the clients for each round. 
The aggregation algorithm always converges if the graph is connected and the weights are positive~\cite{mesbahi2010graph}.

Observe that one key property of \texttt{FedAvg} is that all the clients start from the same model before starting each round of local training. 
The trained model could risk diverging if the starting weights are too far away from each other. %without this guarantee, but we'll see later (where?) that it depends. 
This serves as an important thing to preserve when designing the parameters for the decentralized FL. 

\subsection{Methods} \label{methods}
The network setup is based on the official PyTorch example~\citep{pytorchMNIST}.
The network consists of a 32-channel convolution layer, followed by another 64-channel convolution layer, both with a kernel size of 3 and followed by an activation of ReLU. 
After a $2\times2$ MaxPooling layer and a Dropout layer with probability 0.25 is a fully connected layer with 128 output values and ReLU activation. 
This is followed by another Dropout layer with probability 0.5, and then a second fully connected layer with 10 outputs. 
The outputs are turned into predictions for 10 classes through Softmax, and evaluated with negative log likelihood function against the correct label.

While the preliminary experiments in the following section are run on a local computer in Jupyter Notebook with a Python 3.8 kernel, the experiments in section 3 are run on \cite{PACE}.
%%%%%%%%%%%%%%%%%%%%%%%%%%%%%%%%%
\section{Experiments to Establish Baselines}\label{sec:baseline}
In this section, we first introduce parameters that need to be determined. Next, we present preliminary experiment results to show viability of DFL, and determine ranges of hyper-parameters used in the rest of the article. 
\subsection{Hyper-parameters in Decentralized Federated Learning}
There are several values in the decentralized \texttt{FedAvg} process that are important to the training performance: \begin{itemize}
    \item $s$, \textbf{the number of iterations to execute the consensus protocol, so that clients reach the same model weights before starting another round of local update}:
    If $s$ is too large, then the averaging step would incur too much communication cost when models are already sufficiently close.
    If $s$ is too small, then the client network would not be able to reach a consensus before making another round of update, and the model may diverge in the worst case. 
    To make this parameter fall in an appropriate region, we set it as $s_\ep(t)$ where $s_\ep$ is defined as the least amount of iterations it would need for the entire network's client models to reach consensus with error at most $\ep$ between any pair's parameters. 
    We place $\ep$ at 0.1, and let $s = \min\{100, s_{\ep}\}$ for this section. These values proved to be reasonable.
    \item $E$, \textbf{the amount of model updates performed throughout one epoch, between the model aggregation steps}: 
    This value also represents the inverse of model aggregation frequency.
    For example, if $E=3$, then each client goes through its local dataset for 3 times before sharing their model parameters, and aggregation frequency is $1/3$ times per one pass of dataset.
    If $E = 0.5$, then the client only goes through half of the local training dataset before aggregation. 
    \item $\lambda$, \textbf{communication network density }: The client communication can be described by a graph $G$ where each client is a node, and each pair of communicating clients forms an edge,
    The network density can be described by the second eigenvalue of $G$'s Laplacian. A denser network generally has more connections and a higher magnitude of $\lambda$, and thus would have faster convergence rate $\propto e^{|\lambda| t}$ for the consensus protocol, $\lambda > 0$. 
    This is covered in detail in subsection \ref{smallworld}. 
    Conversely, a less dense network has a slower convergence rate, and this rate is minimized when the network is a tree or a cycle. 
    In this section, we mainly use a \textbf{cyclic topology} to simulate the behavior of model aggregation in a sparser network with slower convergence, as a semi-worse-case scenario.
\end{itemize} 
In addition, the model construction and the training environment could also affect the performance. 
In this section, we fix most of those parameters mentioned above and in subsection \ref{methods} other than $E$.
The goal is to find a good combination of those values, to serve as a baseline reference for experiments in other sections.

\subsection{Preliminary Testing with Centralized Federated Learning}
We first examine the performance of the original \texttt{FedAvg} algorithm. 
We run the centralized \texttt{FedAvg} algorithm from \cite{FLMcM}, and follow the parameters in the original paper, where $K=20$ clients and a central server train over the entire dataset. 
The dataset $D$ is randomly split into IID and mutually disjoint subsets $D_1, ..., D_K$, and distributed across all clients.
In addition, each epoch only updates half ($C = 0.5$) of all the clients, meaning that $CK=10$ clients participate each round of local model updates and parameter aggregation. 
To take care of the limited computational resources, and to reveal more interesting training behaviors across epochs, we reduce $E$ from $>10$ in the original paper to the range between 0.01 and 1. 
Batch size was also chosen to be $B=10$. 

As a comparison, we also train a single model with the entire MNIST training dataset ($|D| \triangleq 60000$ samples) as the baseline. 
The model reached 98.28\% accuracy on the test set (6000 samples) after the first epoch.

The results show that the parameter choices for \texttt{FedAvg} could produce decent models. 
Server side model reaches 98\% accuracy after around 20 epochs. 
Considering that there are $K=20$ clients, $CK|D_1|E = CK|D|E/K = 0.005|D|$ samples are used to update in each round, and $10\%$ of all samples are used in this experiment to achieve 98\% accuracy. 
After 100 rounds, the number of samples processed roughly equal to the size of the entire dataset, and the resulting test accuracy could compare with the 98.28\% baseline accuracy.

\subsection{Preliminary Testing with Decentralized Federated Learning}
Next, we implement decentralized federated learning (DFL), where we choose $K=10$ as a first step, so that the number of clients sharing the parameters each round would be the same as the $CK$ value in the experiment above. 
DFL requires an aggregation step instead of direct averaging, so we pick $s = \min(s_{\ep},100)$ with error threshold $\ep = 0.1$. Different from the methods in \cite{FLMcM}, we switch to $B=64$ for faster training. 

The dataset is partitioned evenly into $K$ local sections for the clients, in an independent and identical distribution manner. 
The accuracy history on the test dataset after each training epoch $t$ is obtained as the prediction accuracy of the client with the lowest training loss $l(w^i(t))$.
All models start at the same initial weights.
The behaviors are displayed in Table \ref{tab:dflE1}.

\begin{table}[!htbp]
\caption{Behavior of DFL with varying $E$}
	\centering
	\begin{tabular}{ccc}
		\toprule
		$E$ & Max test accuracy & Number of samples seen at max accuracy per client 
		\\ \midrule
		1 & 98.8\% at the 38th epoch & 3.8$|D|$ 
		\\ \midrule
		0.05 & 97.85\% at the 490th epoch & 2.45$|D|$ 
		\\ \midrule
		0.005 & < 90\%, maxed at the 5th epoch & 0.025$|D|$ 
		\\ \bottomrule
	\end{tabular}
	\label{tab:dflE1}
\end{table}
It appears that $E=1$ reaches a good model quickly, while $E=0.05$ would yield suboptimal performance. We also observe that training loss for $E=0.005$ starts increasing after the 5th epoch, indicating model overfit. These preliminary experiments confirm that $E=1$ and $E=0.05$ are both reasonable choices in the DFL setup.

However, if we want to study the model behavior during training, we cannot let the models converge too quickly. $E=1$ appears to be a good choice for training better-performing models, but $E=0.05$ offers the opportunity to look closer into the model performance history over several hundreds of training epochs. 
\subsection{Effect of Model Aggregation Frequency without Uniform Initial Weights} 
In this section, we further observe the effect of $E$ on model performance, but in an unconventional setup. 
In the standard federated learning setup, each client is supposed to have disjoint training datasets, for privacy concerns. 
Each training data should only be available to the client that hosts it, and the client is only supposed to share its model parameters without making the local sample somehow traceable by other clients. 
However, in an experimental setting, we could disregard this requirement, grant all clients access to the full or most part of the dataset, so that we can evaluate the robustness of the simple averaging aggregation step while excluding other factors. 

Under this setup, we have an interesting observation: When $E=1$, none of the client models is able to learn a good classification model when the initial model parameters happen to be completely random.
The models quickly diverge, the test accuracy quickly drops to just above 10\%, no better than random guesses. 
At the same time, the loss history for each client increases with $t$, due to the unbounded nature of negative log-likelihood losses. 
The results in Table \ref{tab:dflE2} shows the model behavior with different $E$ values under this setup. 

\begin{table}[!htbp]
\caption{Behavior of DFL with varying $E$}
	\centering
	\begin{tabular}{ccc}
		\toprule
		$E$ & Max test accuracy & Observations
		\\ \midrule
		1 & 10.28\% at the 2nd epoch & Training loss reached plateau at 2nd epoch
		\\ \midrule
		0.05 & 18\% at the 10th epoch & Training loss still decreases after a few epochs, but very slow 
		\\ \midrule
		0.005 & 86\%, at the 15th epoch & Training after 15th made no improvement; Overfit behavior at the 10th epoch
		\\ \bottomrule
	\end{tabular}
	\label{tab:dflE2}
\end{table}
Converse to the observation in the previous section, a higher update frequency (smaller $E$) performs better in this situation. 

The most possible explanation is that, when the aggregation became more frequent, the clients' models would become less different after each epoch, and the model parameters would reach consensus faster. When $E=0.005$, the models reach consensus at a suboptimal point first, and then could not advance forward. When $E=0.05$ or $E=1$, the models are trained for too long before aggregation, and the resulting centroid point is an even more suboptimal local minima. 

The models are supposed to train better with more data available, especially with access to the entire dataset in this section's setup. 
Thus, we conclude that the federated learning method could be fragile without uniform weights across clients. 
In the following sections, we shall make sure all models start with a uniform set of weights or a close set of weights, unless otherwise specified. 

\section{Effects of Different Dataset Distributions}\label{sec:datadis}
In the previous section, we have shown the effects of hyper-parameters on the federated learning when training on the MNIST dataset. 
In this section, as well as the following sections, we mostly settle down on a fixed combination of the hyper-parameters, and instead investigate the decentralized federated learning model behavior with other different adjustments.

In the following subsections, we present the performance of DFL when dataset partition across clients are not of the same sizes, not IID, or both. We also investigate the effect on performances with shared data between clients. 

\subsection{Different Dataset Sizes} 
To create clients with different dataset sizes, we draw samples from a Gaussian distribution $x_{1:K} \sim \mathcal{N}(0,\sigma^2)$, 
and then randomly and uniformly draw $N_i$ samples for client $i$ as its training data using the following equation: 
\begin{align}
    N_i = \min\biggl(\dfrac{|x_i||D|}{\sum_{j=1}^K|x_j|}, B\biggr).
\end{align} 
The resulting dataset distribution for each client is kept IID by the uniform sampling. 

To better describe the model performances, we 
define $e_{p}$ as the number of epochs passed before the overall accuracy first exceeds $p\%$. We call $p\%$ as the \textbf{accuracy threshold} with the corresponding \textbf{epoch number}, $e_p$. These values will be used as a quantifiable indication on model performance for the rest of the paper.
As a preliminary test, we use different combinations of $K$ and $\sigma$. The results are shown in Table \ref{tab:table00}. In all tests, the models are able to achieve 97\% within 100 epochs, so we could conclude that DFL is robust against different dataset sizes when training in MNIST.

\begin{table}
	\caption{Performance with different dataset sizes across clients}
	\centering
	\begin{tabular}{ccccc}
	\hline
	Episode number & $K=10,\sigma^2=1$ & $K=30,\sigma^2=2$ & $K=50,\sigma^2=2$ & $K=30,\sigma^2=10$ \\
		\midrule
$e_{90}$ & 6 & 6 & 10 & 8\\
\hline
$e_{97}$ & 80 & 36 & 52 & 51\\
		\bottomrule
	\end{tabular}
	\label{tab:table00}
\end{table}

Note that, when $\sigma^2 = 10$, the variance is high enough for up to 50\% of the clients 
to only have $B = 64$ data samples to start with. 
Most data are concentrated in a selected few clients. 
This unbalanced distribution facilitated the learning of these few with more training data, but their contributions are weakened during parameter sharing because they have the same weight as the ones with fewer training data. 
Thus, training is slower when $\sigma^2$ is higher, as expected. 
One way to overcome this is to use a weighted average sum, where the weight $w_i$ for the parameters of client $i$ is proportional to $N_i$, as proposed by \cite{hu2019decentralized}.

\subsection{Skewed Training Data} 
\cite{skewedFL} have already investigated the effect of skewed datasets on federated learning. 
In their paper, they partitioned the CIFAR-10 dataset, so that each client only has access to samples of the same label. 
For conveniency, we denote the dataset partition as: 
\begin{align}
    D = \bigcup_{m=1}^{10} D_m; \ \ \forall m\neq n, D_m\cap D_n = \varnothing,
\end{align}
where $D_m$ is the subset containing all data pairs with output labels as $m$. 
During partition, each client $i$ is assigned a dataset $D^{(i)} \subset D_{c_i}$, where integer $c_i \in [1,10]$ indicates a specific label. 
As demonstrated in \cite{skewedFL}, this assignment leads to poor performance on CIFAR-10, and could be alleviated by pre-allocating a shared global dataset containing uniform label distribution. 
Running the similar training procedure on MNIST give different results: The global model converges to an accuracy above 90\%, but this accuracy is relatively poorer compared to the models trained with less skewed datasets.

\cite{skewedFL} proposed a coping mechanism by sharing a small packet of unskewed training data globally. 
The clients train on this global training data on top of their private datasets. 
This method is able to significantly improve the performance. 
In this paper, we slightly modify this procedure by only enabling pairwise sharing, in order to see if it could bring the same level of improvements.

We introduce the parameter $S \in [0,1]$ to describe the amount of training data that one client shares with others. 
Again, we suppose each client $i$ is assigned a partition $D^{(i)}$ on startup.
Next, we perform the sharing step, where client $i$ would share $S|D^{(i)}|$ samples to all other clients.
The shared samples are drawn uniformly, divided uniformly, and not necessarily the same for different clients. 
As such, each client $j$ would receive $\dfrac{S}{K-1}|D^{(i)}|$ samples from $i$.
After this step, each client $i$ would have an expanded training dataset $D'^{(i)}$, where: 
\begin{align}
    |D'^{(i)}| = |D^{(i)}| + S\sum_{j=1,j\neq i}^K |D^{(j)}|
\end{align} 
Each client $i$ would have shared a total of $S|D^{(i)}|$ training samples to all other clients, and would have received $\dfrac{S|D^{(j)}|}{K-1}$ samples from each client $j$. When $S=0$, no client shares any data to each other. On the other extreme, when $S=1$, all the clients use the same set of globally shared training data.

In the following, we compare the results using different optimizers provided by PyTorch with empirically-determined learning rates on a completely skewed dataset, with each label distributed to exactly one of the $K=10$ clients. 
In addition, we investigate how sharing only a small percentage of the data samples can greatly alleviate the performance reduction. 
The learning rates (Table \ref{tab:lr}) are determined empirically to stabilize training and maximize performance, while keeping the training reasonably fast. 

\begin{table}[!htbp]
\caption{Empirical learning rates for each optimizer}
	\centering
	\begin{tabular}{cccccc}
		\toprule
Optimizer & Adam & Adadelta & Adagrad & RMSprop & SGD\\
		\midrule
Learning rate & 0.001 & 1 & 0.001 & 0.0005 & 0.01\\
		\bottomrule
	\end{tabular}
	\label{tab:lr}
\end{table}

\begin{figure}[!htbp]
    \centering
    \subfigure[$E=0.05$, same initial  weights]{\includegraphics[width=0.32\textwidth]{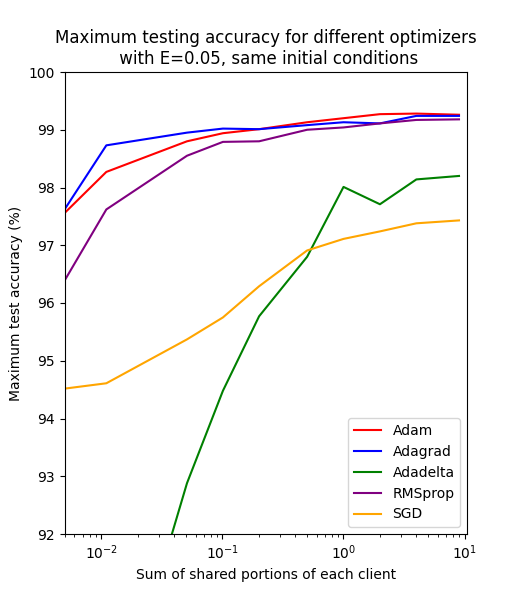}}
    \subfigure[$E=0.05$, different random initial model weights]{\includegraphics[width=0.32\textwidth]{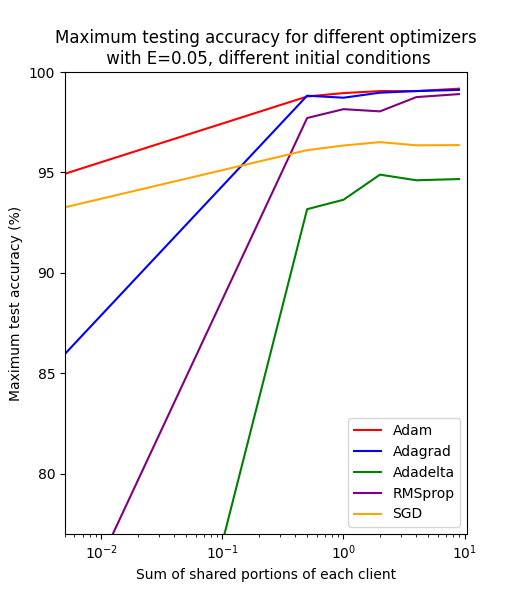}}
    \subfigure[$E=1$, same initial  weights]{\includegraphics[width=0.32\textwidth]{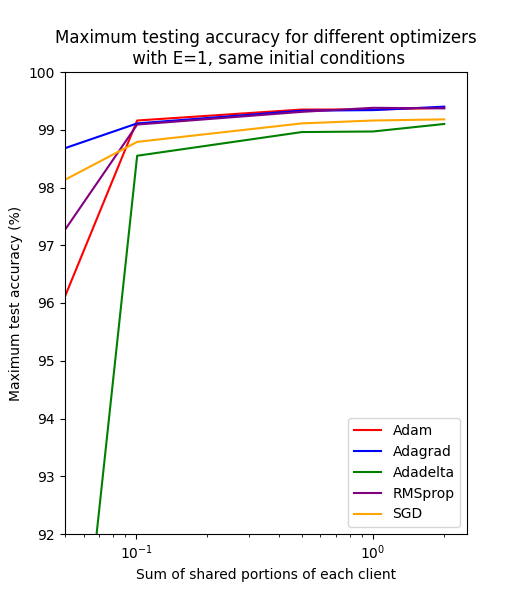}}
    \caption{Comparisons of maximum model test accuracies (\%) with different values of $S$ under different optimizers and training setups. Note that the horizontal axis is the sum of shared portions, meaning that a value of $2$ indicates each client has the parameter $S = \dfrac{2}{K} = 0.2$.}
    \label{fig:Skewed}
\end{figure}

Figure \ref{fig:Skewed}(a) shows the average model performances with $S$ ranging from 0 to 1 under different optimizers.
The accuracy values are obtained as the max average test accuracy across all clients within 500 epochs of training.
The figure indicates that performances are dependent on optimizers.
More importantly, the top performances does not differ by too much, as long as $S > 1\%$.
Note that the horizontal axis is the total shared portions before training started.
Because each client shared $S$ of its training data, the total shared portions would be $K\cdot S$.
Thus, $S=1\%$ corresponds to the $10^{-1}$ mark on the horizontal axis. 

In addition, we conduct further tests on models without same initial weights, shown in Figure \ref{fig:Skewed}(b).
Without any sharing, the results from the previous sections have shown that models tend to fail at training.
However, here it appears that the model training is stabilized with $S \geq 5\%$ for proper optimizers. 

To investigate whether the observed convergence is affected by the higher model aggregation frequency at $E=0.05$, we repeat the experiments using $E=1$, and with same initial model weights. 
The results are similar to having $E=0.05$, as shown in Figure \ref{fig:Skewed}(c). 
Note that these experiments are trained with only 50 epochs instead of 500, so that models at $E=1$ are not trained with significantly more samples than did models at $E=0.05$. 

At this point, one may question what exactly is the difference between the effects caused by sharing data samples and by using less skewed data. 
We now try to combine both ideas.
For simplicity, we again use $K=10$ clients, and assign $D^{(i)}$ as the original training set for client $i$ after partition.
To describe the "skewed-ness" of the training data for model $i$, we define: 
\begin{align}
U_i \triangleq \dfrac{|D_i \cap D^{(i)}|}{|D_i|}
\end{align}
This value is the percentage of the data with label $i$ in $D^{(i)}$.
For example, in Figure \ref{fig:Skewed}, the dataset $D^{(i)}$ consists entirely of data with label $i$ for each $i$, so $U_i = 1$.
This is the most unbalanced way of partitioning the training data.

In the following experiments, we assume $U_i = U$ for all $i$, and we apply the data sharing procedure prior to training. 
Each client starts with partitioned dataset $D^{(i)}$ with skewedness $U_i$. 
This means it contains $U|D_i|$ data samples with the label $i$, and $(1-U)|D_i|$ data samples with  other labels. We have $|D_i| = |D^{(i)}|$ for the dataset size.

Next, client $i$ receives $\dfrac{S|D^{(j)}|}{K-1}$
samples from each client $j$. 
The expected number of samples with label $i$ in each of those datasets is thus $\dfrac{S(1-U)|D^{(j)}|}{(K-1)^2}$.
After the sharing is complete, client $i$ would have an expanded dataset $D'^{(i)}$ with size $|D'^{(i)}| = (1+S)|D^{(i)}|$. 
The expected skewed-ness of this training set is thus: 
\begin{align}
 \dfrac{U|D_i| + \sum_{j=1, j\neq i}^K \dfrac{S(1-U)|D_j|}{(K-1)^2} }{|D'^{(i)}|} = \dfrac{U(K-1)+S(U-1)}{(1+S)(K-1)} + \dfrac{S(1-U)}{(1+S)(K-1)}\dfrac{|D|}{|D_i|}   
\end{align}

\begin{figure}[!htbp]
    \centering
    \subfigure[Adam]{\includegraphics[width=0.32\textwidth]{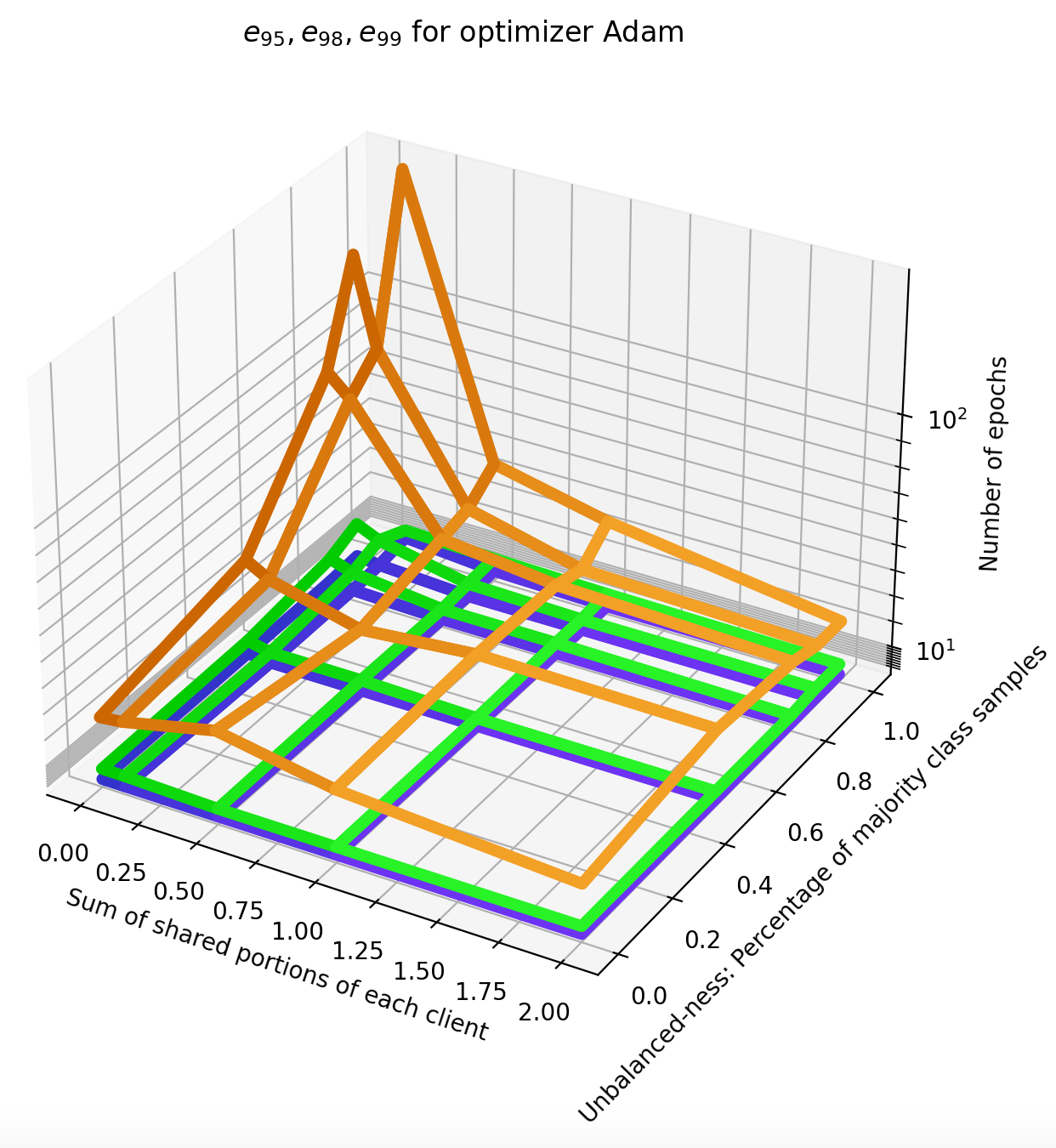}}
    \subfigure[Adadelta]{\includegraphics[width=0.32\textwidth]{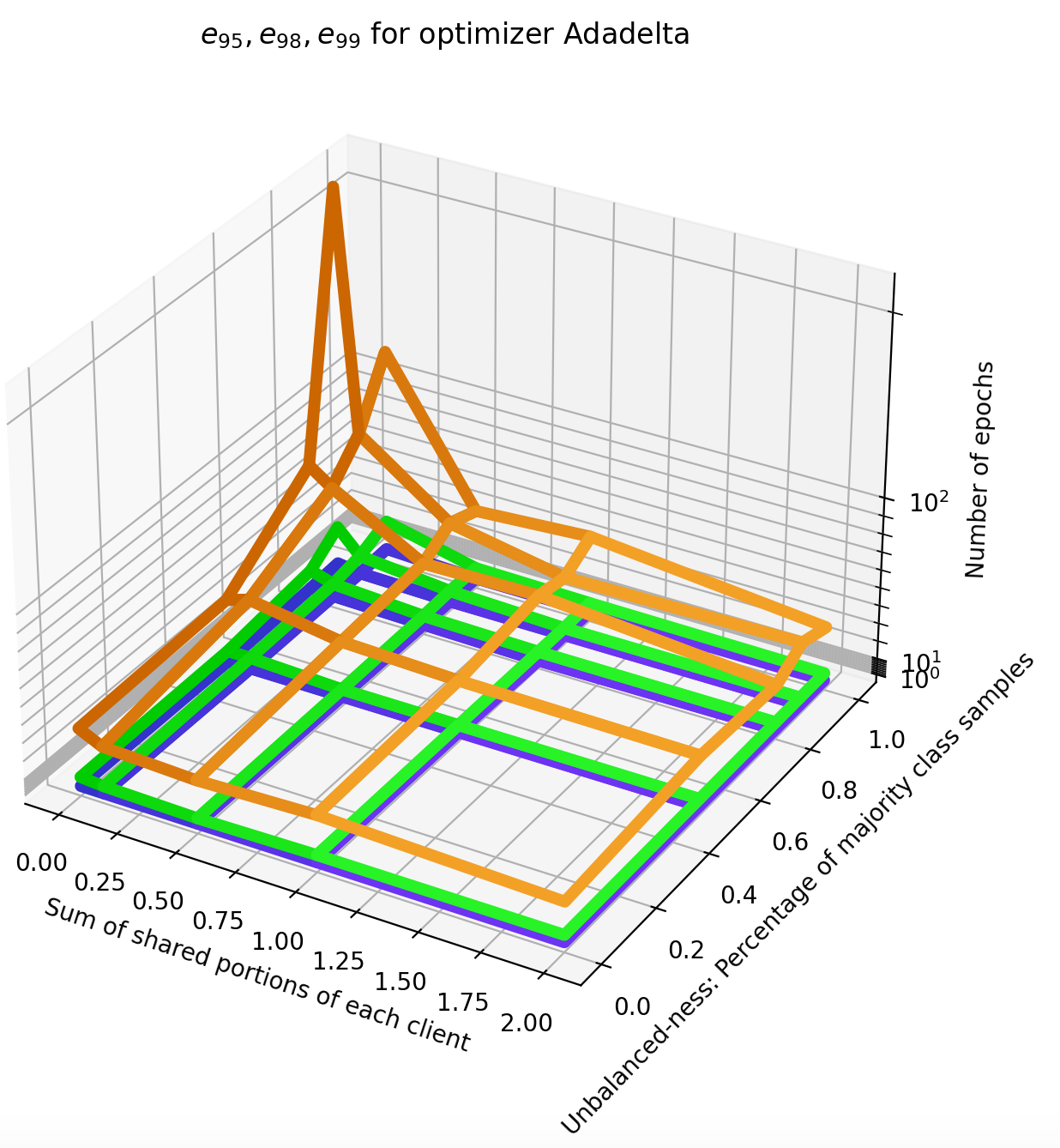}}
    \subfigure[RMSprop]{\includegraphics[width=0.32\textwidth]{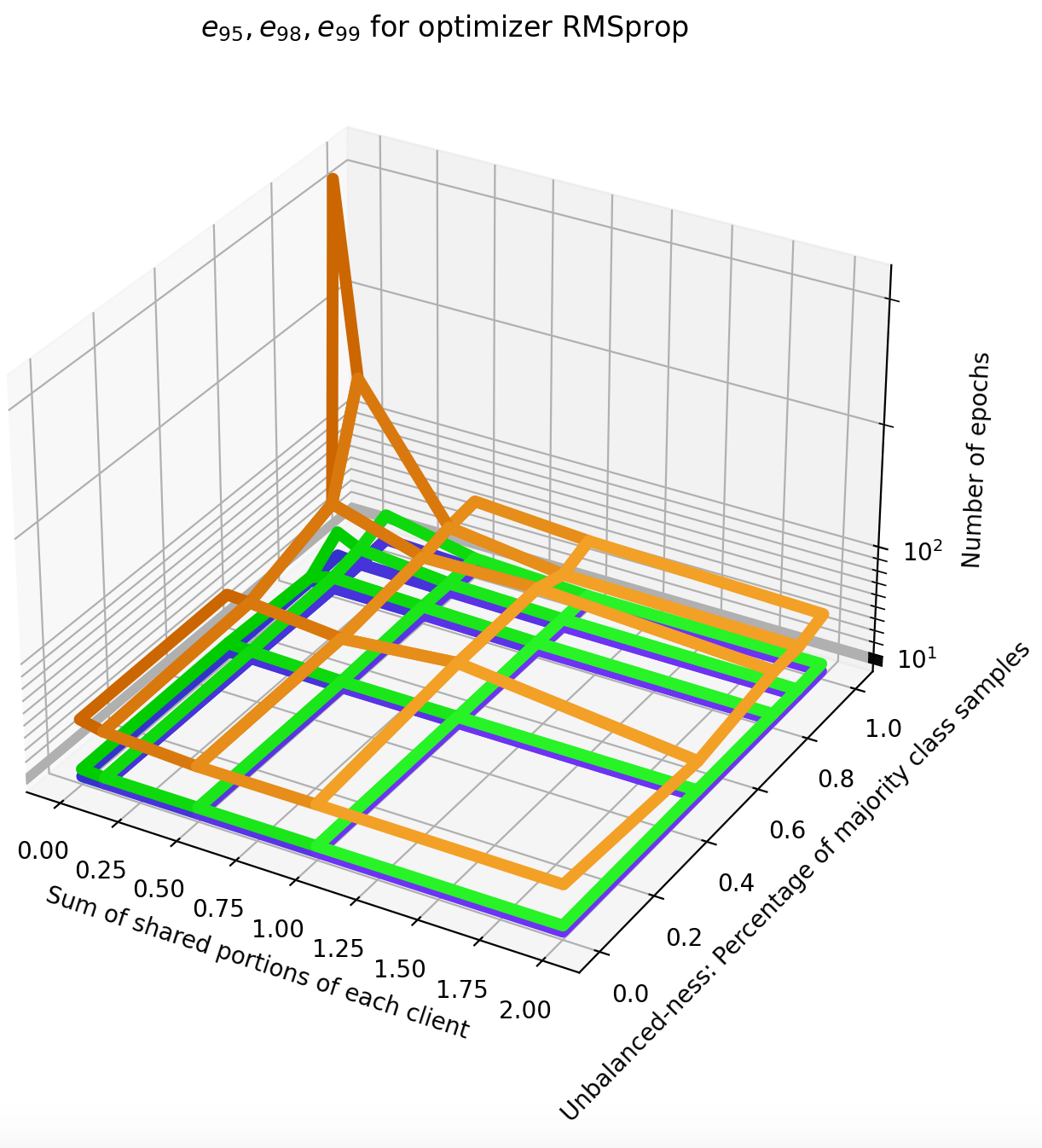}}
    \caption{Comparisons of $e_{95}$(blue),$e_{98}$(green),$e_{99}$(orange), with different values of $S$ and $U$ under different optimizers. Missing points indicate that the corresponding accuracy is never reached before the final epoch.}
    \label{fig:use}
\end{figure}

\begin{figure}[!htbp]
    \centering
    \subfigure[$e_{95}$]{\includegraphics[width=0.44\textwidth]{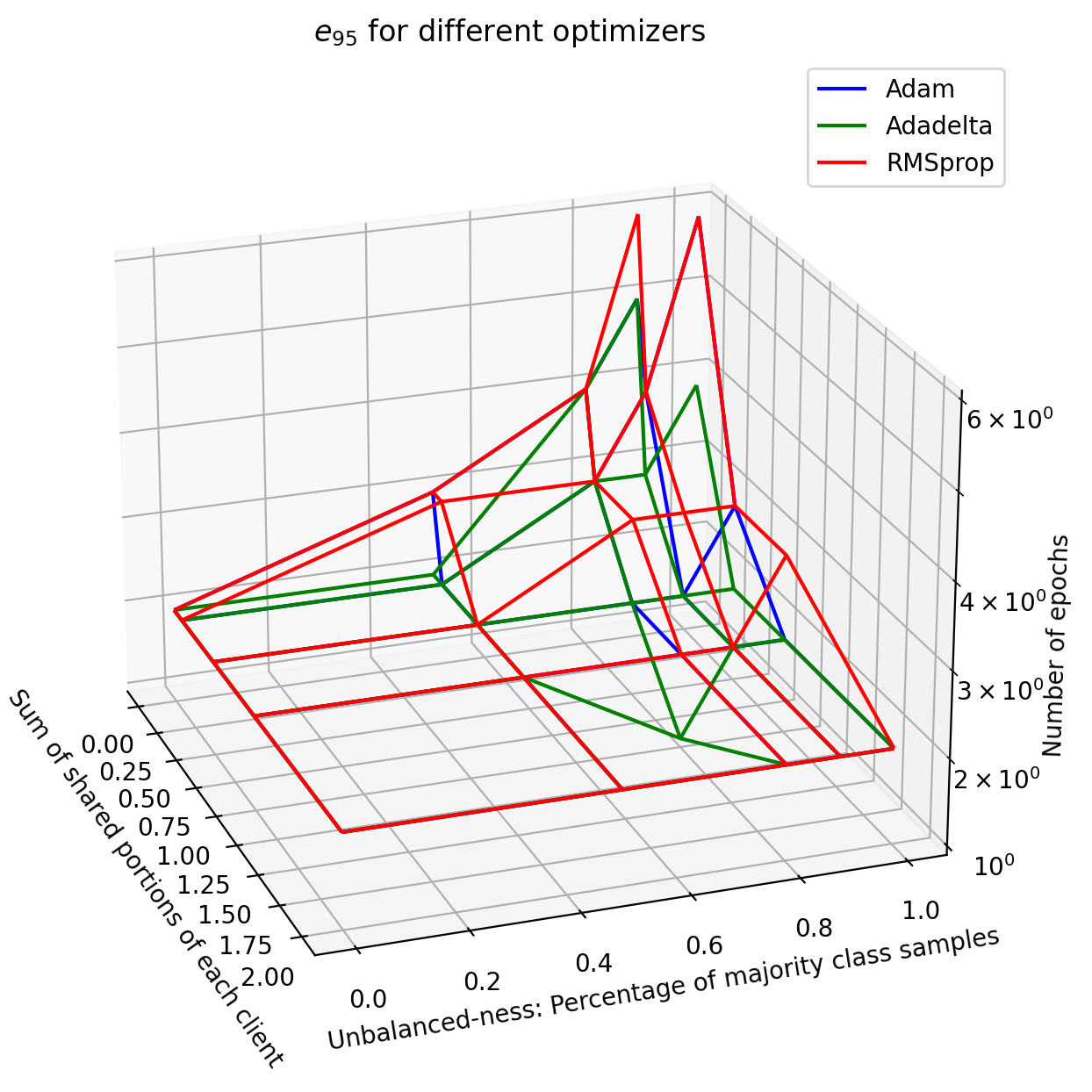}}
    \subfigure[$e_{98}$]{\includegraphics[width=0.44\textwidth]{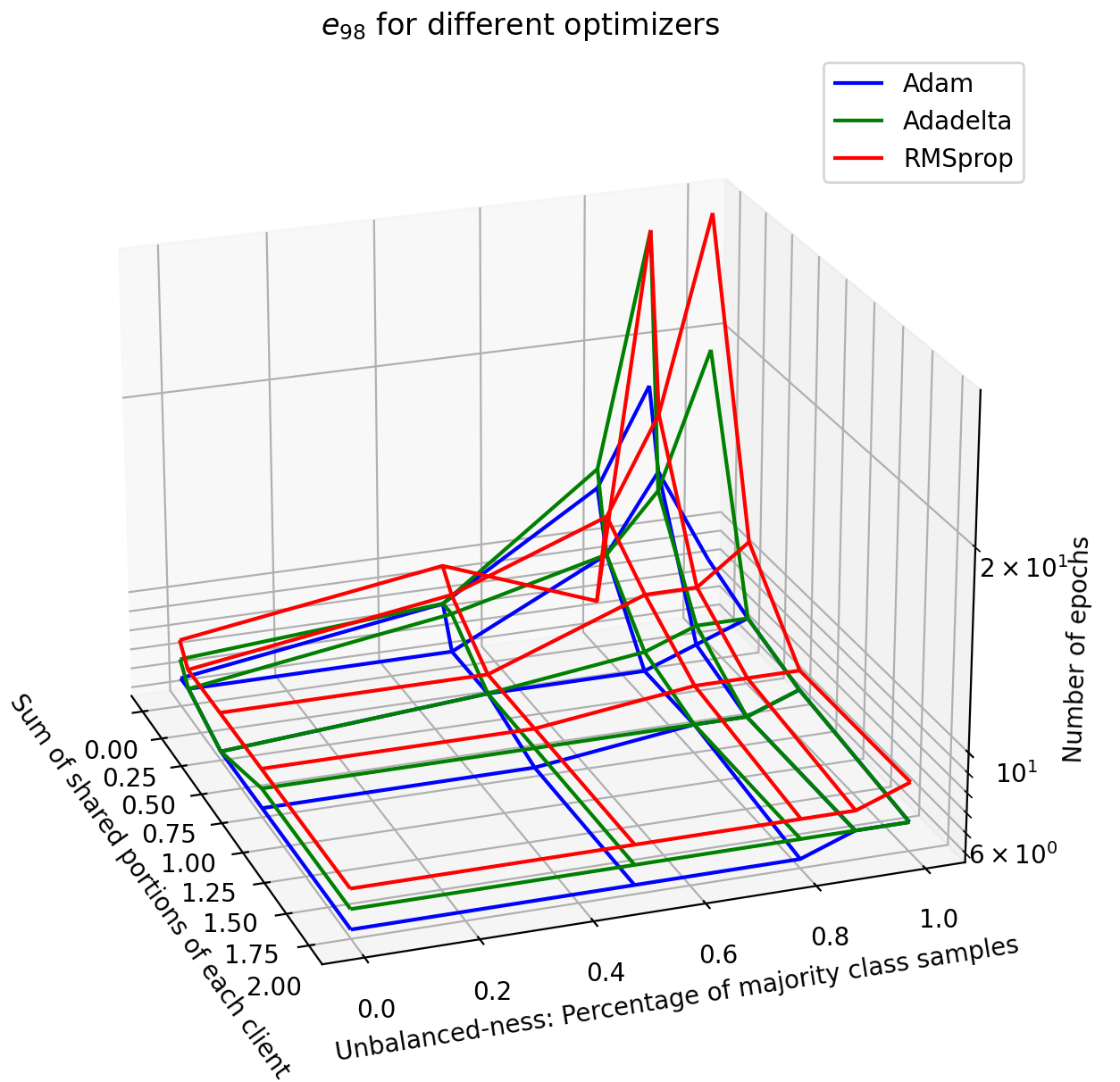}}
    \subfigure[$e_{99}$]{\includegraphics[width=0.44\textwidth]{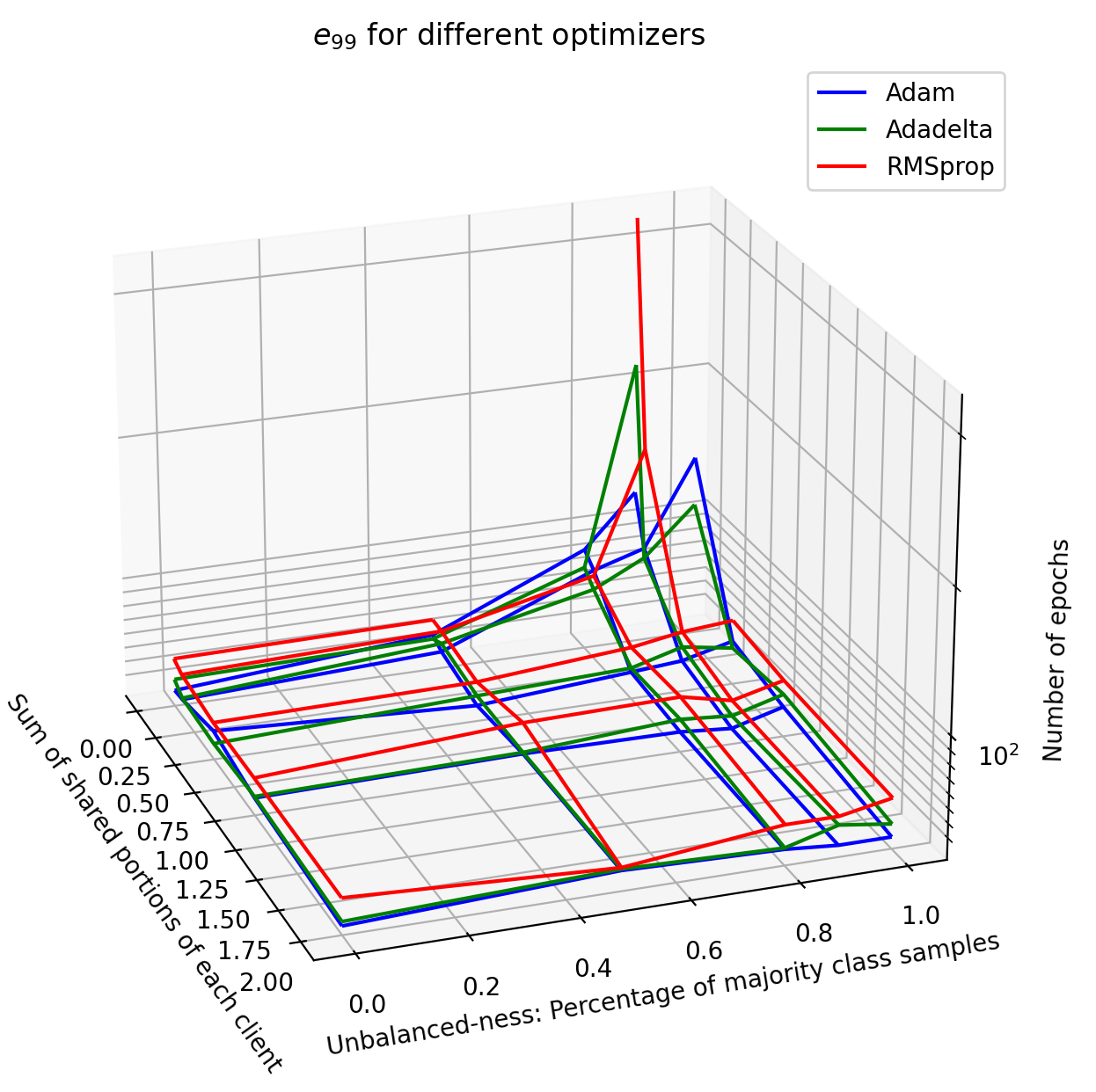}}
    \subfigure[Max performance]{\includegraphics[width=0.44\textwidth]{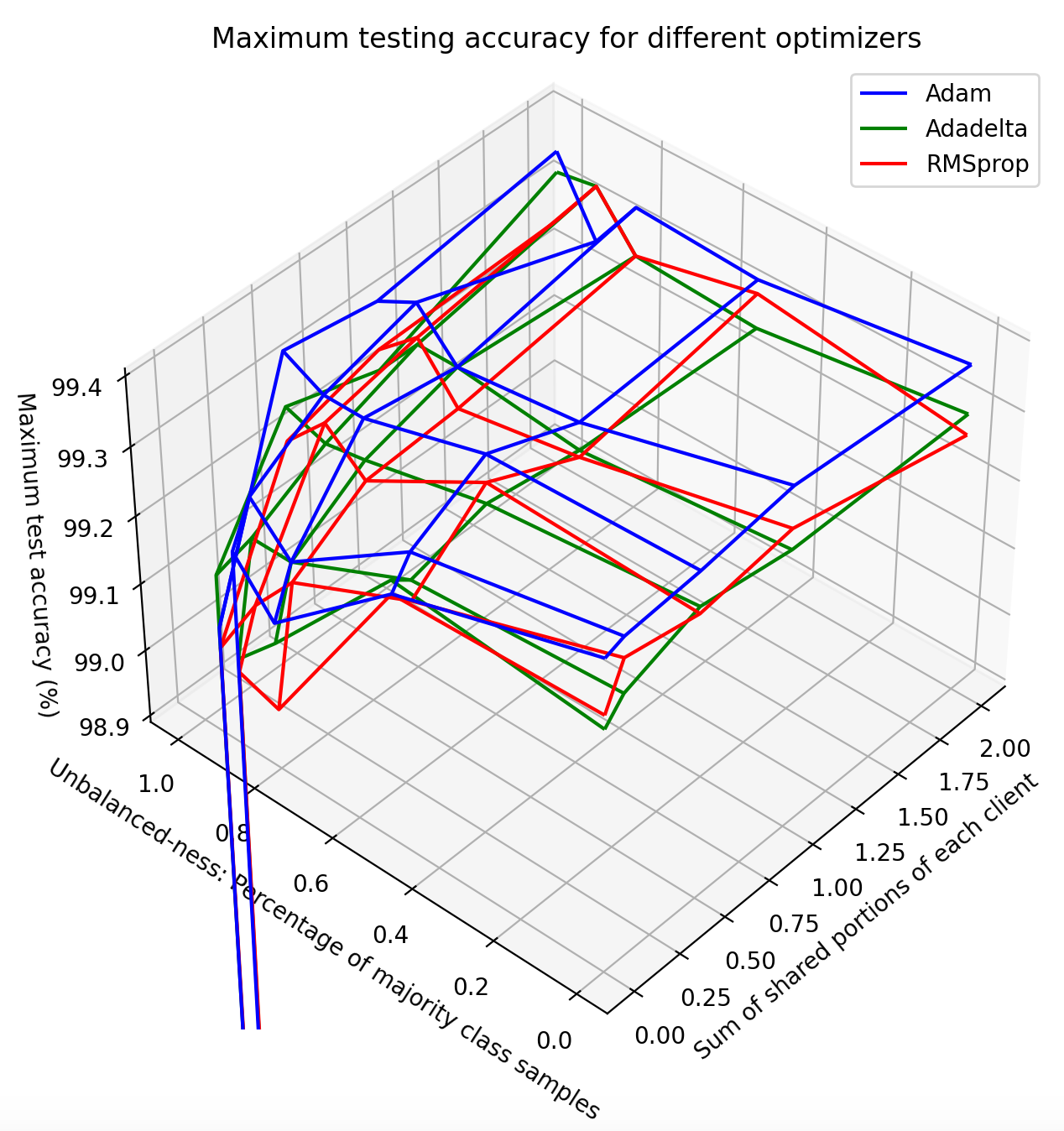}}
    \caption{Comparisons of different optimizers' performances with different values of $S$ and $U$ at  thresholds$e_{95},e_{98},e_{99}$, as well as best performance accuracies. Missing points in (a-c) indicate that the corresponding accuracy is never reached before the final epoch. (d) is cropped to exclude values too low.}
    \label{fig:usmax}
\end{figure}

We train the models with around 300 epochs and with $E=1$ for a more practical update frequency.
In addition, we record the epoch numbers $e_{95}, e_{98}, e_{99}$ for accuracy thresholds 95\%, 98\%, and 99\% respectively. 
The models all start with uniform initial weights.
We test with the top-performing optimizers (Adam, Adadelta, RMSprop) with same learning rates from Table \ref{tab:lr}. 
The results are shown in Figure \ref{fig:use} and \ref{fig:usmax}. 

The general trend, as observed in Figure \ref{fig:use}, is that the number of epochs it takes for the client models to reach a certain accuracy is larger if $S$ is smaller or $U_i$ is larger.
This is most clearly illustrated by $e_{99}$. 
Because we train with $E=1$, the shape of $e_{98}$ and $e_{95}$ has much less variation, but the trend is still visible. 
Figure \ref{fig:usmax}(a-c) plots the same data, but in a different view that compares optimizers. 
In addition, Figure \ref{fig:usmax}(d) plots the max average model accuracies, which mostly stay around 99\% regardless of $S$ and $U$.
The only bad performance comes from when $S=0$ and $U=1$, the same condition that leads to the worse performances in \cite{skewedFL}.

It should be pointed out that other researchers have also come up with different ideas for reducing variance caused by skewed datasets, such as the SCAFFOLD algorithm proposed by~\cite{scaffold}.

\subsection{Differently Normalized Training Data}
\begin{figure}[!htbp]
    \centering
    \subfigure[Adam, $S=0$, same initial weights. Max accuracy 97.64\%. ]{\includegraphics[width=0.32\textwidth]{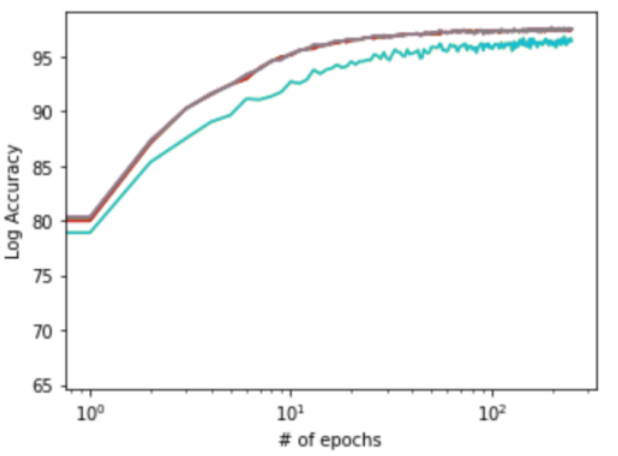}}
    \subfigure[Adadelta, $S=0$, same initial weights. Max accuracy 97.34\%. ]{\includegraphics[width=0.32\textwidth]{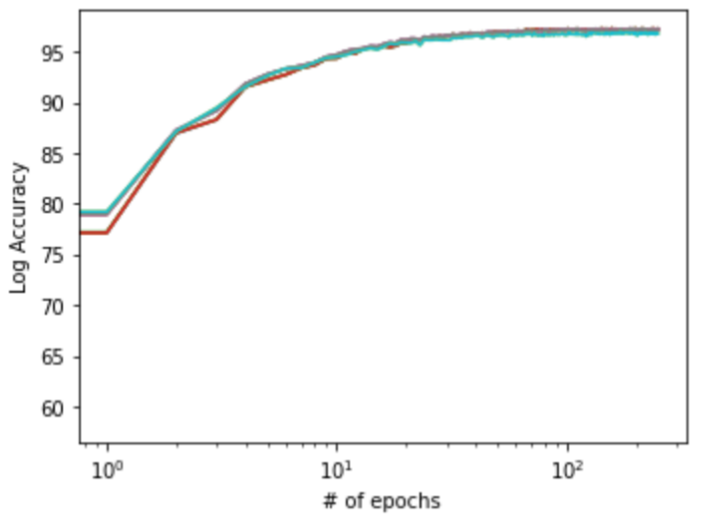}}
    \subfigure[Adadelta, $S=0$, different initial weights. Max accuracy 97.55\%. ]{\includegraphics[width=0.32\textwidth]{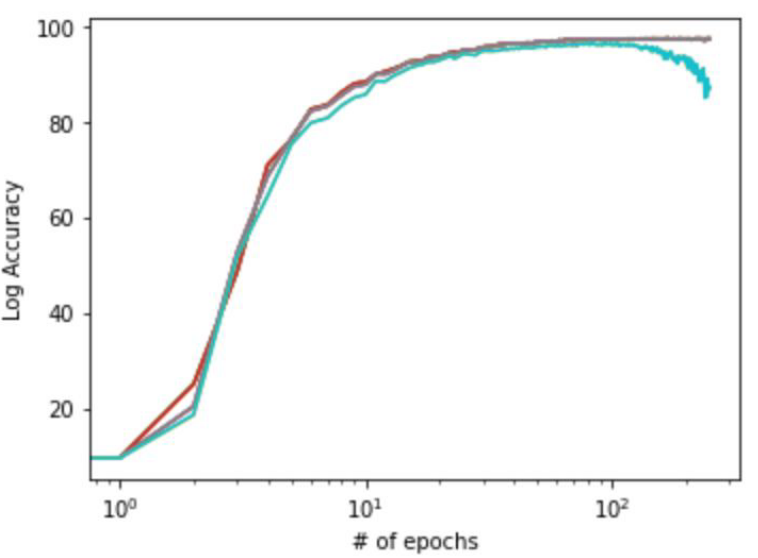}}
    \subfigure[Adam, $S=4$, different initial weights. Max accuracy 96.49\%. ]{\includegraphics[width=0.32\textwidth]{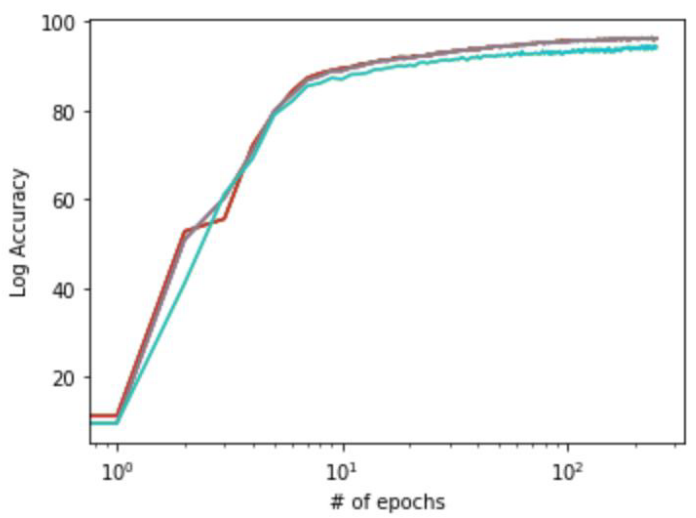}}
    \subfigure[Adam, $S=0$, same initial weights. Max accuracy 98.96\%. ]{\includegraphics[width=0.32\textwidth]{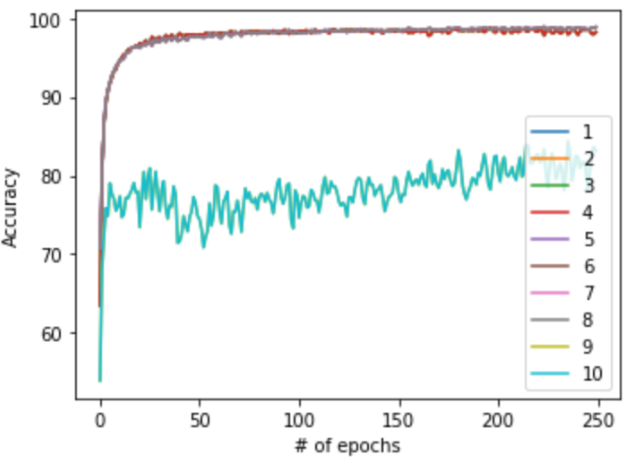}}
    \subfigure[Adadelta, $S=0$, same initial weights. Max accuracy 98.97\%. ]{\includegraphics[width=0.32\textwidth]{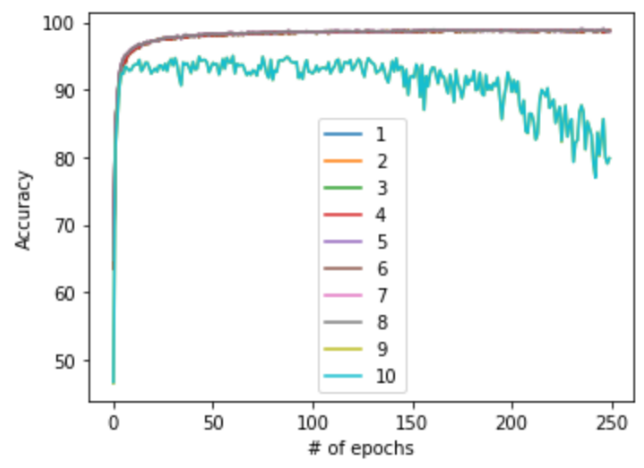}}
    \caption{Comparisons of test accuracy histories for clients with different normalizations. In (a-d), clients were trained with normalization parameters $\mu_{1:4}=\mu_M-0.1,\mu_{5:8}=\mu_M+0.1, $and $\mu_{9:10}=\mu_M+0.3$. The histories of client 1-8 overlap as the brown line, while those of client 9-10 overlap as the cyan line. In (e-f), the normalization parameters are $\mu_{1:4}=\mu_M-0.2,\mu_{5:8}=\mu_M+0.2, $and $\mu_{9:10}=\mu_M+0.6$, and the figures contain legends for clients.}
    \label{fig:norm01}
\end{figure}

In the previous experiments, the training data are pre-processed by normalization using the collective mean $\mu_M$ and variance  $\sigma_M$ computed from the entire training dataset, the same as the preprocessing in \cite{pytorchMNIST}.
However, in applications where the data are truly distributed and private, it could be hard to obtain the correct global statistics of the training data.
Instead, clients would have to rely on local data, and calculate normalization parameters from what's available.
Thus, in this section, we investigate if the training could be successful with different normalizations on the MNIST dataset.
In particular, our experiments shift the normalization mean for each client by varying amounts.
$\mu_i$ is the normalization mean that each client uses on its raw data. 
For the first experiment, we define
\begin{align}
 \mu_{1:4}=\mu_M-0.1, \; \mu_{5:8}=\mu_M+0.1, \; \mu_{9:10}=\mu_M+0.3   
\end{align}
Figure \ref{fig:norm01}(a-d) reports the model performance history in this setting by picking representative experiment results. 
The normalization parameters $\mu_{9:10}$ for clients 9 and 10 were further from $\mu_M$, and those models show worse performances.
However, also note that the top accuracy for each model is lower than the accuracy seen in other experiments with uniform normalizations.

Further increasing the difference in mean values provided more drastic outcomes, as shown in Figure \ref{fig:norm01}(e-f). 
While some models are able to achieve better performances, the clients with more normalization mean deviation perform much more poorly.
This indicates that, while bad normalization parameters could hurt the performances (as shown by client 9 and 10), the majority of the clients' models are robust against this with averaging. 

\begin{figure}[!htbp]
    \centering
    \subfigure[$S=0$ ]{\includegraphics[width=0.32\textwidth]{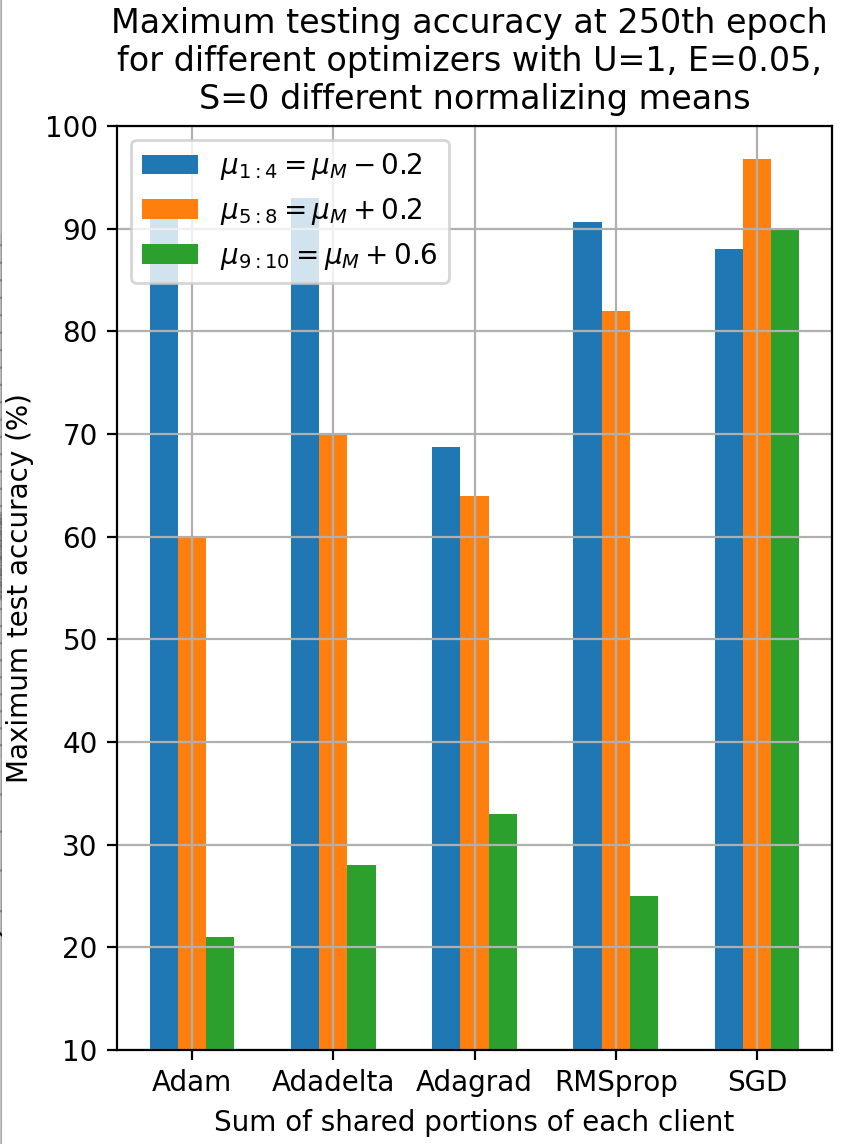}}
    \subfigure[$S=0.1$ ]{\includegraphics[width=0.32\textwidth]{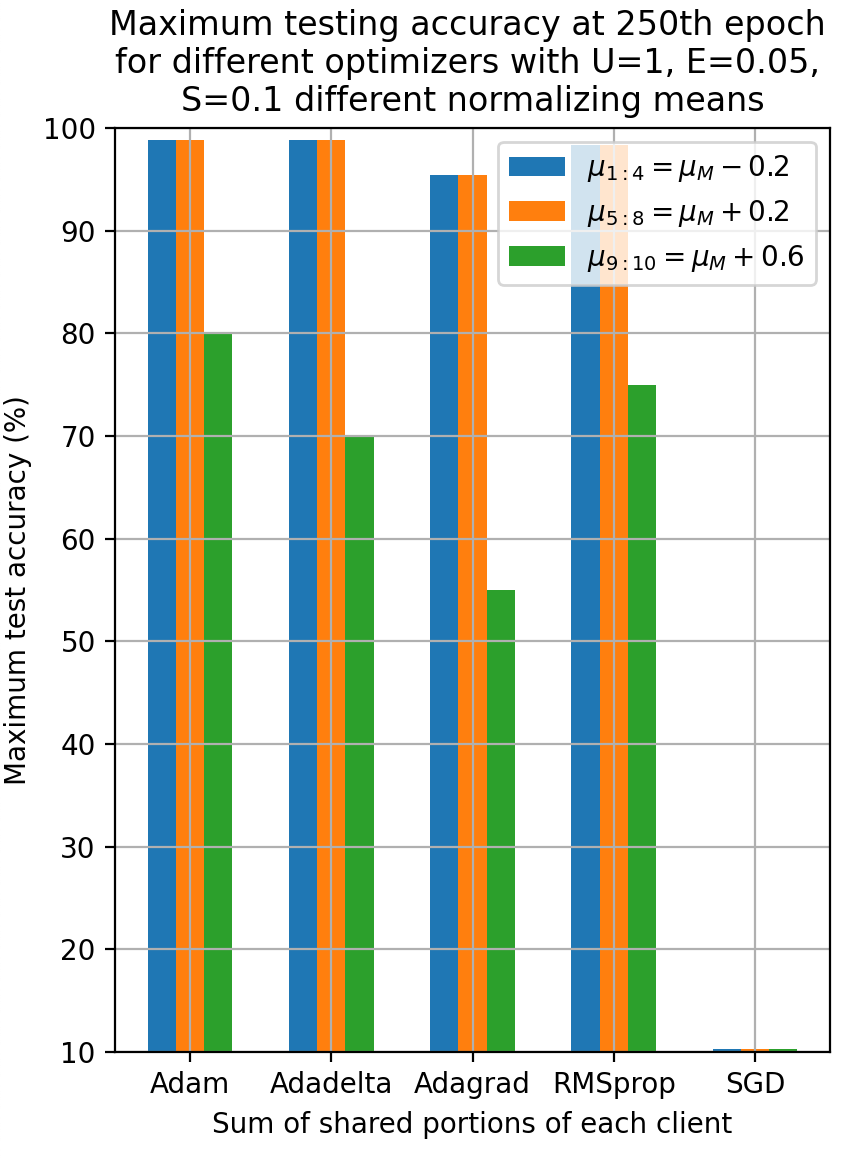}}
    \subfigure[$S=0.4$ ]{\includegraphics[width=0.32\textwidth]{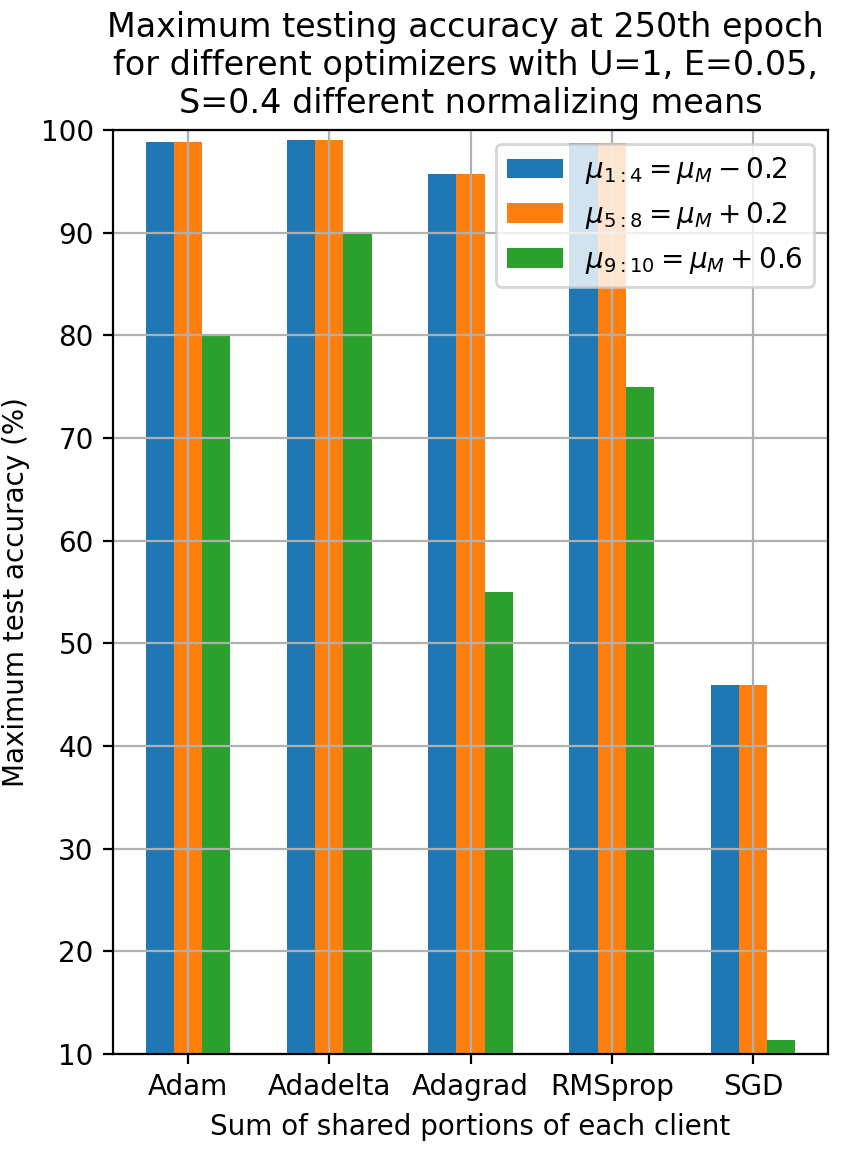}}
    \caption{Accuracies for training under skewed dataset with normalization parameters $\mu_{1:4}=\mu_M-0.2$, $\mu_{5:8}=\mu_M+0.2$, and $\mu_{9:10}=\mu_M+0.6$. Each presented value is the peak accuracy at the final 10 epochs of a 250-epoch training process.}
    \label{fig:normU1}
\end{figure}

\begin{figure}[hbt!]
    \centering
    \subfigure[Epoch counts ]{\includegraphics[width=0.47\textwidth]{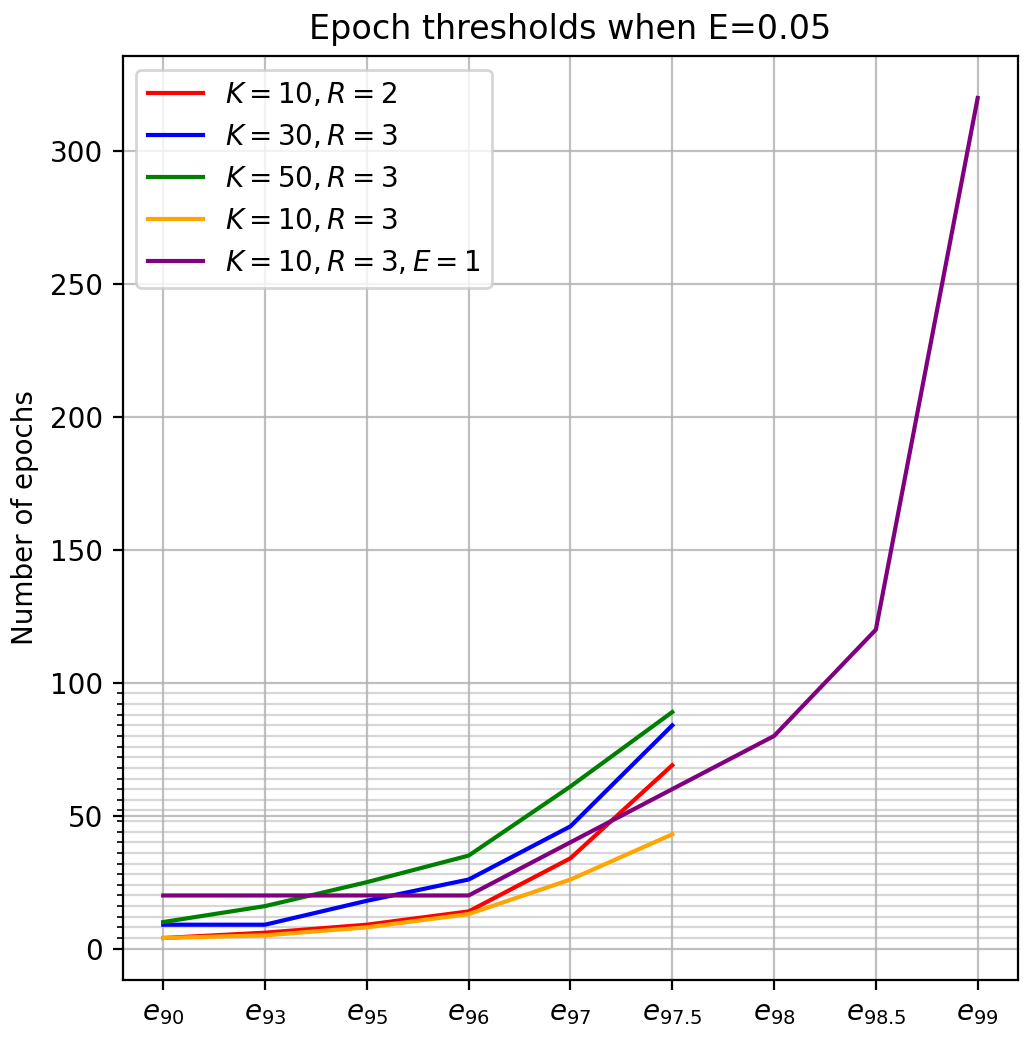}}
    \subfigure[Max accuracies ]{\includegraphics[width=0.5\textwidth]{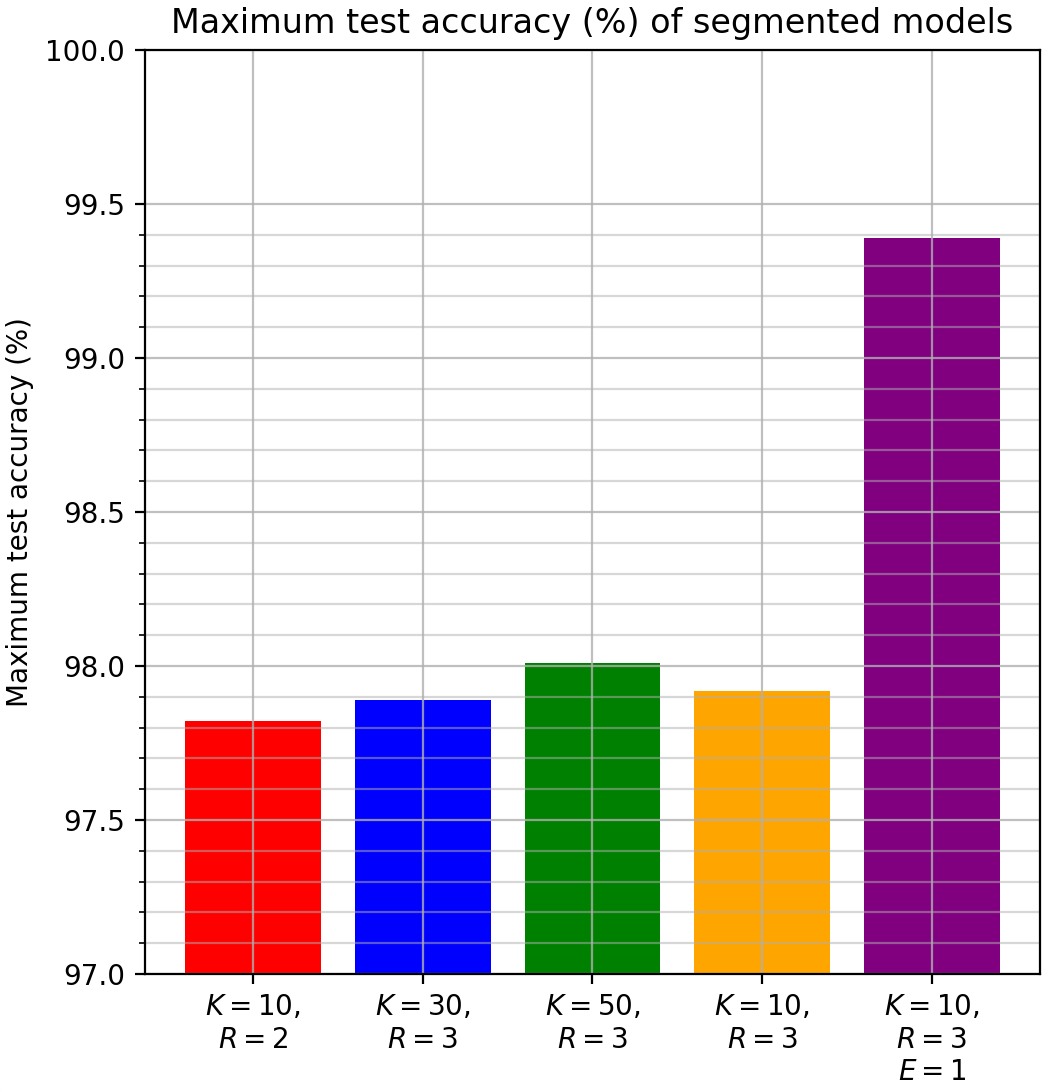}}
    \subfigure[Pairwise model distances for $K=10,R=3$ ]{\includegraphics[width=0.75\textwidth]{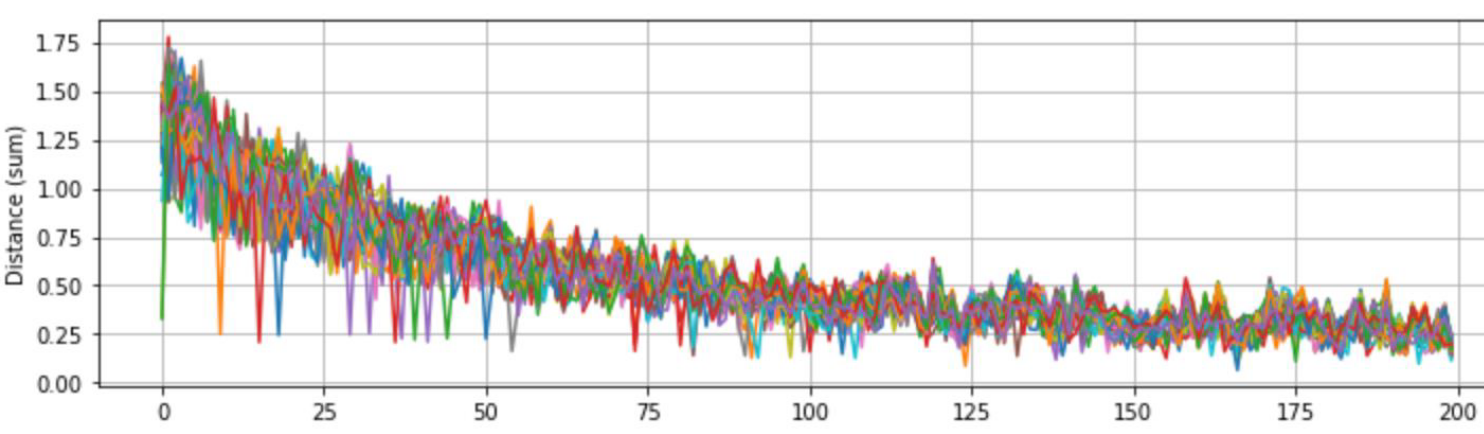}}
    \subfigure[Pairwise model distances for $K=50,R=3$ ]{\includegraphics[width=0.75\textwidth]{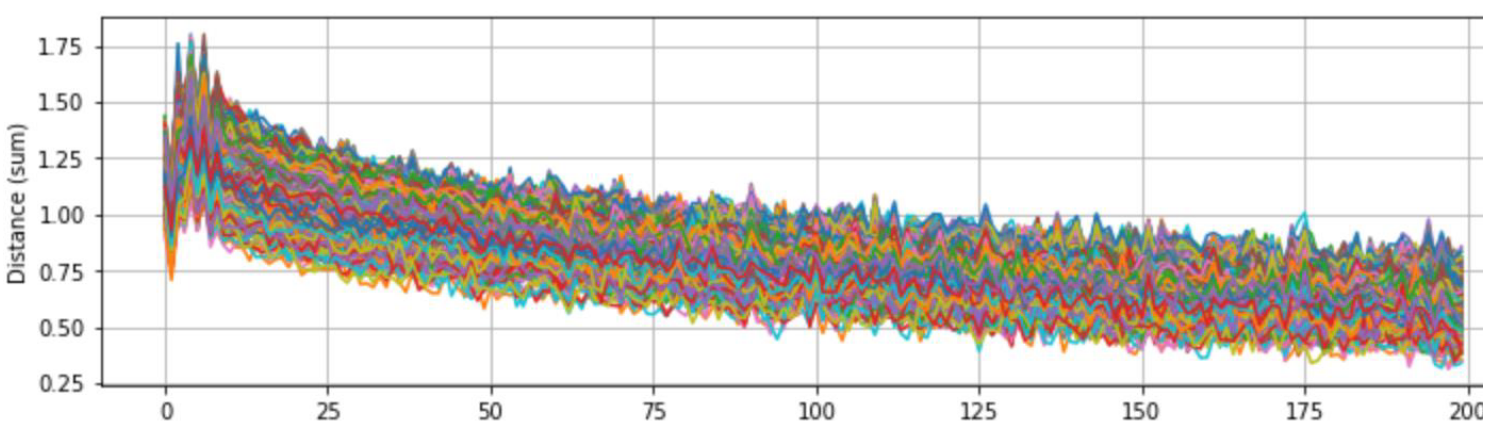}}
    \subfigure[Pairwise model distances for $K=10,R=3,E=1$ ]{\includegraphics[width=0.75\textwidth]{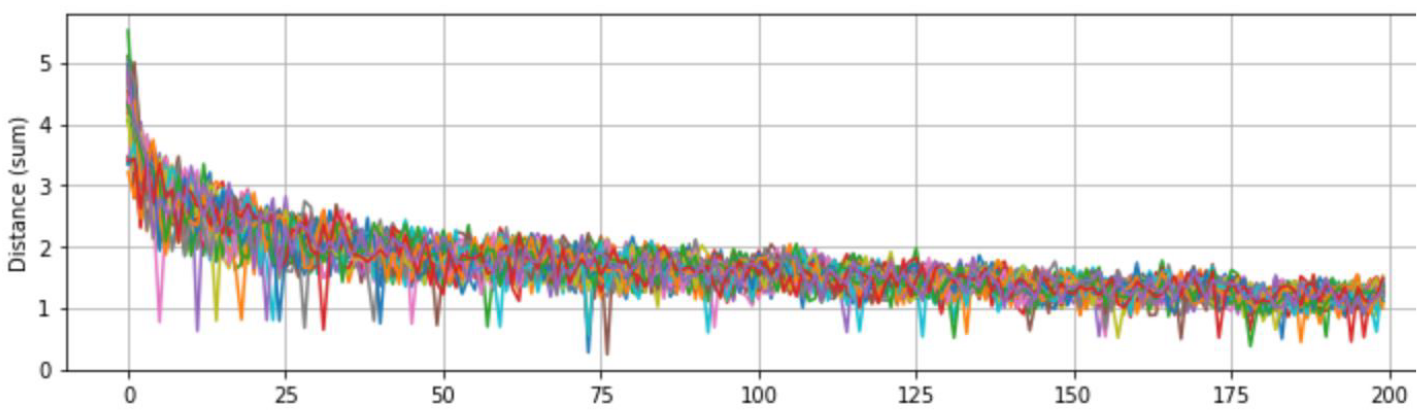}}
    \caption{(a-b): Epoch threshold and maximum accuracies of segmented federated learning with layer-wise sharing. All tested with $E=0.05$ unless otherwise specified. In (a), the $E=1$ data was obtained by multiplying the original data with $20 = \frac{1}{0.05}$. (c-e) History of every possible pair of clients' model parameter distance, using $\ell_2$ norm, over 200 epochs.}
    \label{fig:seg0205}
\end{figure}

Figure \ref{fig:normU1} presents the results from all experiments. 
When $S \geq 0.1$, the training is again stabilized, as demonstrated by the performance improvements. 
The maximum accuracy that each group of clients attained reveals the same trends that more sharing improves performance, and more normalization parameter deviation leads to worse performance.

Note that the values when $S=0$ are obtained from the last 10 epochs of training over 250 epochs, instead of from picking a historical max accuracy. 
This is because the max accuracy histories are noisy, and display large fluctuations. 

\section{Different Optimization Methods} \label{sec:optimethods}
Previously, we have seen that certain optimizer-learning rate pairs perform better than others in the same job, such as in Figure \ref{fig:Skewed}. 
In practice, we may deploy clients with different optimization setups, such as using different optimizers and learning rates, when it is not clear which combination would work best.
For this section, we test the robustness of DFL, when clients use different optimization methods, but share the same optimization goal. 

\subsection{ Different Learning Rates}
In this section, we use Adadelta as the optimizer, with learning rate ranging from 0.001, 0.01, 0.1, 1, and 5 across $K=10$ different clients. 
The training dataset is partitioned with $U=0$, but sizes differed with $\sigma^2 = 0.7$.
In addition, to stabilize the training, we use data sharing with $S=0.1$.
$lr_i$ is the learning rate of client $i$. 

\begin{figure}[!htbp]
    \centering
    \subfigure[$E=0.05$]{\includegraphics[width=0.32\textwidth]{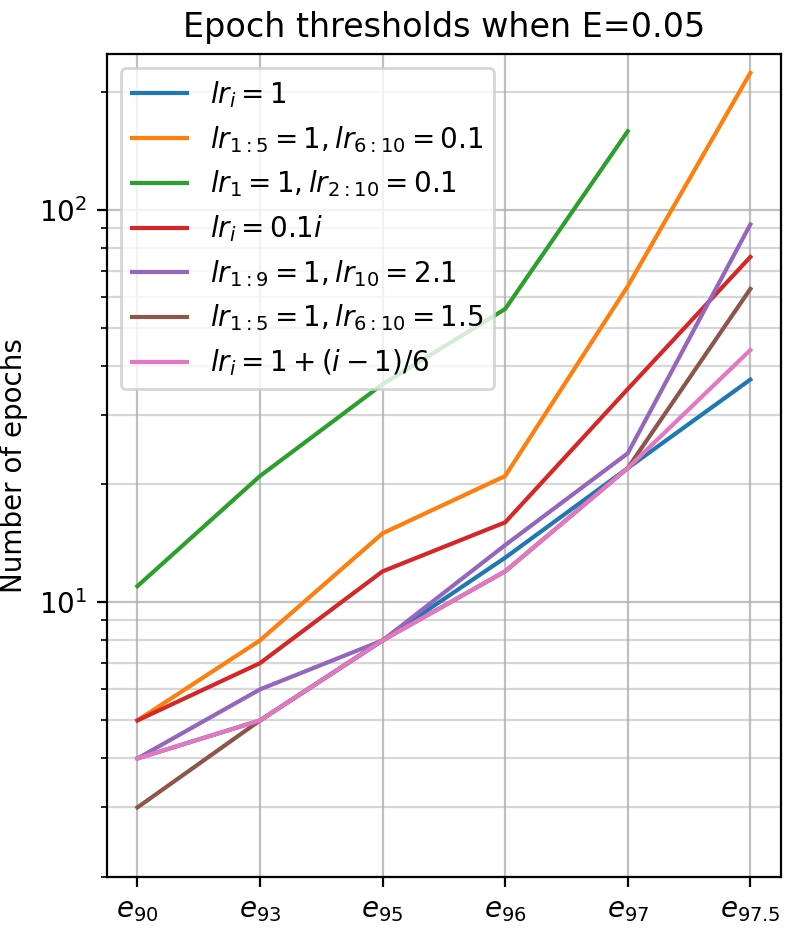}}
    \subfigure[$E=1$]{\includegraphics[width=0.32\textwidth]{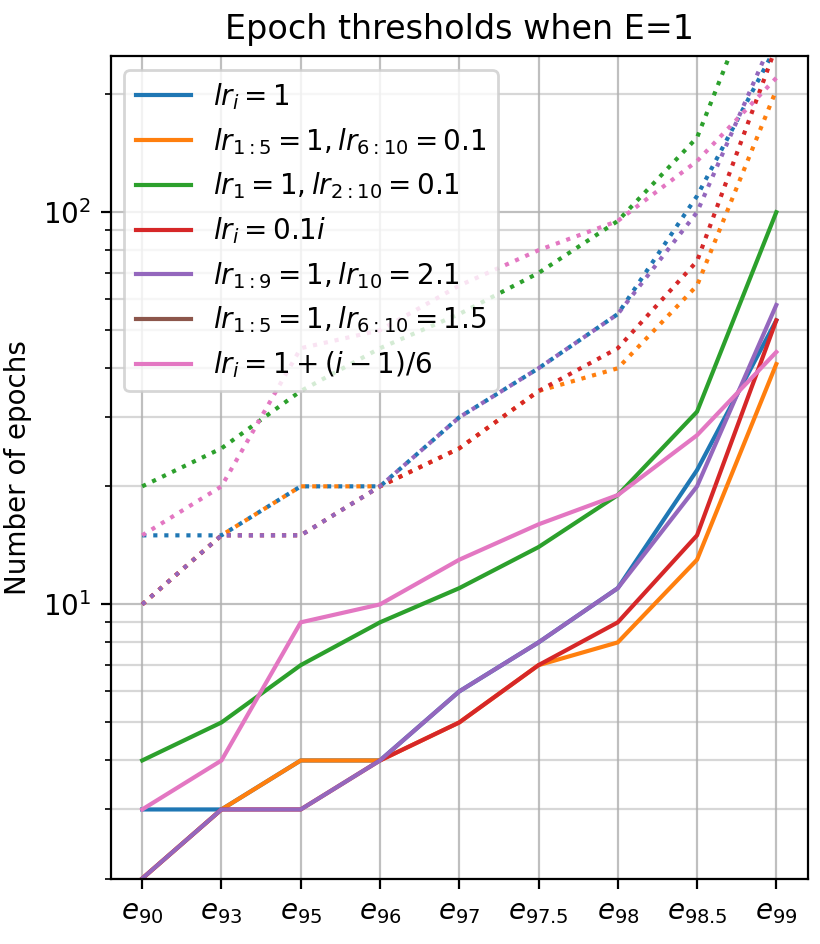}}
    \subfigure[Various $E$]{\includegraphics[width=0.32\textwidth]{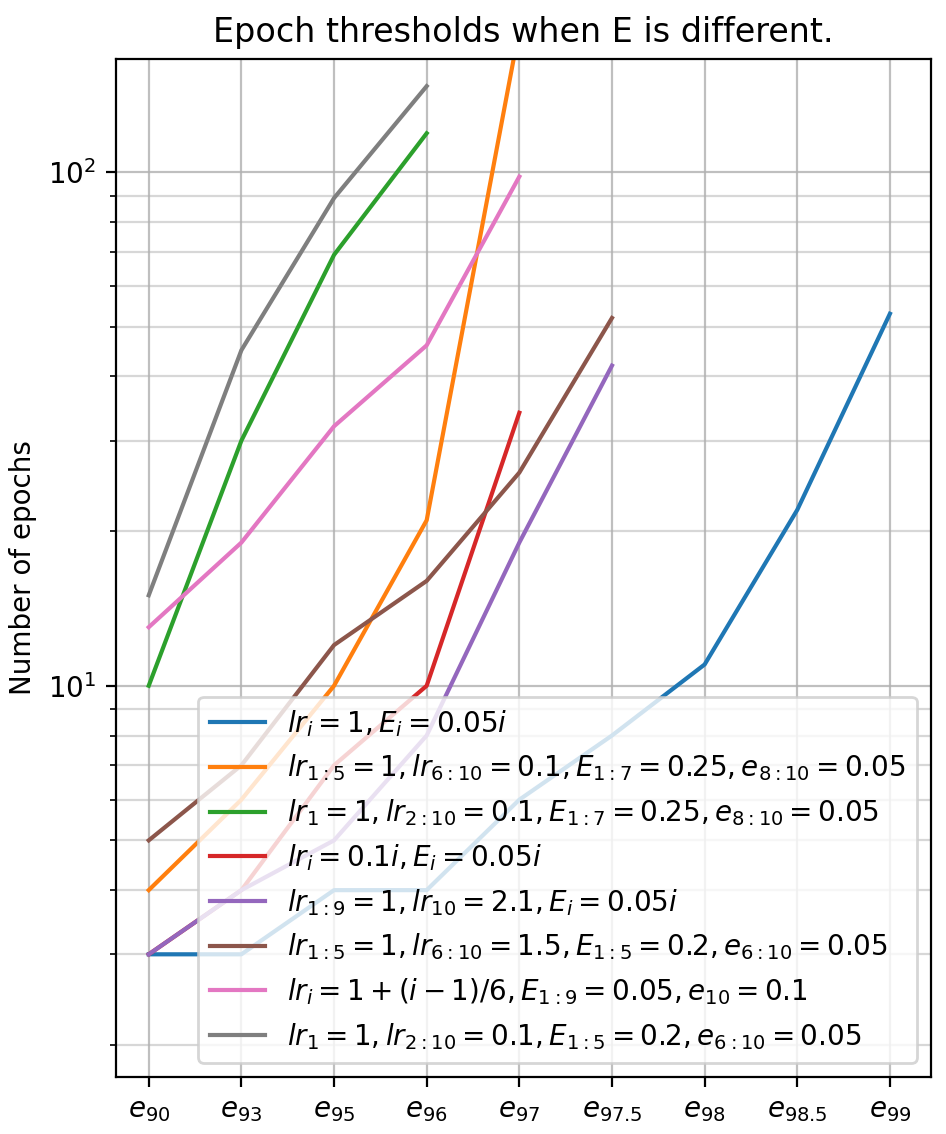}}
    \subfigure[Maximum test accuracy percentages]{\includegraphics[width=0.85\textwidth]{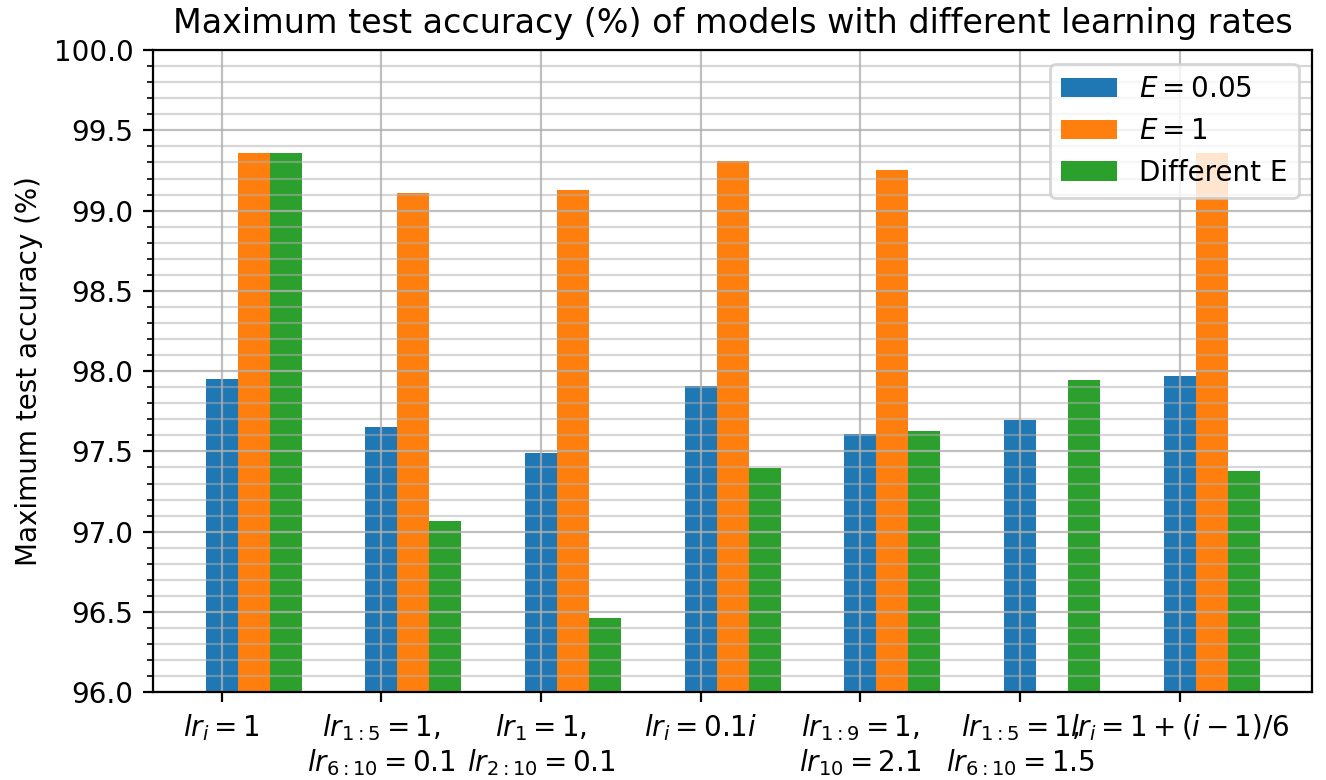}}
    \caption{(a-c): Epoch threshold values with different combinations of learning rates and learning speeds over 500 epochs, as specified in the legend. (d) Maximum test accuracy values of models in (a-c); the missing bar is where models diverged. In (b), the dashed lines are the solid lines scaled up by 5.}
    \label{fig:lr0205}
\end{figure}

The results are shown in Figure \ref{fig:lr0205}, in which we try different kinds of learning rate combinations: 
\begin{itemize}
    \item $lr_1 = ... = lr_5 < lr_6 = ... = lr_{10}$;
    \item $lr_1 \neq lr_2 = lr_3 = ... = lr_{10}$;
    \item $lr_i = a+bi, i = 1,...,10$
\end{itemize}

Figure \ref{fig:lr0205}(a) and (b) show the performance with $E=0.05$ and $E=1$, respectively.
Those figures show that the clients could train successfully in these settings.
In general, higher average learning rates lead to smaller epoch numbers and faster learning rate, but there is no clear-cut trend from the results.

Figure \ref{fig:lr0205}(c) presents results when we also combine different learning speeds for different clients. 
For example, if agent $i$ has learning speed $E_i =0.25$ while others have speed $E = 0.05$, then client $i$'s model gets updated 5 times more than other clients within one epoch.
According to the figure, all the models are able to converge. However, from Figure \ref{fig:lr0205}(d), the setups with uniform aggregation frequencies would perform better.
We also observe that slower aggregation frequency with $E=1$ achieves better test accuracy, similar to what we observe in the preliminary experiments as well.

The results from the figures showed that, if the learning rates are conservative, then varying learning rate does not significantly affect the rate of training.  
However, we note that the presented results are from experiments that converge to good-performing models.
If the learning rates are too large, the models risk diverging.
This is what happened when $E=1$ with learning rate distributions as $lr_{1:5}=1$, $lr_{6:10}=1.5$, as shown by the lacking of data from $E=1$ in Figure \ref{fig:lr0205}(d).
Conversely, if the learning rates are too small, then model learning would be too slow to keep track of efficiently. This is tested, but not presented here.

\subsection{ Different optimization algorithms}
In this section, we assign different optimizers for clients.
We pick five optimizers included in \cite{pytorchMNIST} that are suited for classification tasks:
\textit{Adam}, \textit{Adagrad}, \textit{Adadelta}, \textit{RMSprop}, and \textit{SGD}.
The learning rate for each optimizer is chosen empirically: 0.001, 0.01, 0.005, 0.0005, and 0.05, respectively.
Note that these values are slightly different from what is in Table \ref{tab:lr}, but the orders of magnitudes are similar.

\begin{figure}[!htbp]
    \centering
    \subfigure[]{\includegraphics[width=0.5\textwidth]{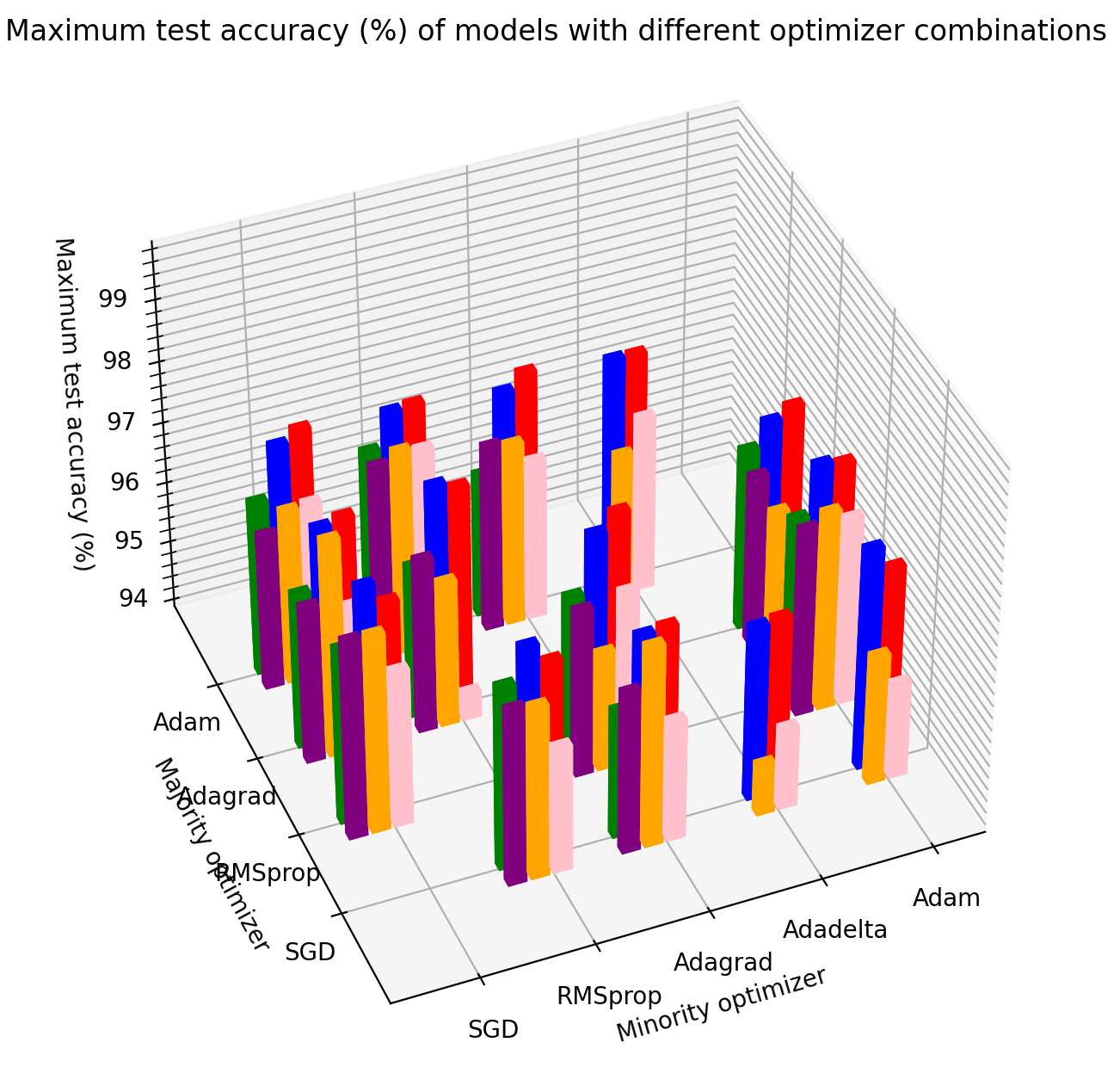}}
    \subfigure[With the plane marking 97\%]{\includegraphics[width=0.47\textwidth]{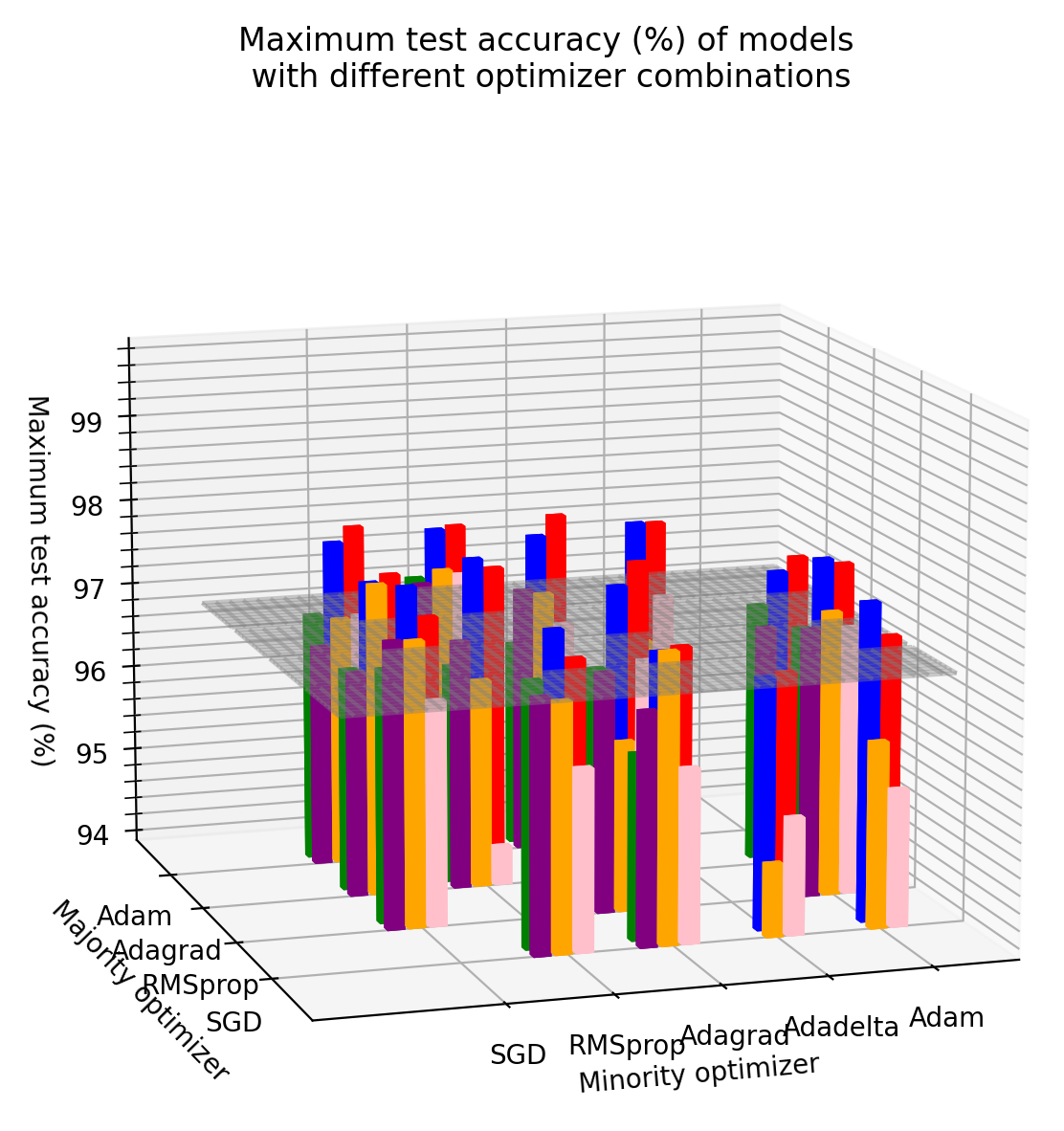}}
    \caption{Maximum test accuracy for clients with mixed optimizers, trained over 200 epochs; empty positions represent combinations that were not run. The red, blue, and green bars represent optimizer numbers as $9+1, 7+3$, and $5+5$. The other three bars correspond to experiments without same initial weights.}
    \label{fig:twoSGD}
\end{figure}

Again, we choose the training setup parameters to be $E=0.05$, $S=0.1$, $\sigma^2 = 0.7$, $U=0$ with $K=10$ clients. 
Figure \ref{fig:twoSGD} plots the maximum accuracy achieved by each of the combination that is tested.
For each pair of optimizer combination that we test, 
we try 3 different ratios between the optimizers: 9 optimizers of one type plus 1 optimizer of another type; 7 of one type plus 3 of another; and 5 of each type.
In addition, we test the performances of those setups under random initial model weights.

The results reveal that combining any two optimizing methods could lead to convergence to a satisfactory model for the MNIST classification problem, with or without uniform initial weights.
However, the resulting accuracy values were lower than the baseline and the experiments done with uniform optimizers in previous sections. 

In Figure \ref{fig:twoSGD}(b), this is revealed by the grey surface that marks the 97\% accuracy threshold.
Note that none of the training where 5 clients used one optimizer and 5 used the other is able to reach an accuracy threshold of 97\%.
In comparison, experiments where one optimizer is used by the majority of the clients tends to have higher accuracy values.

\begin{figure}[!htbp]
    \centering
    \subfigure[$E=0.05$]{\includegraphics[width=0.35\textwidth]{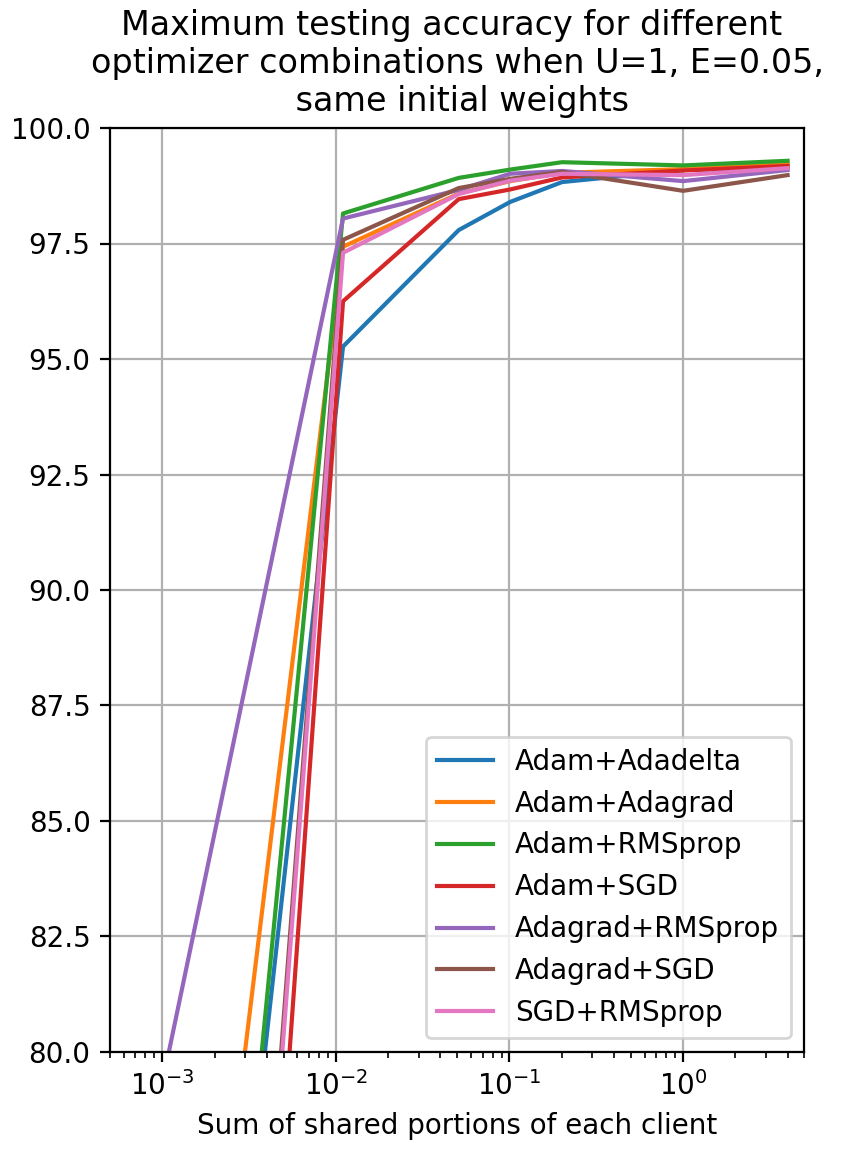}}
    \subfigure[$E=1$]{\includegraphics[width=0.35\textwidth]{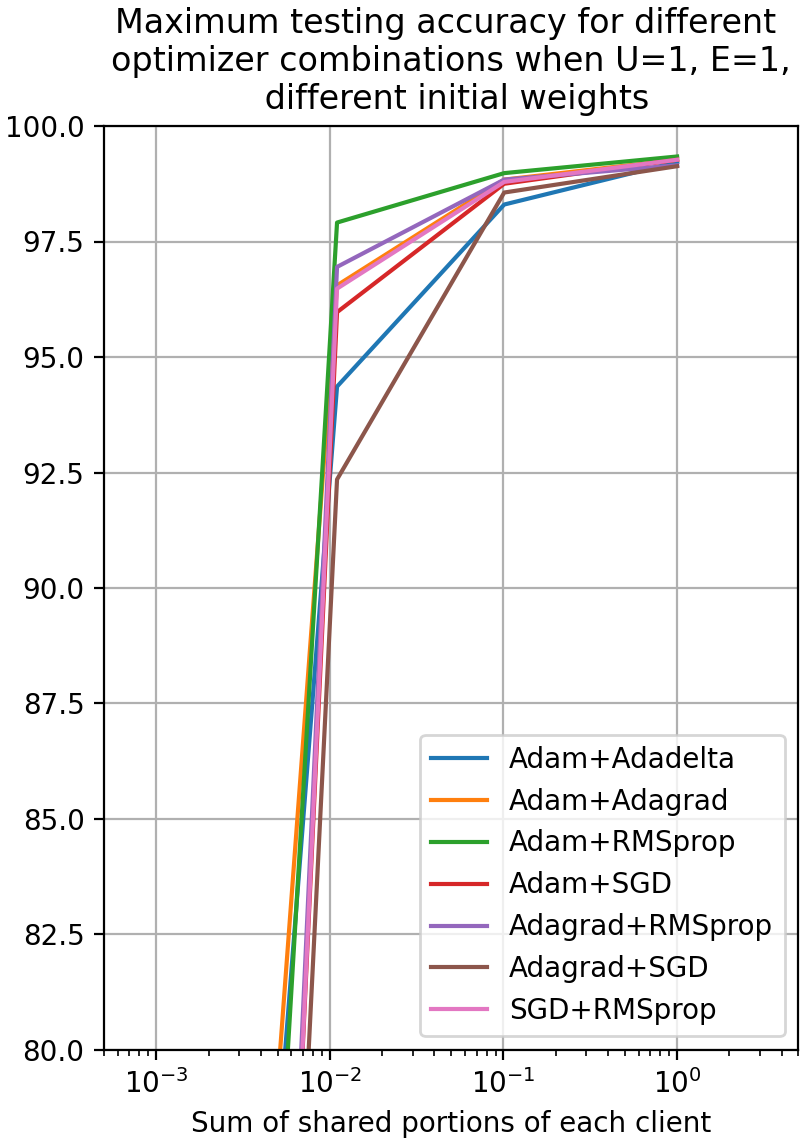}}
    \caption{Maximum test accuracy for clients with mixed optimizers, trained over 200 epochs, with fully skewed datasets.}
    \label{fig:twoSGDunb}
\end{figure}

The models with different optimizers and skewed training data partitions with $U=1$ are still able to converge to a satisfactory global model, as long as $S \geq 1\%$, as shown in Figure~\ref{fig:twoSGDunb}. 
Those models are trained for 500 epochs for $E=0.05$, and the optimizers are assigned to 5 clients each.
It appears that combining different optimizers for skewed datasets lead to better performances than combining different optimizers on IID data partitions. 
These observations hint that, averaging the models trained using different optimizers could result in suboptimal performances when compared to baseline models, 
but might provide effective learning when the training dataset is also nonuniform. 
However, more experiments would be needed to verify this hypothesis.

\section{Segmented Federated Learning} \label{sec:segfl}
One common setup for all previous experiments is that they have to share the full set of weights each epoch. 
This can be costly in communication, especially in edge computing applications with less resources for communication, and scenarios with huge amount of devices. 
Thus, we are curious if it's feasible to reduce communication cost by only sharing a segment of the entire model to a selected few neighbors. 
In this section, we introduce existing work, then propose more segmentation options, and examine the results.

\subsection{Existing Work}
\cite{hu2019decentralized} test this idea by dividing the entire model $W_i$ of a client $i$ into several segments $W_i^1, ..., W_i^M$. 
In the aggregation step, each client polls segments from a fixed set of $R$ other clients, and performs weighted averaging according to each client's dataset size. 
This procedure could be done in a gossip approach, to save communication cost while still giving convergence. However, it is unclear how they divided the model into "segments". 

As a proof-of-concept, we treat each layer as a segment, and randomly pick $R = 3$ clients out of the $K$ clients for each segment. 
The training setup is the same as previous default setups, where 
models start from the same initial weights, are
trained using the Adadelta optimizer with learning rate of 1, 
and use disjoint and IID partitions ($S=0$, $U=0$).
The results after 200 epochs is shown in Figure \ref{fig:seg0205}. 
The models can successfully converge to a set of parameters with sufficiently good performance with accuracy above 97\%.
For experiments with smaller $K$, the communication graph would be denser with fixed $R$. We observe that denser graph requires less number of epochs for each accuracy threshold. 

Figure \ref{fig:seg0205}(c-e) examines the model $\ell_2$ distance between clients.
The model distances between all pairs of clients are examined, regardless of whether the clients are connected or not.
The fact that the model distances are non-zero reveals that the model weights are not completely uniform at the end of each aggregation step, and at the start of each training epoch.
This is expected in this segmented method, because the communication graph is directed, and is possible to have several components.
 
However, from the results, we see that the models don't have to be the same to get good performance. 
For example, as \cite{matchedFL} has observed, networks with permutations in parameters could give exactly the same performance.  

In the next section, we formally define the Segmented Federated Learning approach in a decentralized manner.
\subsection{ Segment Units}
The model used under current study, a simple convoluted neural network (CNN), can be described by a list of tensors with dimension up to 4D.
Denote each layer as $W_{L_i}$, so a CNN with $l$ layers would have its whole model weight be 
$W = (W_{L_1}, ..., W_{L_l})$, where 
$W_{L_i} \in \R^{C_{i-1}\times C_{i} \times w_i \times h_i}$ if the $i$-th layer is a convolution layer,
$W_{L_i} \in \R^{(C_{i-1}\cdot I)\times C_i}$ if the $i$-th layer connects to a previous convolution layer where the image output would have size $I$,
and $W_{L_i} \in \R^{C_{i-1}\times C_{i}}$. 
Here, we use $C_{i-1}$ as the input channel size, $C_i$ as the output channel size, and $w_i\times h_i$ as the convolution kernel  size for layer $i$. 

Between client $i$ with weights $W^i$ and client $j$ with weights $W^j$, the whole model distance is defined as: 
\begin{align}
\norm{W^i-W^j} \triangleq \sum_{k=1}^l \norm{W^i_{L_k} - W^j_{L_k}}_2.
\end{align}

For some client $j$, we use the following notations for its CNN model weights: 
\begin{align*}
    \text{Full model } W^j &= [W^j_{L_1}, ..., W^j_{L_l}] \text{ for client } j \text{, with the number of layers being } l;\\
    \text{Single layer } W^j_{L_i} &= [W^j_{L_{i,out},c_1}, ..., W^j_{L_{i,out},c_{C_{i}}}] \text{, aggregated by output channels, or}\\
    &= [W^j_{L_{i,in},c_1}, ..., W^j_{L_{i,in},c_{C_{i-1}}}]^T\text{, aggregated by input channels};\\
    W^j_{L_{i,out},c_i} &= [k^j(i,1,c_{i}), ..., k^j(i,C_{i-1},c_{i})]  \in \R^{C_{i-1}\times w \times h} 
    \text{, with kernels for all }\\
    &\text{input channels corresponding to the }c_i\text{-th output channel,} 1 \leq c_{i} \leq C_i, \\
    W^j_{L_{i,in},c_{i-1}} &= [k^j(i,c_{i-1},1), ..., k^j(i,c_{i-1},C_i)] \in \R^{C_{i}\times w \times h}
    \text{, with kernels for all }\\
    &\text{ output channels corresponding to the }c_{i-1}\text{-th input channel,} 1 \leq c_{i-1} \leq C_{i-1};\\
    k^j(i,c_{i-1},c_i) &\in R^{w\times h} \text{ is a convolution kernel at the i-th layer between two channels.}
\end{align*}

Denote $k^j(t)$ as one instance of the smallest sharing unit of the $j$-th client at the $t$-th iteration.
This unit could be a kernel, a channel, a layer, or the whole model.
We also use $\Neib_j^t = \{j_1^t,...,j_R^t\}$ to denote the index of clients that communicate with $j$ at $t$.
Then, the algorithm proposed by \cite{hu2019decentralized} can be generalized as:
\begin{align}
k^j(t+1) = \frac{|D^{(j)}|}{|D(j,t)|} +  \sum_{m=1}^{|\Neib_j^t|}\dfrac{ |D^{(j_m^t)}| k^{j_m^t}(t)}{|D(j,t)|} \ \ \text{ , where } |D(j,t)| = |D^{(j)}|+\sum_{m=1}^R|D^{(j_m^t)}|
\end{align}

\subsection{Connected graphs and small-world phenomenon} \label{smallworld}
We use a graph $G=(V,E_G)$ to represent the communication topology between clients, where $V = \{1,...,K\}$. 
If client $i$ would receive a message containing the model parameters from client $j$, then a corresponding edge $(j,i)\in E_G$ represents this connection. 
The consensus algorithm in DFL updates the model weights using the following expression: 
\begin{align}
 W^i \leftarrow \frac{1}{|\mathcal{N}_i|}\sum_{j\in\mathcal{N}_i}(W^j-W^i), \ \ \mathcal{N}_i \triangleq \{j : (j,i) \in E_G\}   
\end{align}
In other words, each client updates its model by adding the average difference from its neighbors in the topology $G$. 
This procedure is guaranteed to converge to the centroid of all clients' models, as long as the undirected topology graph $G$ has only one component (or, equivalently, contains a spanning tree, or exists at least one path between each pair of clients). %fully connected. 
In addition, the rate of convergence is determined by the second eigenvalue $\lambda_{2,G} \geq 0$ of $G$'s Laplacian $L_G$, which is the non-zero eigenvalue with the smallest magnitude for a connected graph.
When using the consensus algorithm, there is a trade-off between having a dense and a sparse network. It is well-known that the rate of convergence is dominated by this eigenvalue. 
In general, a denser network has larger $|\lambda_{2,G}|$, and has faster convergence rate.
On the other hand, however, a sparser network could require less communication between clients at each iteration.
Thus, we observe a trade-off between less number of iterations and smaller communication cost per iteration.

\begin{figure}[!htbp]
    \centering
    \subfigure[Cycle-like graph, where each node has $k=4$ local connections. $\lambda_{2} = -0.738$]{\includegraphics[width=0.32\textwidth]{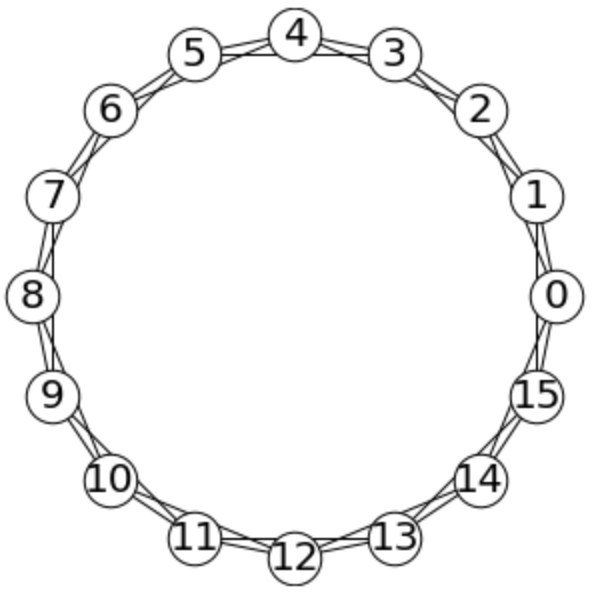}}
    \subfigure[Small-World graph, where edges from (a) were rewired with $p=0.5$. $\lambda_{2} = -1.071$]{\includegraphics[width=0.32\textwidth]{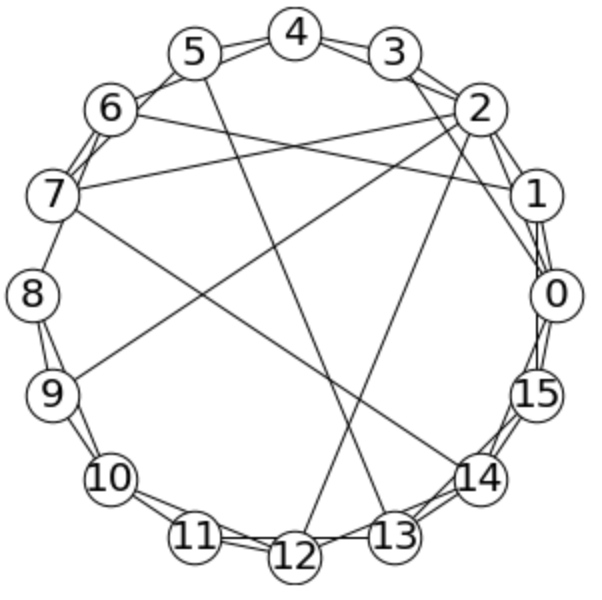}}
    \subfigure[Idealistic graph with $k=4$. $\lambda_{2} = -2.0$]{\includegraphics[width=0.32\textwidth]{MARL_pic/SW01.png}}
    \caption{Comparison of local topologies with Small-World topologies.}
    \label{fig:SWexample}
\end{figure}

Notice that the \texttt{FedAvg} algorithm is not easily generalized to the consensus algorithm. 
In the consensus algorithm, all clients update simultaneously in one iteration. 
In \texttt{FedAvg}, however, the aggregation is done in 2 iterations:
Only the central client updates its model in the first iteration, using the average of all its neighbors \textit{but} itself;
then, the other clients get updated in the second iteration by directly replacing its model with the central client's model. 
It achieves the same outcome as the consensus step in DFL, but the mechanism has a subtle difference.

Over two decades ago, \cite{Watts1998} revealed the Small-World phenomenon in their seminal paper. 
The observation is that, starting from a circular graph topology where each node only connects to the few closest nodes ("local" connections), the eigenvalue $\lambda_2$ could be drastically increased by randomly rewiring some of those local connections to long-distance pairs. 
With enough rewiring, the convergence rate of the consensus algorithm could be increased without additional communication costs.
For example, as shown in Figure \ref{fig:SWexample}, performing rewiring on the graph from \ref{fig:SWexample}(a) resulted in \ref{fig:SWexample}(b), which has the same number of edges but significantly higher $\lambda_2$ magnitude.
\ref{fig:SWexample}(c) offers an example where the $\lambda_2$ could become even larger. However, in this topology, the clients would have to form connections with ones further away, instead of with local neighbors.
In addition, the graph connections follow a delicate cyclic pattern, which is not practical.

Thus, the Small-World rewiring scheme allows us to maintain a sparse, realistic, yet effective topology. In the next section, we attempt to combine the segmented federated learning approach with the small-world topology to further reduce the communication cost.

\subsection{Simulation Results of Segmented Federated Learning with Small-World Topology}
In this segment, we present the experimental results from combining the segmented federated learning approach with Small-World communication topologies.
%of combining the two sections above. 
Out of the possible smallest units, we would stick to using output channels ($W^j_{L_{i,out},c_i}$), because:
\begin{itemize}
    \item Using layers ($W^j_{L_i}$) should be equivalent to the existing work by \cite{hu2019decentralized};
    \item Using kernels ($k^j(i,c_{i-1},c_i)$) proves to be taking too long in the simulation, and the result does not show successful training.
\end{itemize}
\subsubsection{Segment Sharing with Clientwise Topology}
\begin{figure}[!htbp]
    \centering
    \subfigure[Epoch counts]{\includegraphics[width=0.58\textwidth]{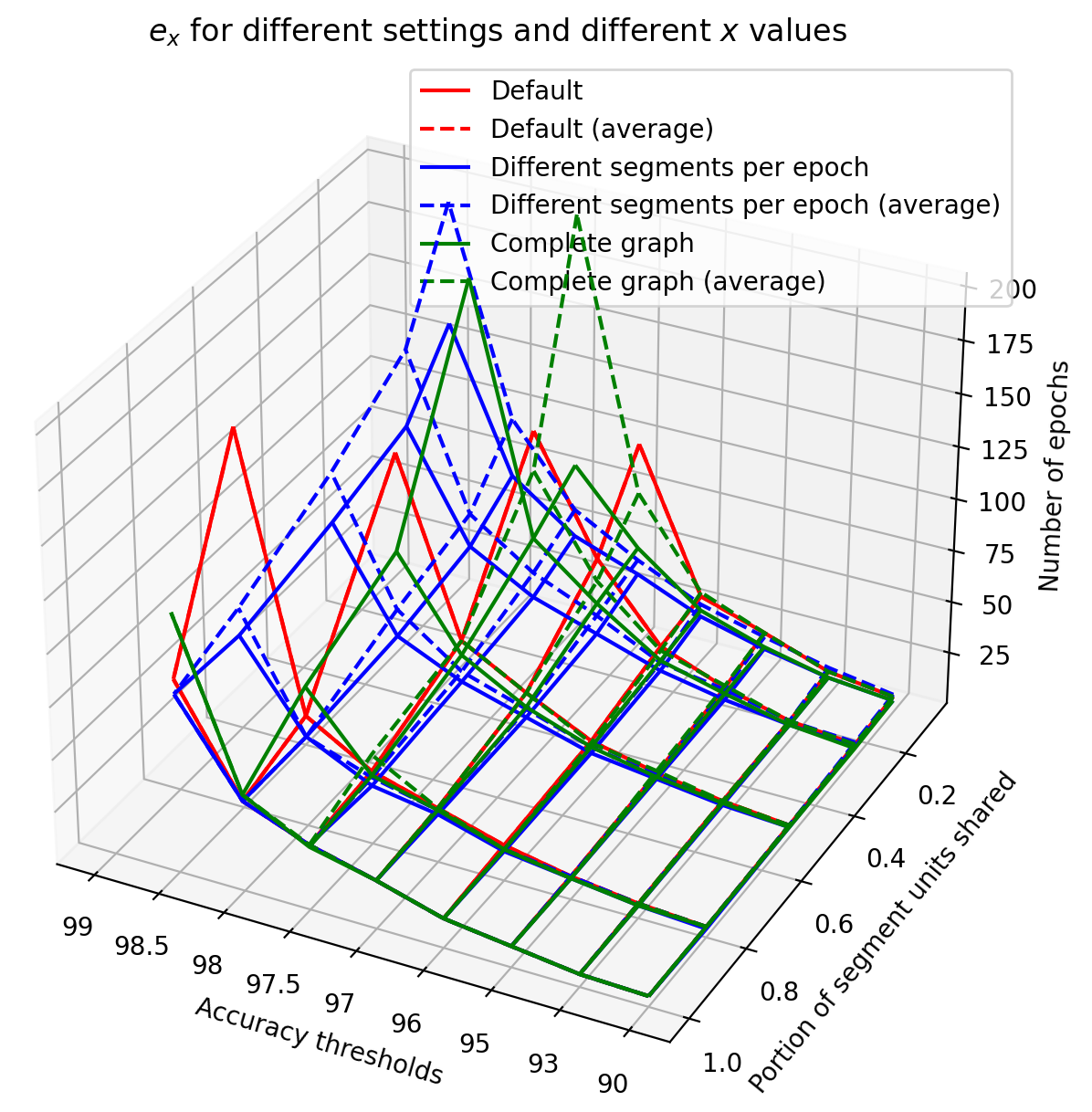}}
    \subfigure[Epoch counts, another angle]{\includegraphics[width=0.55\textwidth]{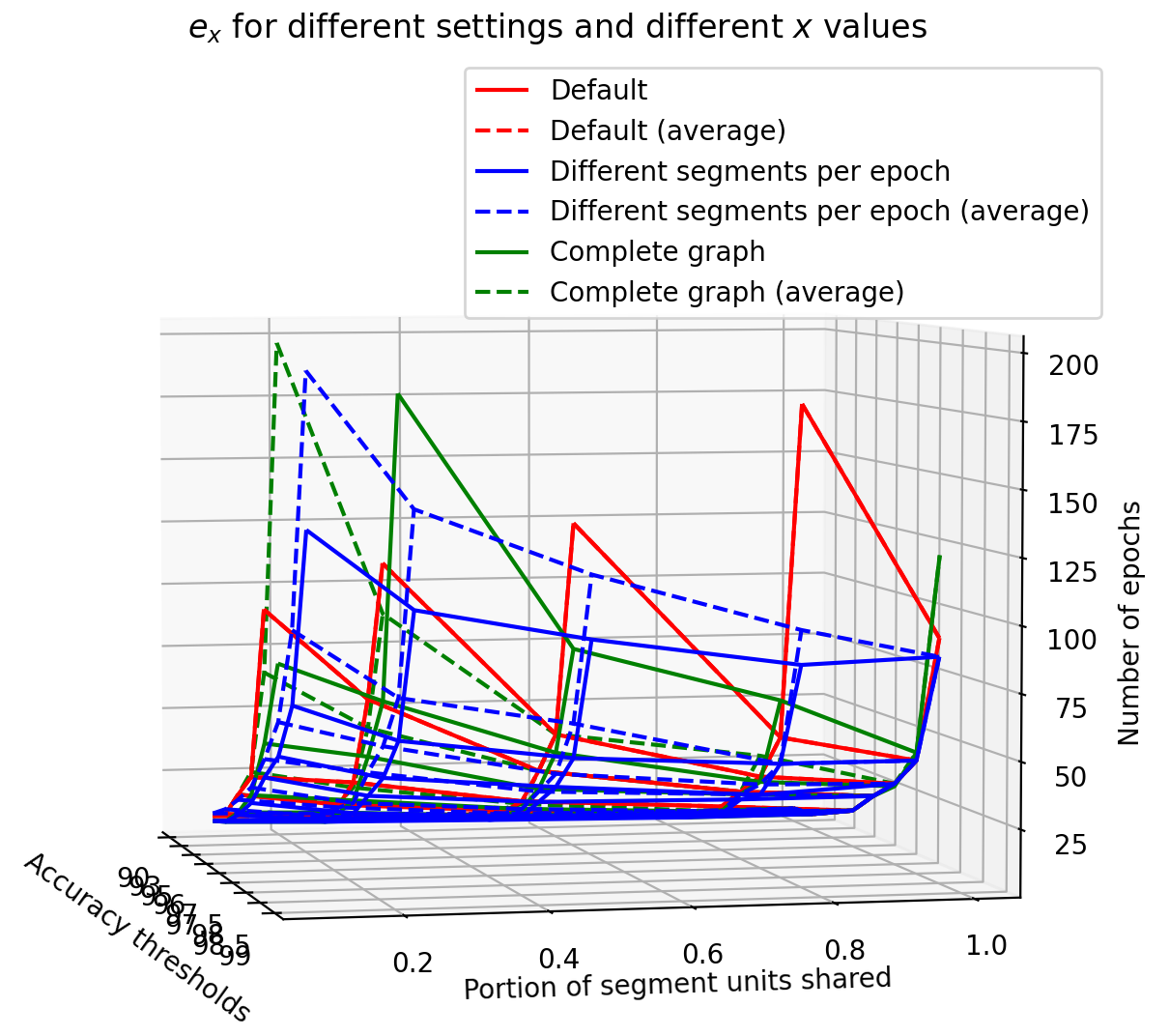}}
    \subfigure[Max accuracy]{\includegraphics[width=0.42\textwidth]{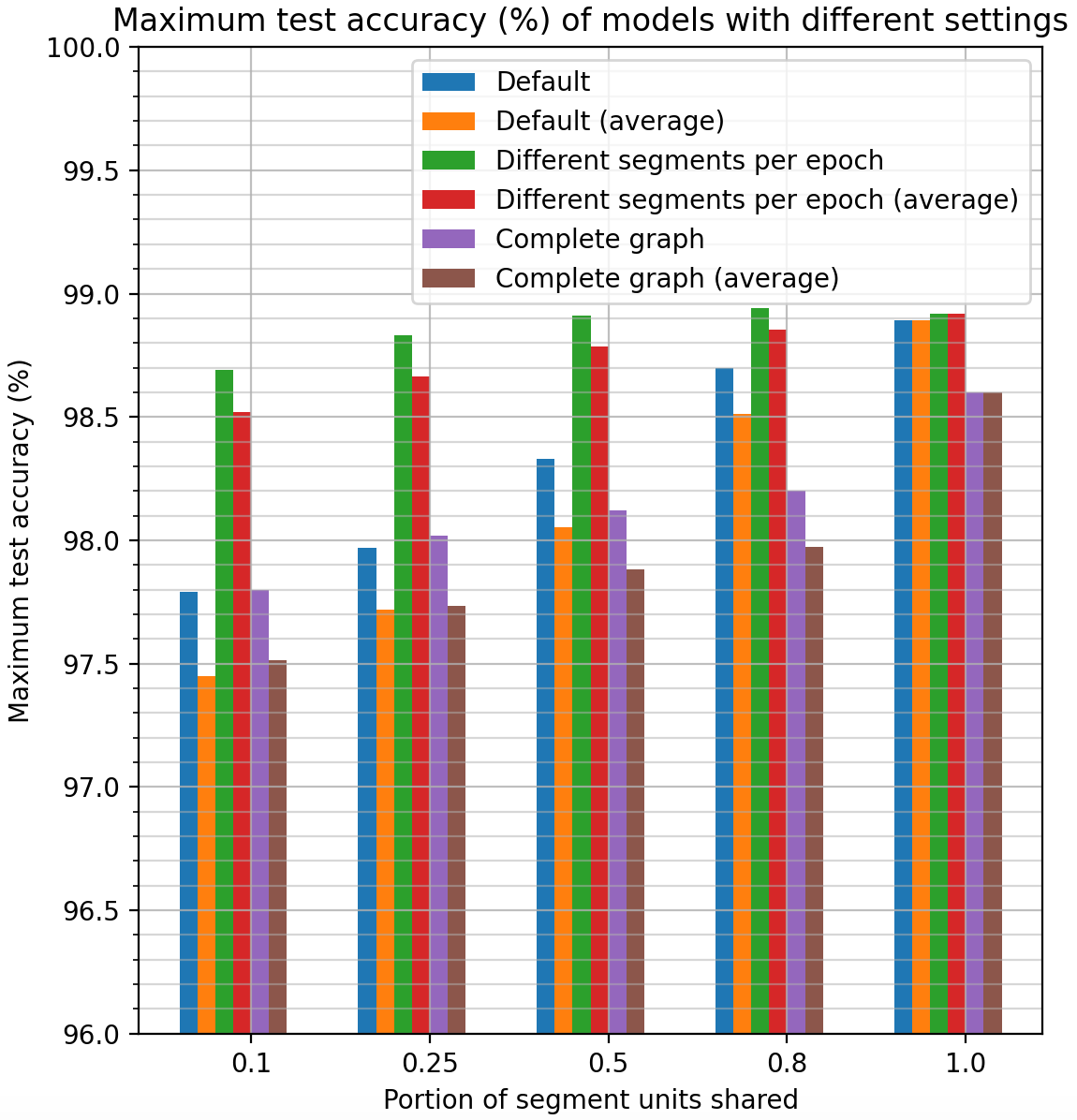}}
    \caption{(a-b). Number of epochs to reach $e_{90}$ to $e_{98.5}$, and (c). Max accuracies, for the top-behaving model and the model averages, using different aggregation settings.}
    \label{fig:cuPre}
\end{figure}

In this section, we assume that the communication graph is undirected. 
All segment sharing are performed under the same Small-World client communication topology as Figure \ref{fig:SWexample} shows, which is generated with $k=2$ as the original number of neighbors per client, and $p=0.5$ as the rewiring probability for each edge. 
Further, we define $PSS$ as the portion of segment units shared per client.
For example, if $PSS=0.5$, then each client $j$ would share half of all its segments $W^j_{L_{i,out},c_i}$ at each epoch. 
$PSS$ is one of the parameters that are compared in Figure \ref{fig:cuPre}.

To aid further discussion, we also denote the set of indices of the chosen segments as $SegInd^j$.
Each item inside $SegInd^j$ is a tuple $(i,c_i)$, where $i$ indicates the layer index of the segment, and $c_i$ is the index of the segment in layer $L_i$. 
Note that the segments are not picked layer by layer. Segments in all layers are considered together in this paper:
\begin{align}
\forall i,j, \,\, \dfrac{|\{(i,c) \in SegInd^j\}|}{C_i} \neq PSS = \dfrac{|SegInd^j|}{\sum_{i=1}^l C_i}    
\end{align}

Figure \ref{fig:cuPre} compares the performances of three different settings. They all simulate segment-wise aggregations, albeit in slightly different manners. The settings' descriptions are listed below: 
\begin{enumerate}
    \item \textbf{"Default"} setting pre-determines the segment indices $SegInd^j$ that each client $j$ would poll (receive) from others. 
    Here, we do not necessarily guarantee that $SegInd^{j_1} = SegInd^{j_2}$ for any pair of clients $(j_1,j_2)$. 
    The set of clients that $j$ would poll from is exactly $\Neib_j$, the set of neighbors of $j$ in the client-wise small-world communication topology $G$. 
    \item \textbf{"Different segments per epoch"} setting means that $SegInd^j_t$ for client $j$ is dependent on the epoch number $t$:
    At each new epoch, the segment indices is randomly selected again. 
    Everything else is the same as "Default".
    \item \textbf{"Complete graph"} uses a complete graph as the topology, which means $\Neib_j = \{k: 1\leq k \leq K, k \neq j\}$. The rest are the same as "Default".
\end{enumerate}
All the experiments are run with the default hyper-parameters: $E=0.05, S=0, U=0$, and trained with 200 epochs. 
They are run with 5 values of $PSS = \{0.1,0.25,0.5,0.8,1.0\}$. 
Figure \ref{fig:cuPre}(a-b) display the epoch numbers of different accuracy thresholds from $e_{90}$ to $e_{98.5}$ for each $PSS$ value.
Note that, if an accuracy threshold is never reached, then the corresponding value is not plotted.
For example, the "Default" at $PSS=0.1$ never reaches an accuracy over 97.5\% within 200 epochs, so $e_{98}, e_{98.5}$, and $e_{99}$ are missing from the plot.
The maximum epoch number is at $e_{97}=100$. 
Thus, even though it appears that the "Different segments" setting has the highest epoch numbers at $PSS=0.1$, those correspond to accuracy thresholds that the other two settings never achieved within 200 epochs. 
The correct interpretation is not that "Different segments" setting takes the longest to reach accuracy thresholds, but that "Different segments" setting is the only one to reach those accuracy thresholds, and the other settings either could not reach them, or would have required much longer training.
This is validated by Figure \ref{fig:cuPre}(c) as well.

From Figure \ref{fig:cuPre}, we observe that models with larger $PSS$ values reached accuracy thresholds faster, and could reach a higher accuracy thresholds.
This observation is true for all settings.
Larger $PSS$ means that more segments of the model would be at the same weights before each epoch starts, and the $PSS=1$ case is exactly equivalent to the training setup in previous sections without segmenting. 

Generally speaking, the performance of the "Different segments" setting is better than the "Default" setting. 
When clients poll different segments at each epoch, each part of the model would have a same chance of being aggregated in the long run. 
This helps make sure no segment of the client deviates from the global consensus too far.
% Also, from the results, the "Complete graph" setting had surprisingly worse accuracy than both. 

Interestingly, the three settings end with different performances when $PSS=1$.
Theoretically, the full model weights are shared when $PSS=1$, so the algorithm should be equivalent to $\textit{FedAvg}$ and to DFL.
One possibility is that procedures that required randomness cause the discrepancies.
Although each experiment shares the same random seed, different settings follow different procedures. For example, "Different Segments" setting would generate a random topology at each epoch, but "Default" setting would not have to.
As a result, the model training procedure is not exactly the same between settings, even when starting with the same seeds.

From the performances above, we decide to only focus on "Default" and "Different segments per epoch" settings in the following section. They are different enough in implementation, and their performances are good enough.
In addition, we limit the number of iterations per aggregation into $s = \min(20, s_{\ep})$ instead of $s = \min(100, s_{\ep})$ for the rest of the section. The reason is that 20 iterations has been enough for converging models to reach consensus.

\subsubsection{Effect of Aggregation Frequency on Segmented Federated Learning}
\begin{figure}[!htbp]
    \centering
    \subfigure[$E=0.05,0.1$]{\includegraphics[width=0.47\textwidth]{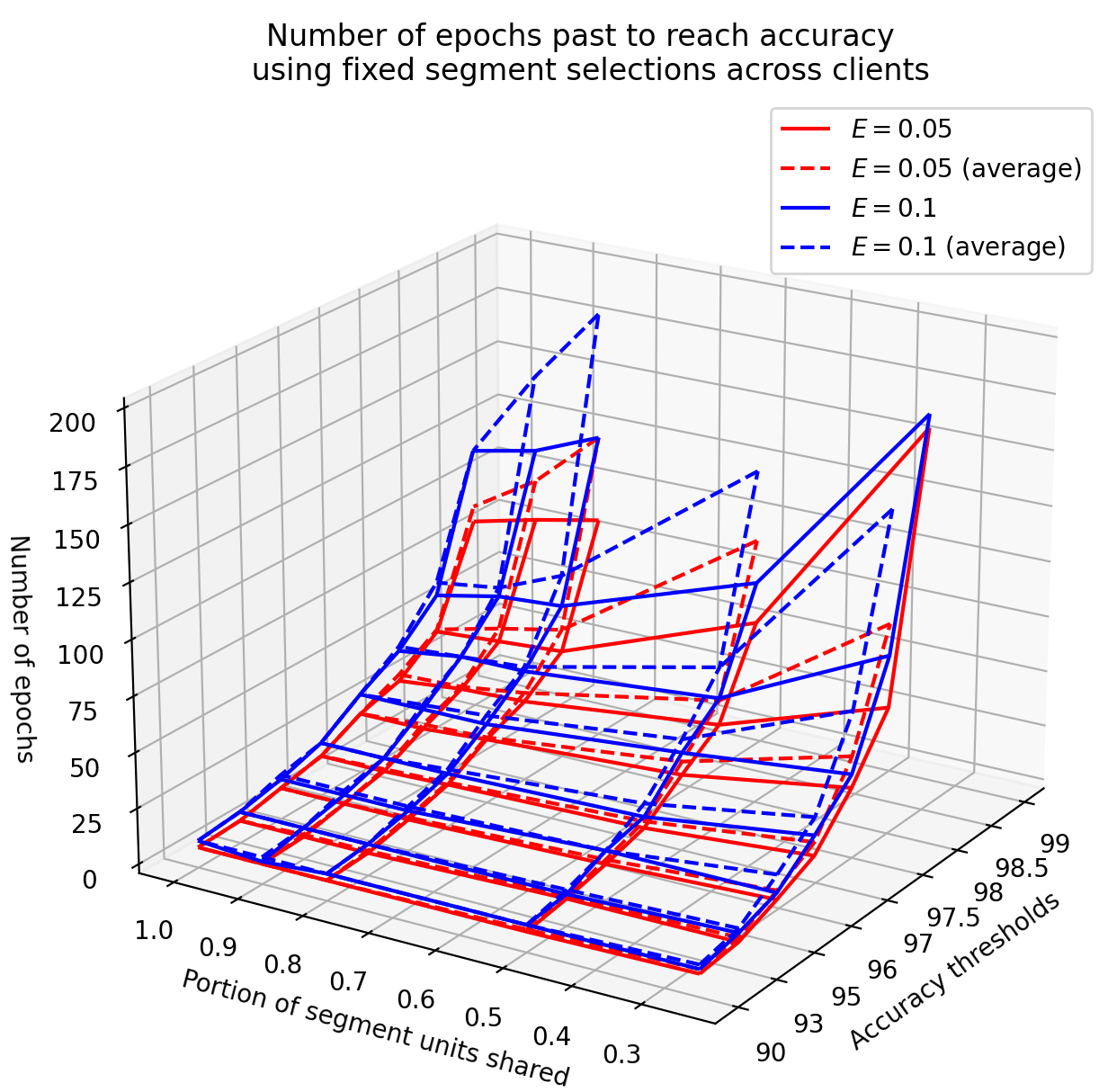}}
    \subfigure[$E=0.25,0.5$]{\includegraphics[width=0.47\textwidth]{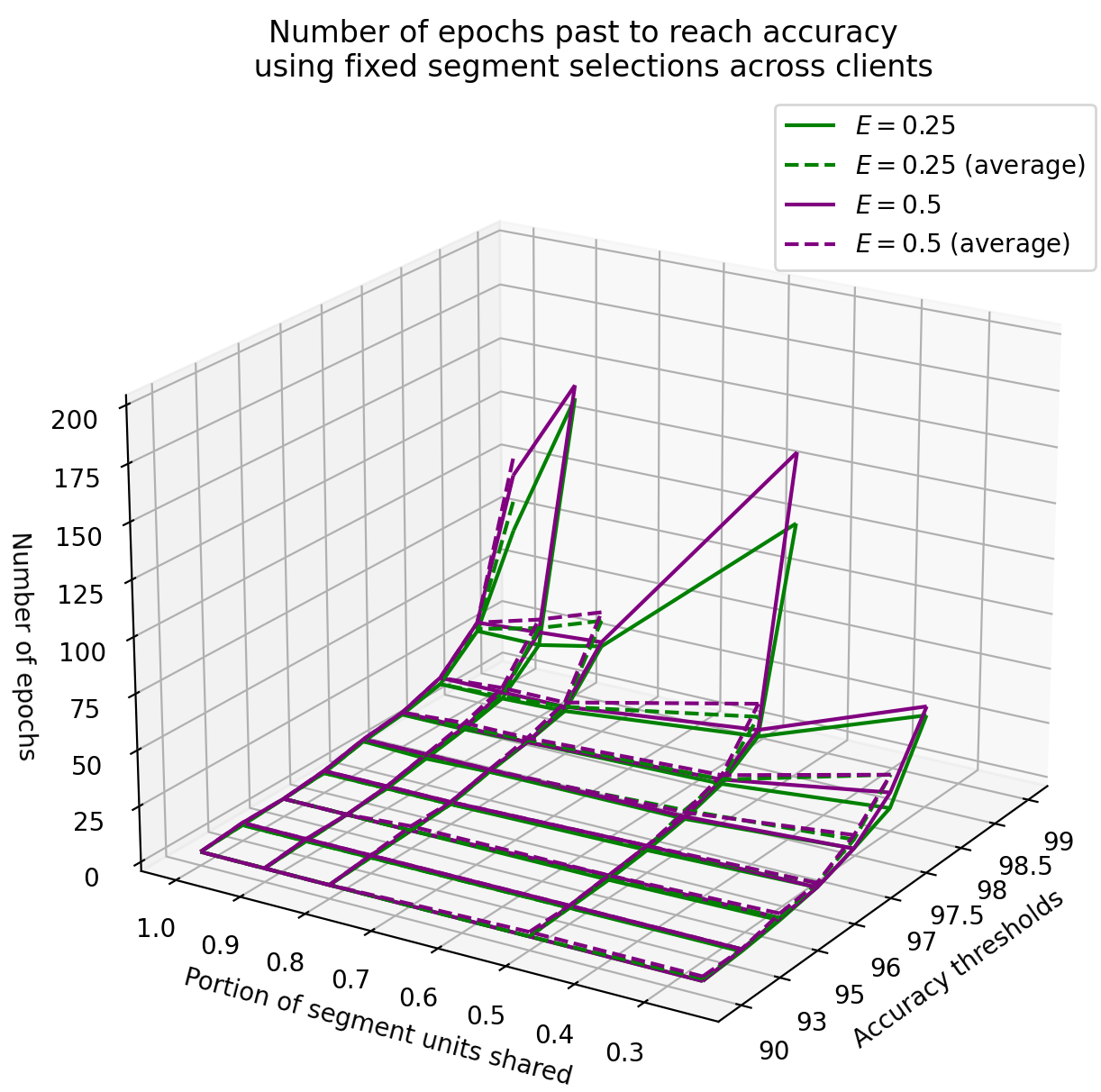}}
    \subfigure[max accuracy]{\includegraphics[width=0.47\textwidth]{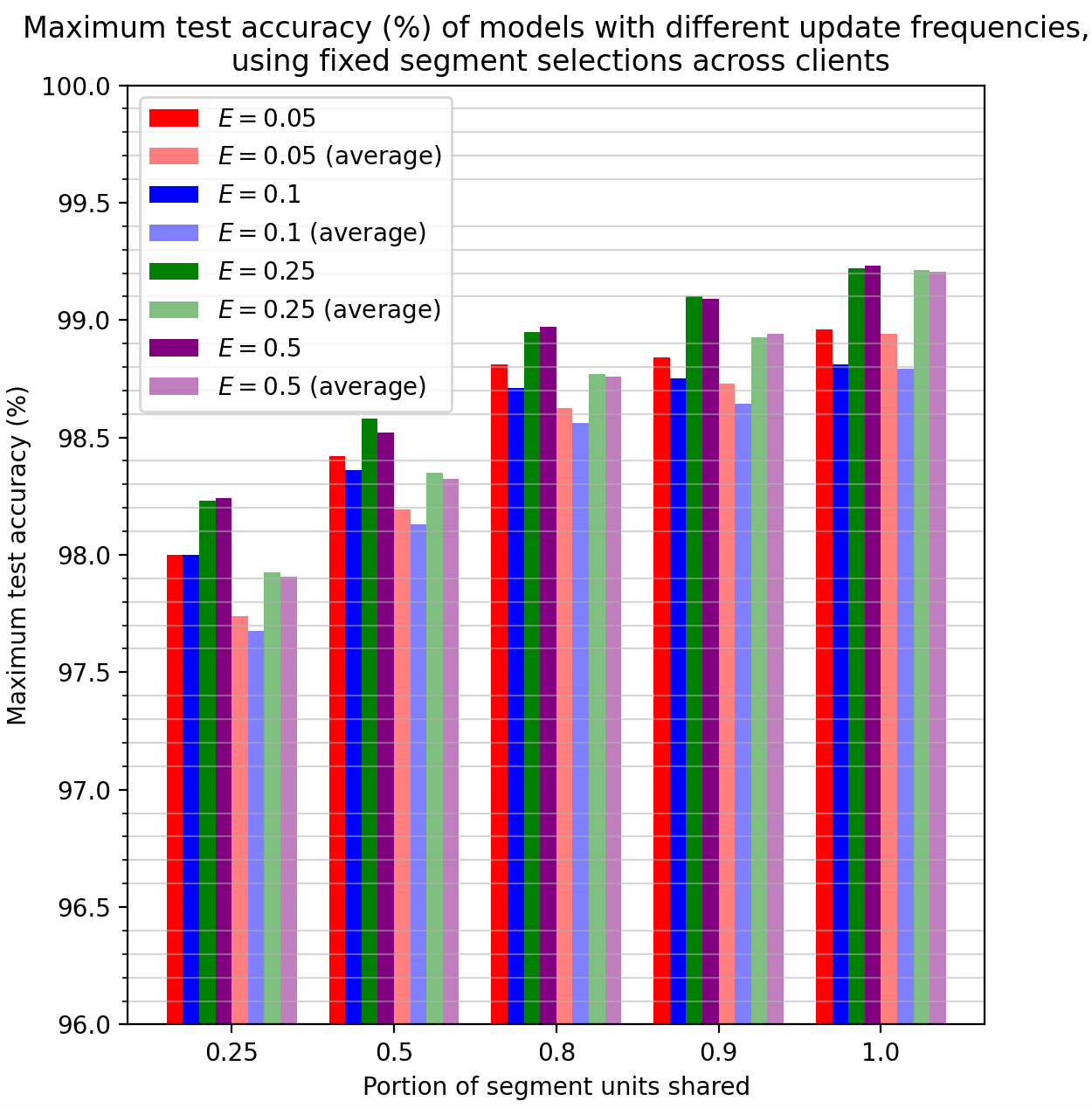}}
    \subfigure[model distance range]{\includegraphics[width=0.37\textwidth]{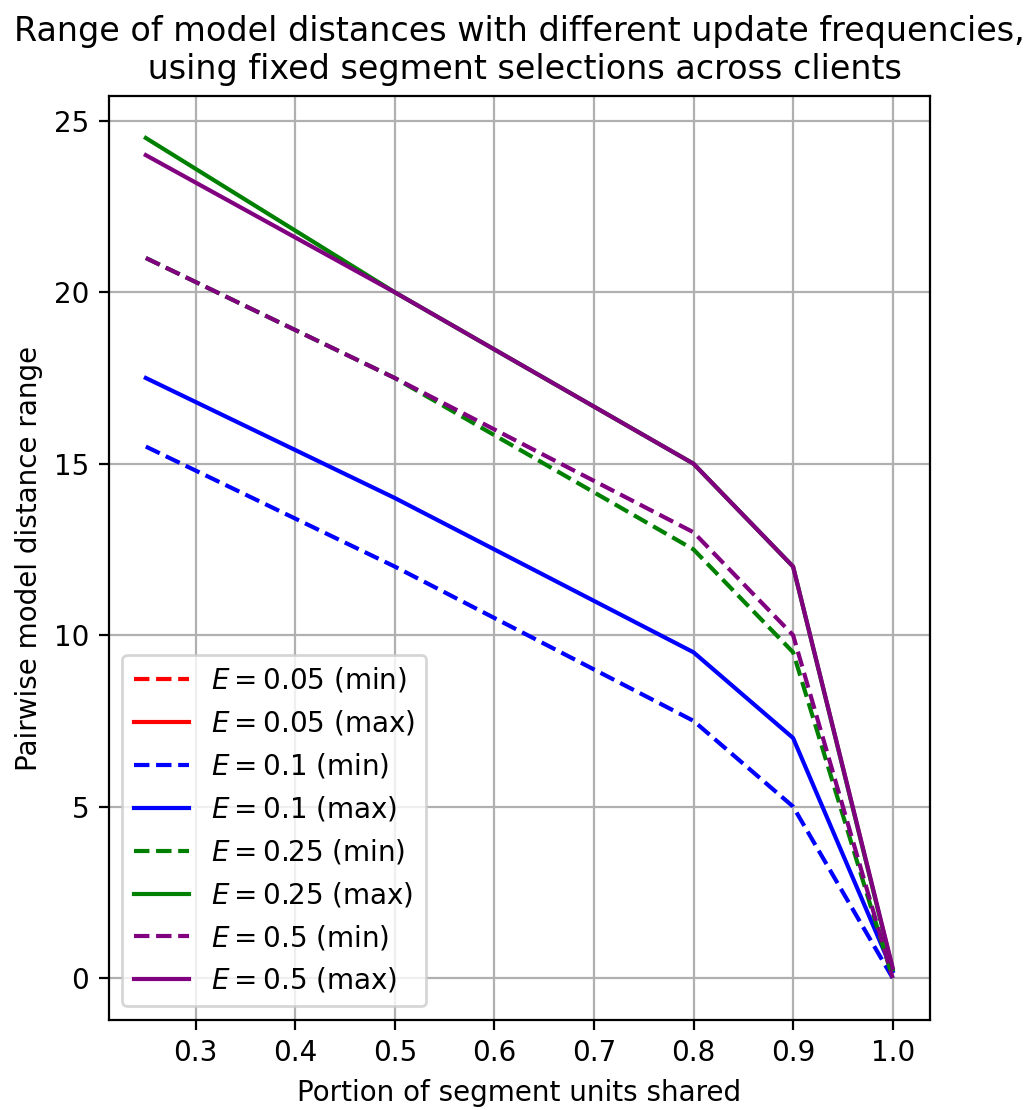}}
    \caption{(a-b) Number of epochs to reach $e_{90}-e_{98.5}$ for the top-behaving model and the model averages, (c) max accuracies, and (d) range for pairwise model distances in $\ell_2$ norms, compared when using different $E$ when clients always share the same set of segments. Values in (a-b) for $E=0.1$ was multiplied by 2 to compare with $E=0.05$; the same goes for $E=0.5$ for $E=0.25$. Data for $E=0.05$ in (d) is missing.}
    \label{fig:cuno}
\end{figure}

\begin{figure}[!htbp]
    \centering
    \subfigure[]{\includegraphics[width=0.47\textwidth]{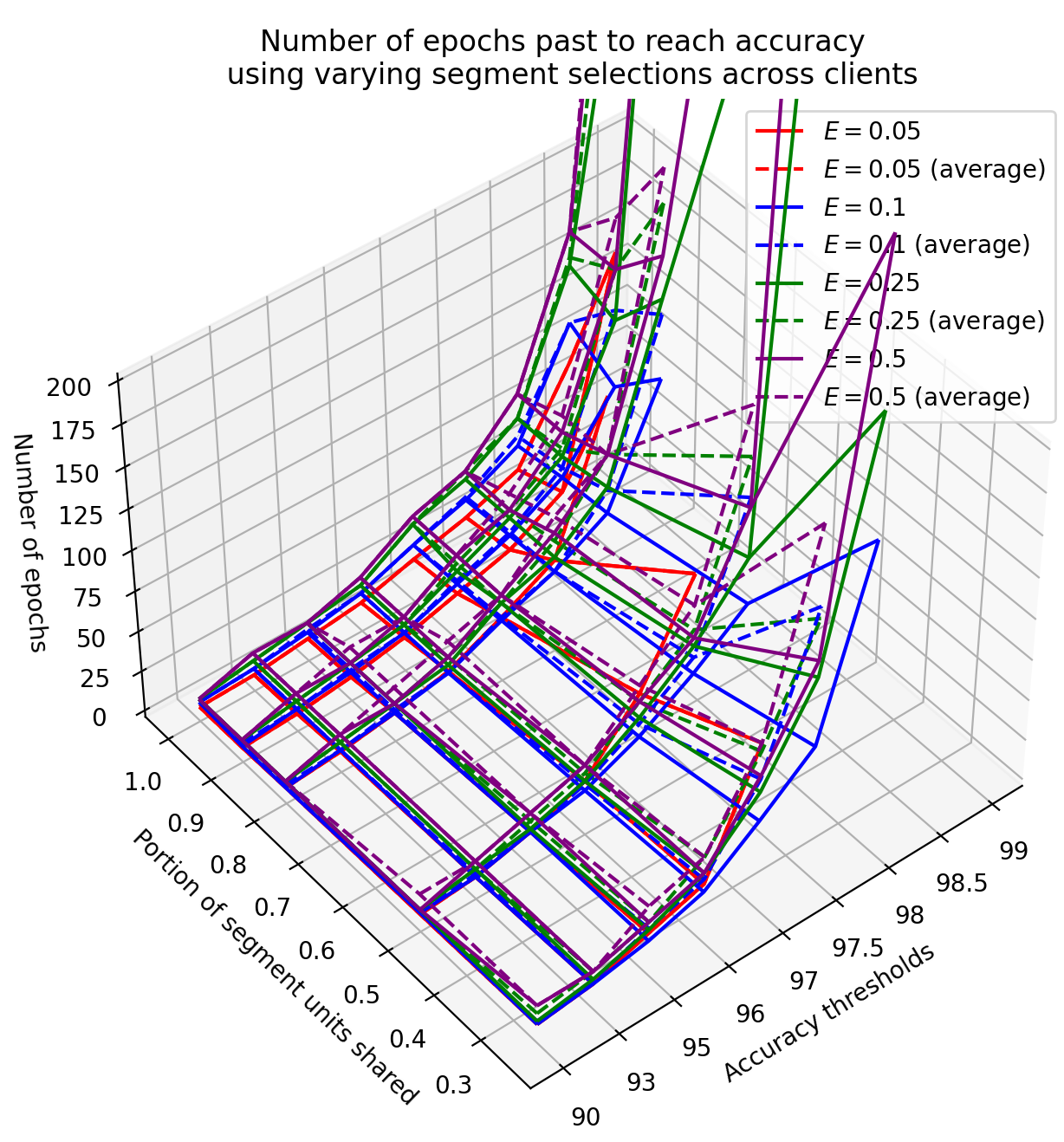}}
    \subfigure[another angle]{\includegraphics[width=0.47\textwidth]{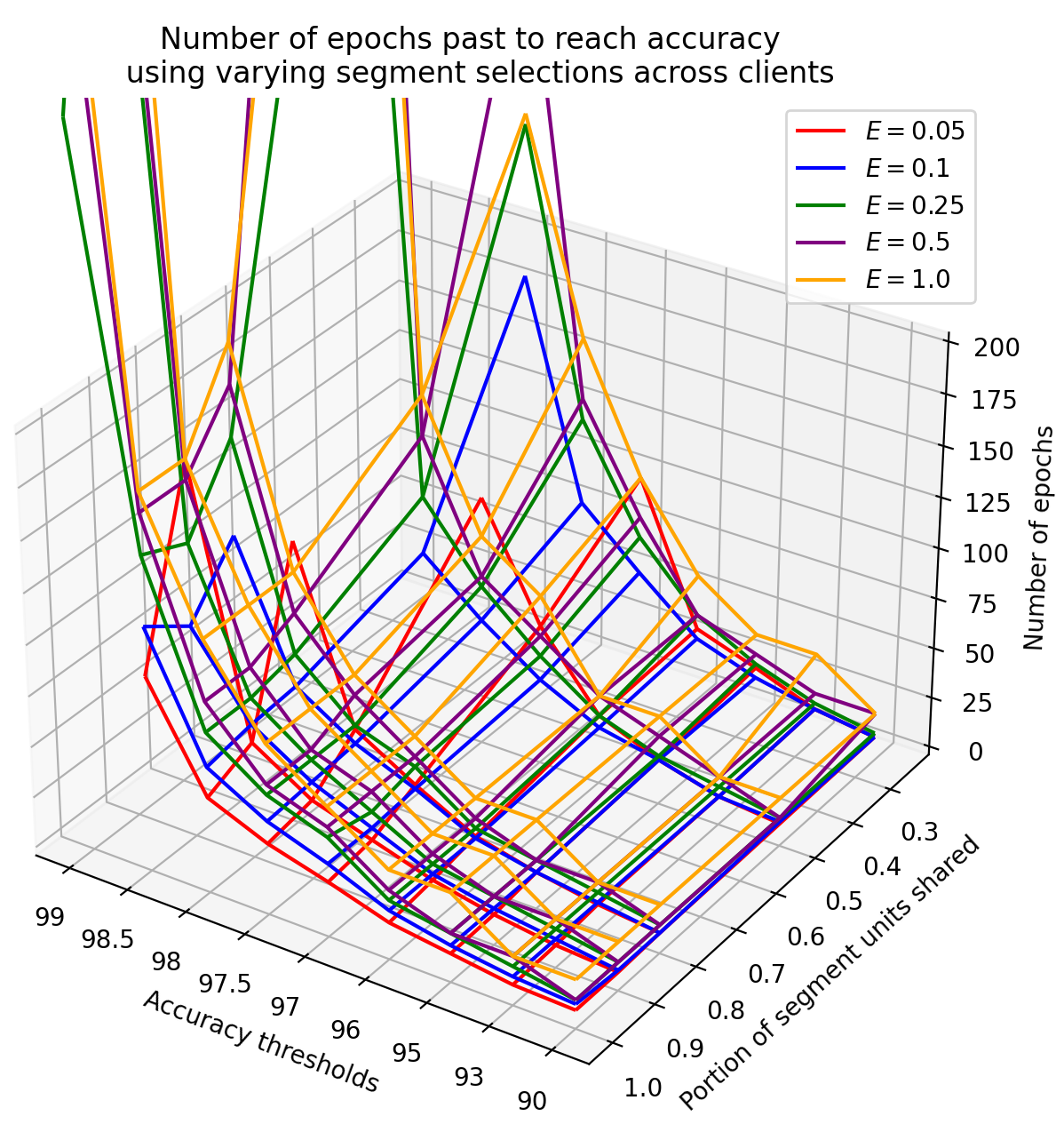}}
    \subfigure[max accuracy]{\includegraphics[width=0.47\textwidth]{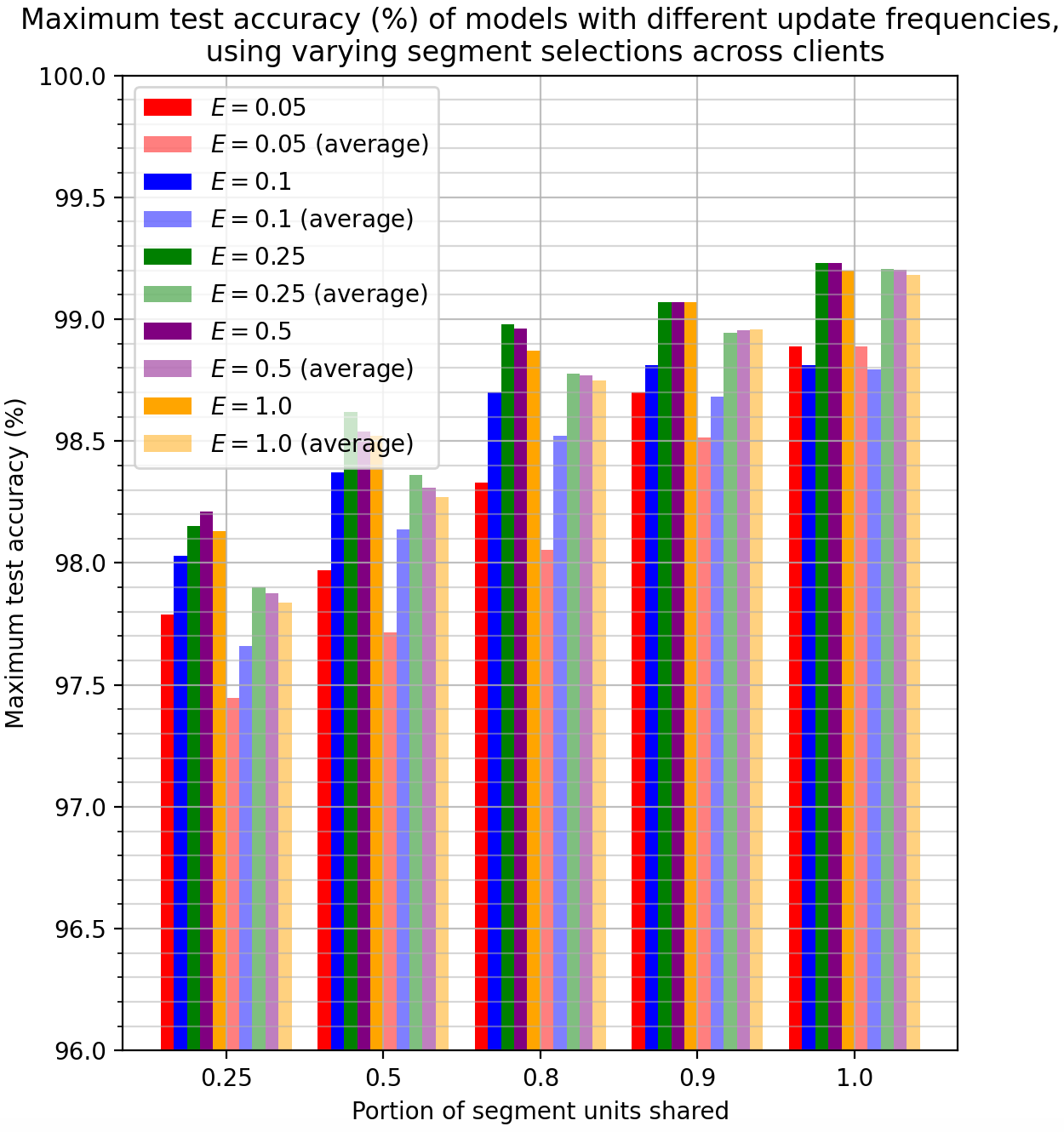}}
    \subfigure[model distance range]{\includegraphics[width=0.37\textwidth]{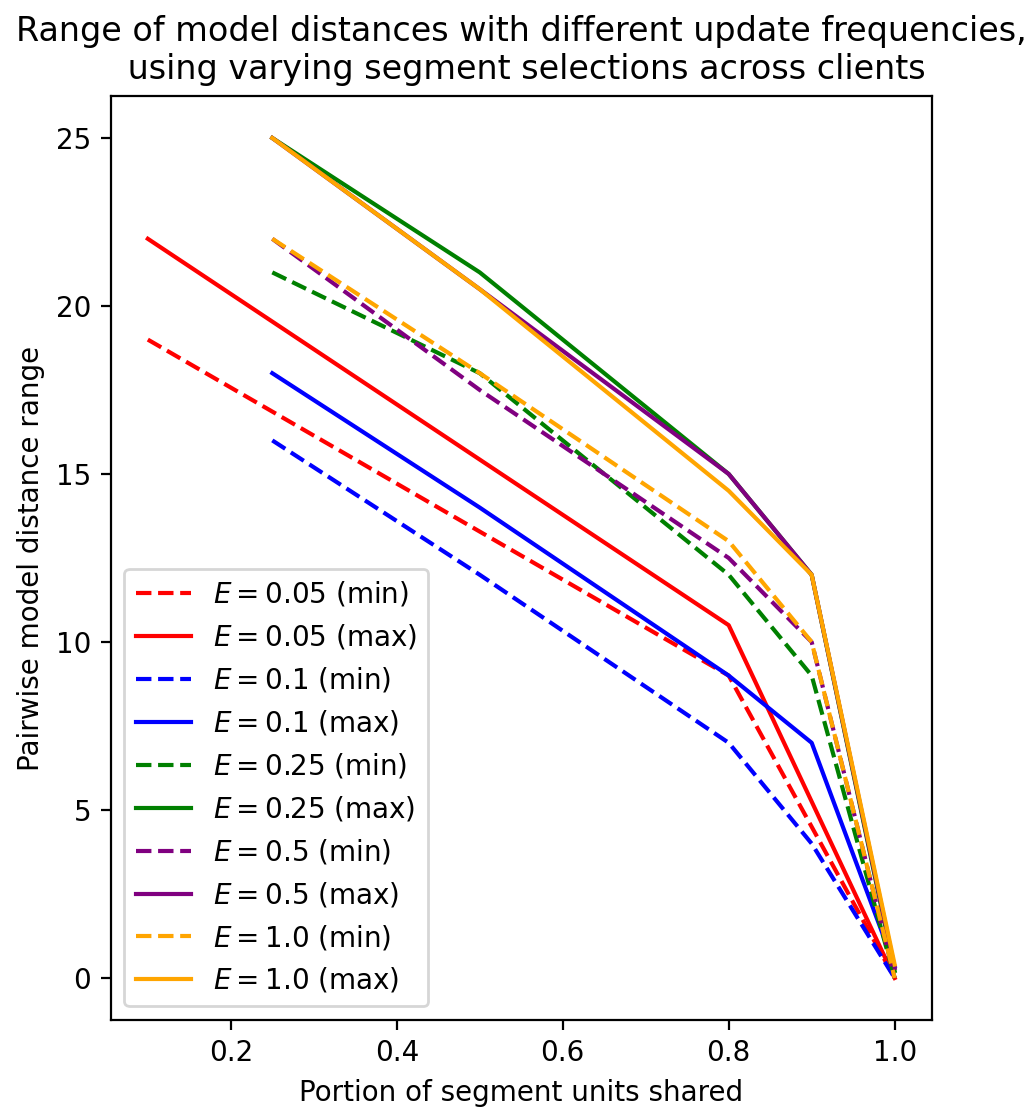}}
    \caption{(a-b) Number of epochs to reach $e_{90}-e_{98.5}$ for the top-behaving model and the model averages, (c) max accuracies, and (d) range for pairwise model distances in $\ell_2$ norms, compared when using different $E$ when each client sends different set of segments. Values in (a-b) for $E=0.1$ is multiplied by 2 to compare with $E=0.05$; the same goes for $E=0.5$ for $E=0.25$.}
    \label{fig:cuw}
\end{figure}

We also investigate the effect of aggregation frequency in the segmented setting. 
The results are shown in Figure \ref{fig:cuno} and \ref{fig:cuw}, where both figures follow the "Default" setting, but all clients pull the same set of segments in Figure \ref{fig:cuno}. 
Both figures display the epoch numbers in (a-b), the accuracy by the top-performing clients in (c), and the range of whole-model distances in (d).
$E$ values are chosen from $0.05,0.1,0.25,0.5,1.0$.
Models with $E=0.05$ and 0.25 are trained with 200 epochs, while models with $E=0.1$ and 0.5 with 100 epochs, so that they form comparable pairs with similar amount of training:
$$0.05 \times 200 = 0.1 \times 100; \ \ 0.25 \times 200 = 0.5 \times 100$$
Note that the $PSS$ values in Figure \ref{fig:cuno} and \ref{fig:cuw} are switched to $\{0.25,0.5,0.8,0.9,1.0\}$, where  $0.1$ is replaced with $0.9$ for finer observation near where behavior changes.

In Figure \ref{fig:cuno}, the shared segments are the same for each client at each epoch: 
$SegInd^{j_1}_{t_1} = SegInd_{t_2}^{j_1} = SegInd_{t_2}^{j_2}, \ \forall j_1,j_2,t_1,t_2$.
In this figure, (a) compares the amount of epochs required for models to reach certain accuracies with $E=0.05$ and $E=0.1$. Because the models with $E=0.1$ has doubled number of samples and optimization steps than models with $E=0.05$ does, the values for $E=0.1$ are multiplied by 2 to better reflect the difference between the two cases. The same is done on values for $E=0.5$ in (b) to compare with $E=0.25$.
In both subfigures, experiments with higher $E$ would take a larger number of effective epochs to reach the same accuracy threshold. For example,
$0.05e_{97}|_{E=0.05} > 0.1e_{97}|_{E=0.1}$.
However, the top accuracy is generally higher with higher $E$ values (less frequent aggregations). This is consistent with the general observation that training is more effective with less frequent aggregation. 


The experiments in Figure \ref{fig:cuw}(a-b) follow the "Default" setting, which means, $SegInd_{t_1}^{j_1} = SegInd_{t_2}^{j_1}, \;\; \forall t_1,t_2,j_1$, but $SegInd^{j_1} \neq SegInd^{j_2}$.
We observe similar behaviors as in \ref{fig:cuno}; the only difference is that the $E=0.05$ case turns out to take longer than the others to reach certain accuracies when $PSS$ is low. 

Additionally, both Figure \ref{fig:cuno}(d) and Figure \ref{fig:cuw}(d) depict the range of possible pairwise model distances $\norm{W^{j_1}-W^{j_2}}$ at each epoch $t$.
Both subfigures indicate that pairwise model distances never reach 0, until $PSS$ reaches the special case of 1 where the entire model is shared between clients.
The range of model distances monotonically shrinks as $PSS$ increases, because more parameters are shared and synchronized at the start of each epoch. 
In addition, less frequent aggregations generally result in larger model distances.

\begin{figure}[!htbp]
    \centering
    \includegraphics[width=0.9\textwidth]{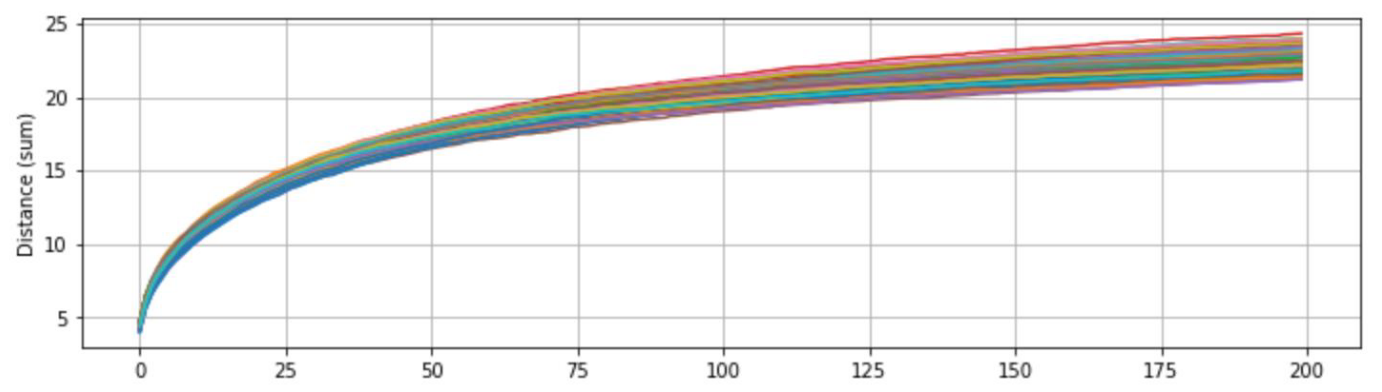}
    \caption{Pairwise model $\ell_2$ distance range at each epoch, obtained from the $PSS=0.25, E=0.25$ experiment with different segment sets polled by each client.}
    \label{fig:cunoDistExample}
\end{figure}
Figure \ref{fig:cunoDistExample} is an additional plot showing the pairwise model distance history for  a specific case where $PSS=0.25$ and $E=0.25$. 
The pairwise distances are increasing over time, even though the models are trained and aggregated normally. 
This is observed in all experiments shown in Figure \ref{fig:cuno} and \ref{fig:cuw} where $PSS < 1$.
It is expected to see nonzero model distances from Figure \ref{fig:cuno}(d). 
Without aggregation, any segment $(i,c_i) \notin SegInd^j$ would not be aggregated, and would not become uniform across clients. 
In fact, different segment weights could demonstrate similar behaviors, as observed by \cite{matchedFL}. 

On the other hand, the nonzero distances in Figure \ref{fig:cuw}(d) could be explained differently, while it should be expected that all the segments are polled by at least some client, i.e.,
\begin{align}
  \bigcup_{j=1,...,K} SegInd^j = \bigcup_{i=1,...,l} \{(i,c_i): 1\leq c_i\leq C_i\}  
\end{align}
Those clients that share one segment only constitute a subset of the client-wise Small-World topology.
The aggregation topology for each segment could contain more than 1 components, where different components never aggregate with clients from each other.
This, again, explains why some, if not most, segments would not contain the same weights at each epoch.
Again, it is known that models could give similar behaviors without the same set of weights.
Hence, we again see model distances increasing over time due to segment differences over individual training without aggregation, while the models are still giving similar performances. 
\begin{figure}[!htbp]
    \centering
    \subfigure[Distance for each segment at each aggregation iteration at the last epoch ]{\includegraphics[width=0.87\textwidth]{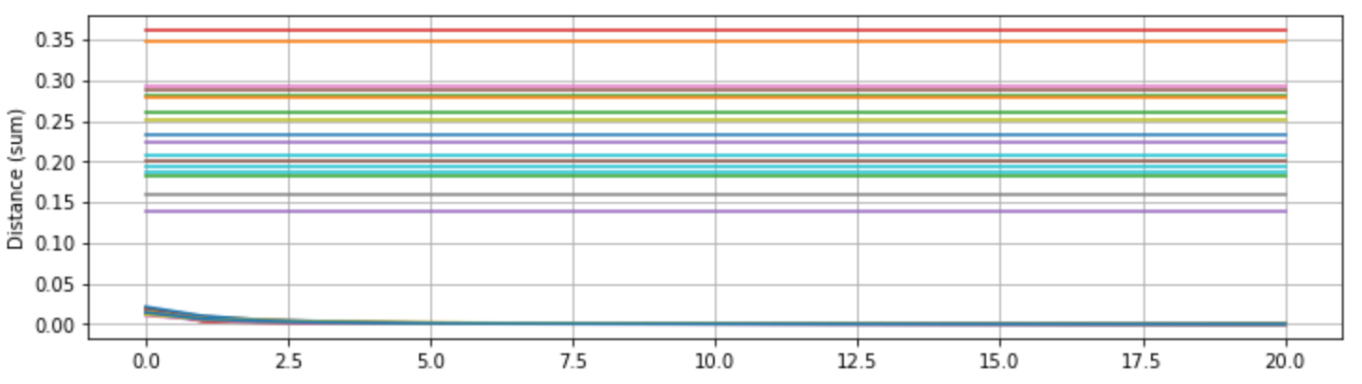}}
    \subfigure[Distance for each segment at the start of aggregation at each epoch]{\includegraphics[width=0.87\textwidth]{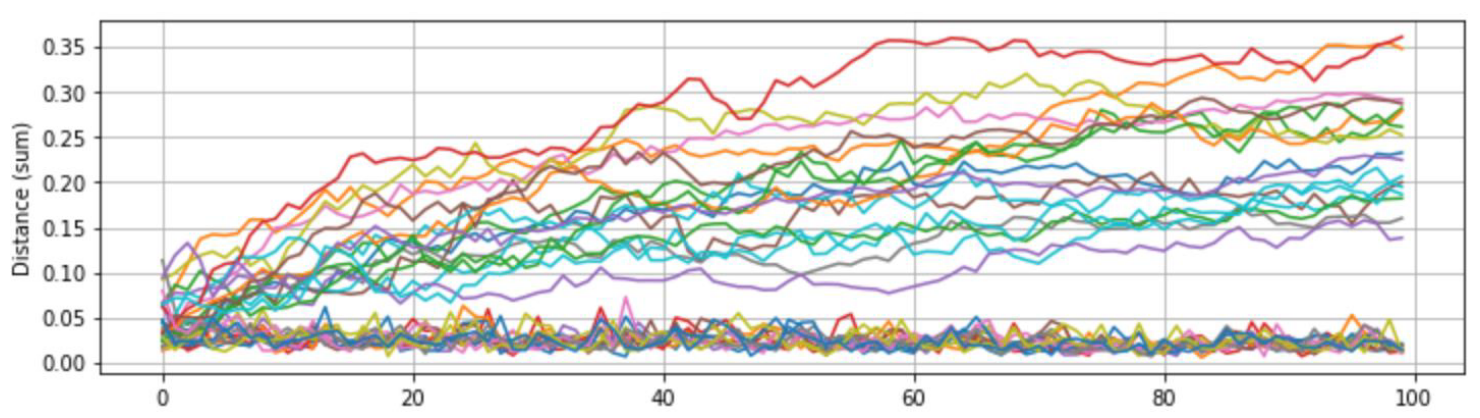}}
    \subfigure[Distance for each segment at the end of aggregation at each epoch]{\includegraphics[width=0.87\textwidth]{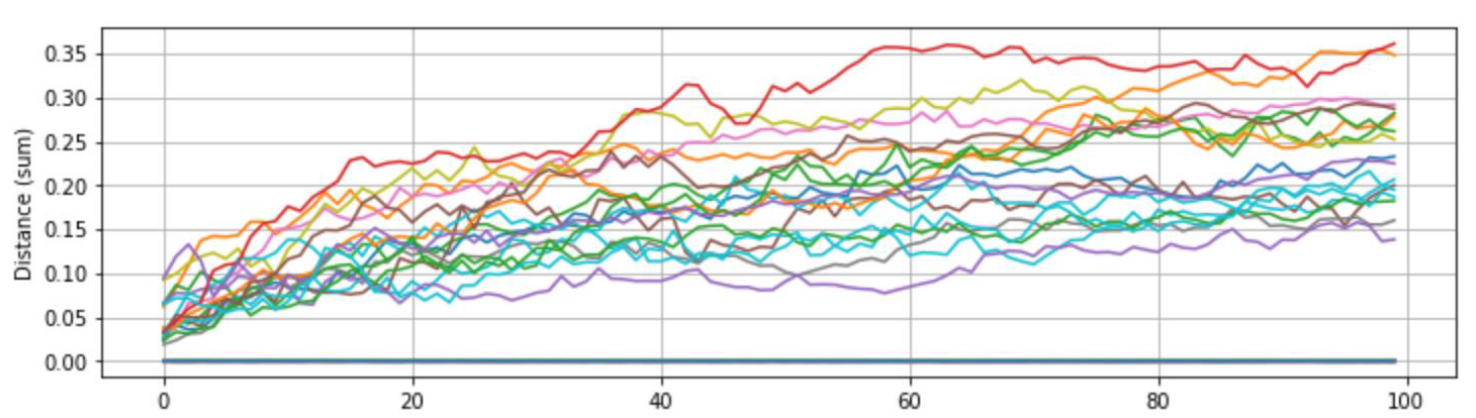}}
    \caption{Segment-wise distance history for a pair of clients in the "Different segment" setting with $E=1, PSS=0.5$.}
    \label{fig:cusegdist}
\end{figure}
Figure \ref{fig:cusegdist} offers a view of segment-wise distance history. Figure \ref{fig:cusegdist}(a) shows the distance over all aggregation iterations, where the distances between shared segments quickly drop to 0, while the others remain the same. Similarly, Figure \ref{fig:cusegdist}(b) and Figure \ref{fig:cusegdist}(c) shows the distances at the start and end of each epoch. The distances between shared segments increase a bit during individual training, but then dropped to 0. The distances between segments that are not shared would remain the same.

\subsubsection{Segment-wise Sharing Topologies}
Instead of relying on the original Small-world communication topology between clients, we propose aggregation with segment-wise topologies, as an attempt to guarantee consensus among aggregated segments. Two methods are developed and detailed below.

\begin{figure}[!htbp]
    \centering
    \subfigure[]{\includegraphics[width=0.47\textwidth]{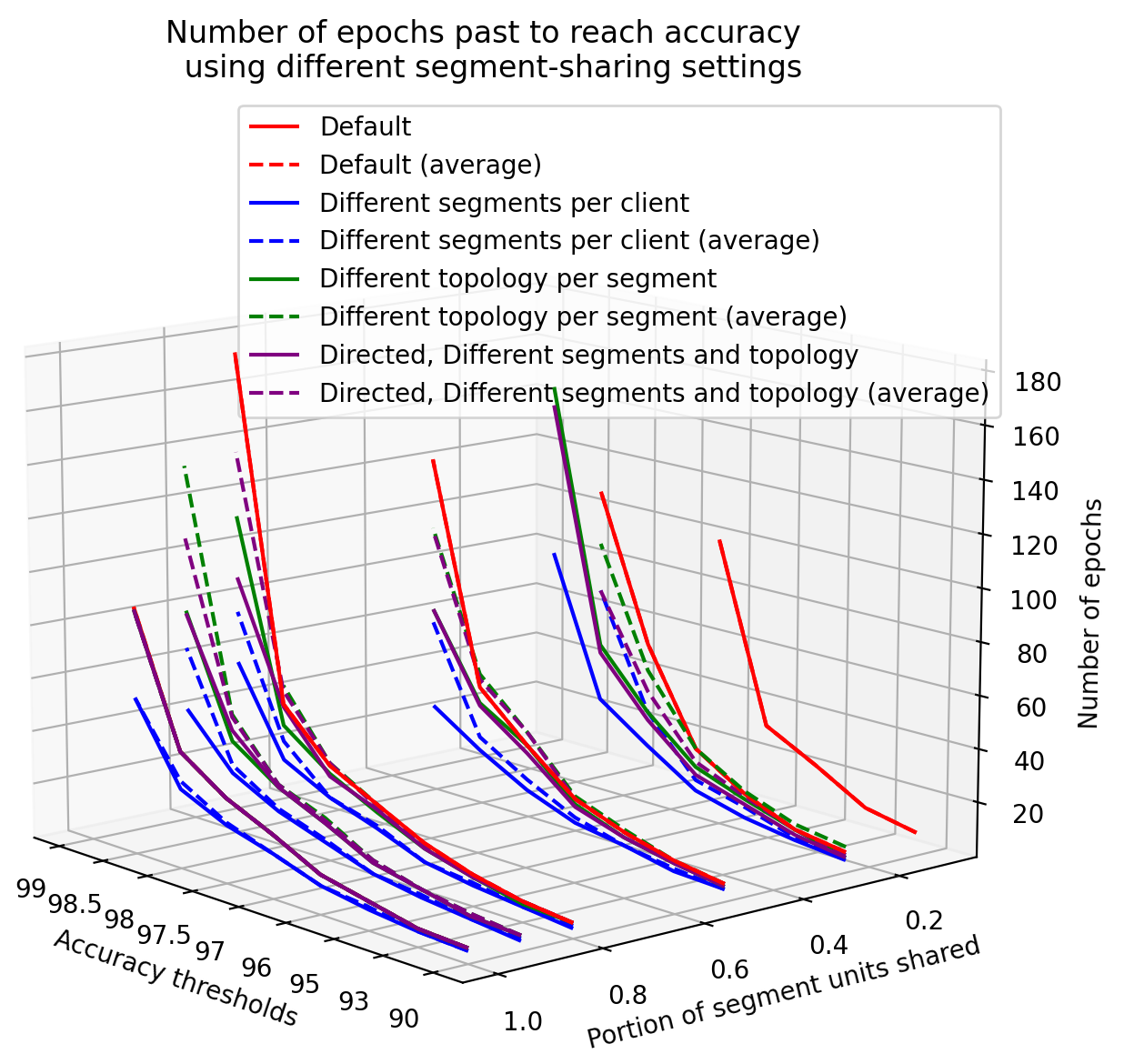}}
    \subfigure[another angle without averages]{\includegraphics[width=0.47\textwidth]{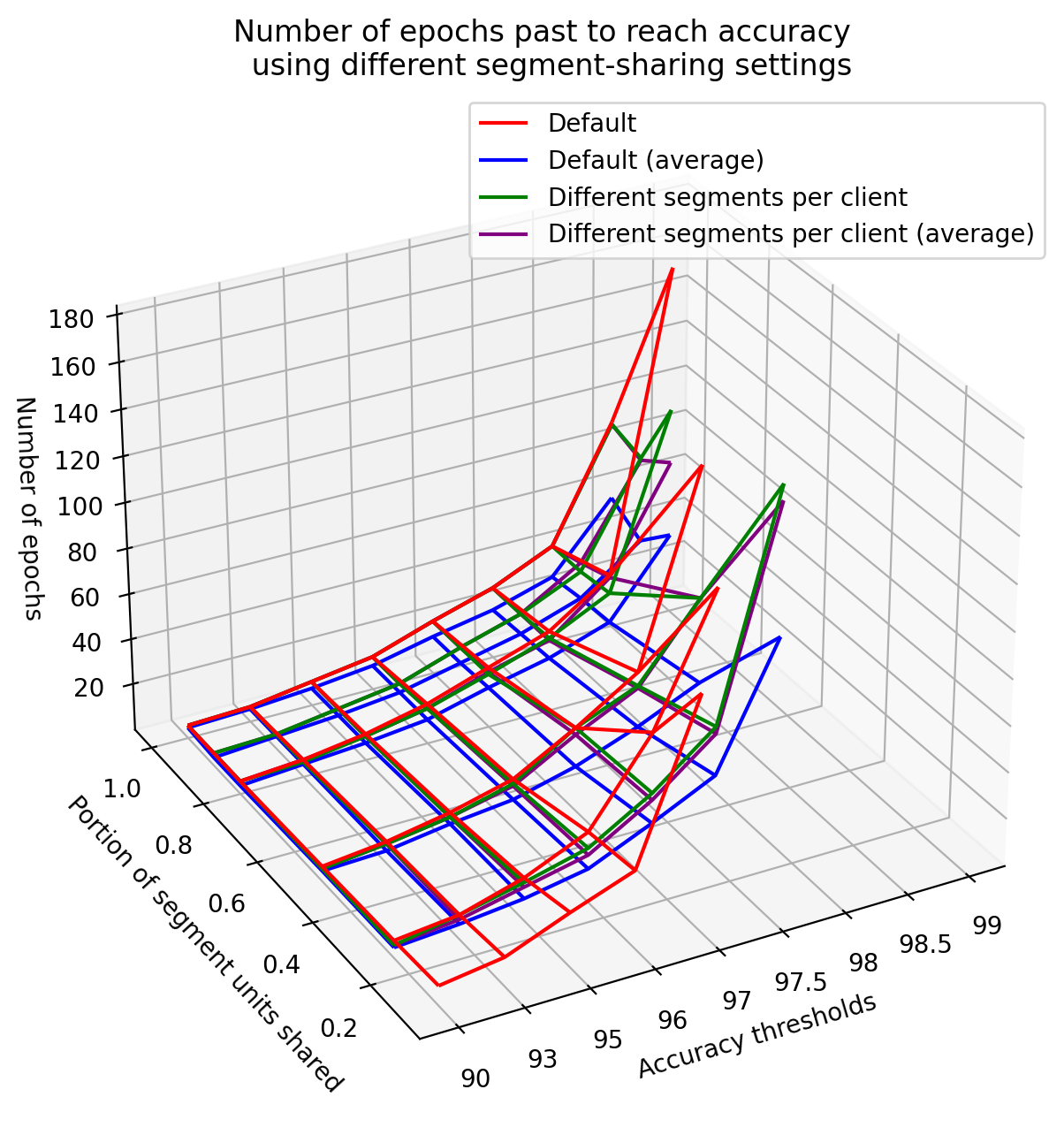}}
    \subfigure[max accuracy]{\includegraphics[width=0.47\textwidth]{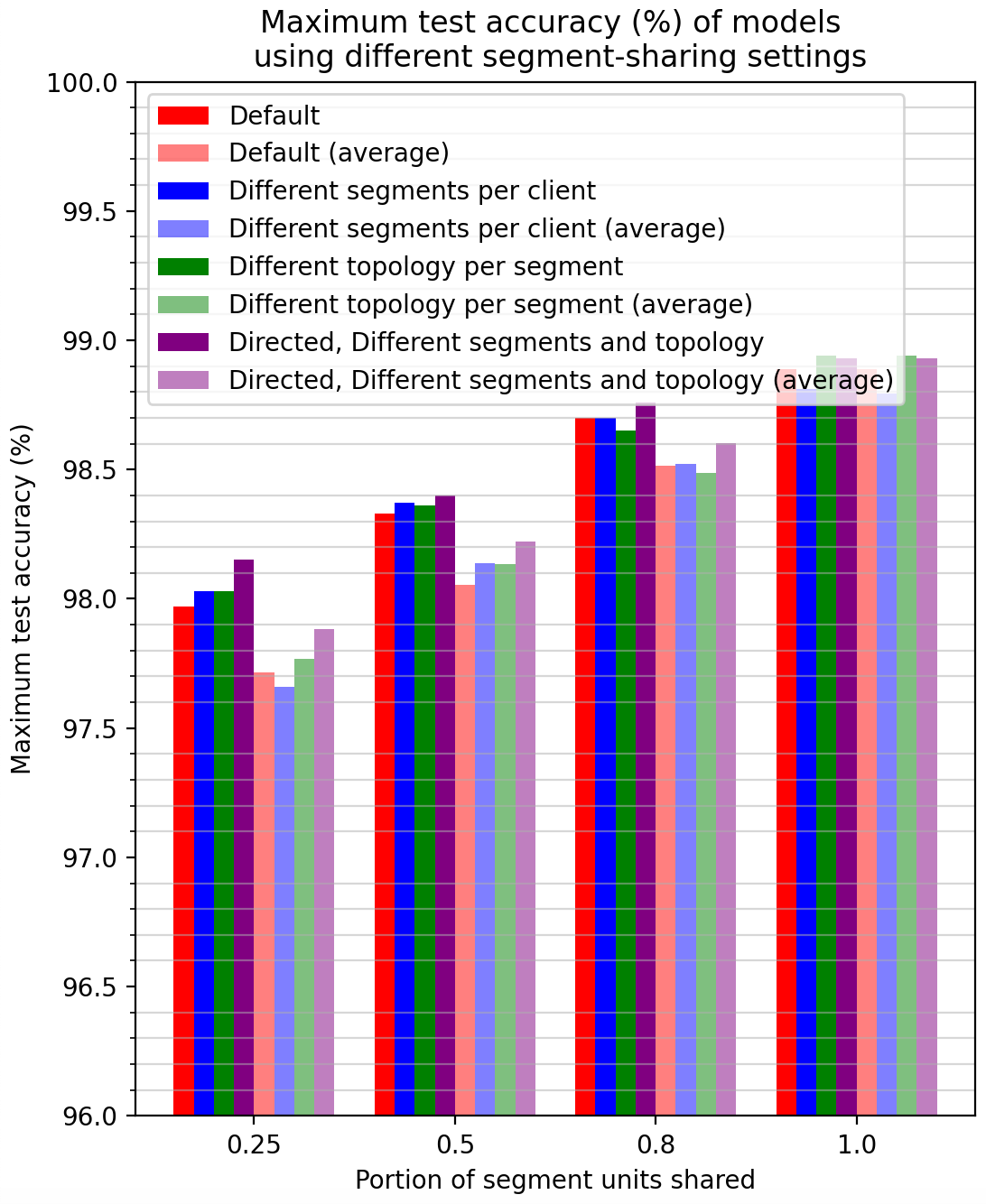}}
    \subfigure[model distance range; "Default" and "Different segments" ones used $E=0.1$ instead of $E=0.05$]{\includegraphics[width=0.37\textwidth]{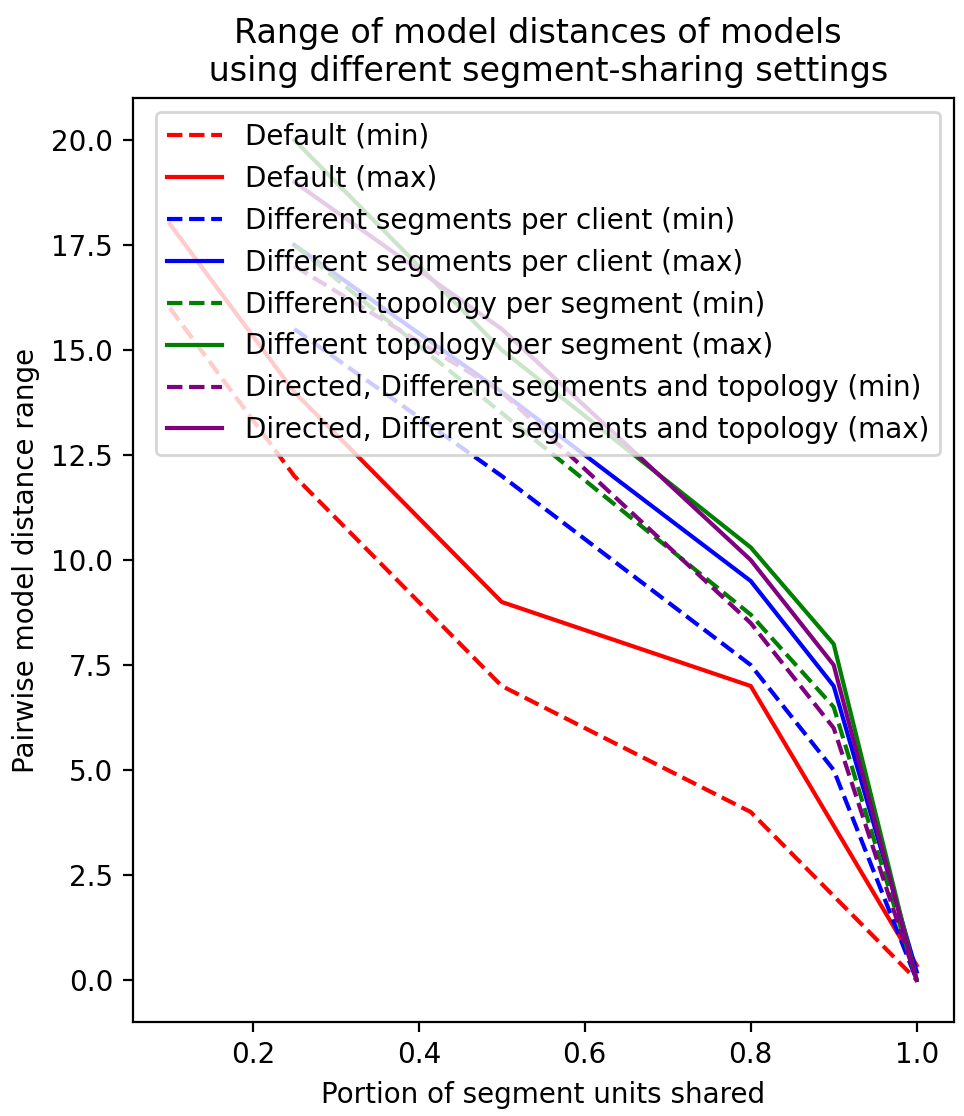}}
    \caption{(a-b) Number of epochs to reach $e_{90}-e_{98.5}$ for the top-behaving model and the model averages, (c) max accuracies, and (d) range for pairwise model distances in $\ell_2$ norms, compared between different sharing settings described in the legend.}
    \label{fig:cucross}
\end{figure}

\textbf{"Different topology per segment" setting:} Here, we randomly generate a small-world sharing topology graph $G_{i,c_i}$ for each segment $W_{L_i,out,c_i}$ with index $(i,c_i)$. 
Each client $j$ still keeps a set of segment index $SegInd^j$ that it shares with others at each aggregation iteration, and its size is still $|SegInd^j| = PSS\cdot\sum_{i=1}^l C_i$.
Similar to "Default" setting, we still guarantee $SegInd^{j_1} = SegInd^{j_2}$ for any $j_1\neq j_2$.
Figure \ref{fig:cucross} includes results that compare the previous two settings and this setting.\\
\textbf{"Directed, Different segments and topology" setting:} Here, each segment $W_{L_i,out,c_i}$ still follows a small-world connection topology $G_{i,c_i}^{(t)}$.
However, to simulate the stochastic factors in real application that could affect the communication, this small-world connection topology is randomly formed at each epoch $t$. 
This is to simulate an application where each client $j$ does not know which segment's message would arrive first during polling.
To deal with this uncertainty with less communication cost, it stops polling from other clients once it has received $PSS\cdot\sum_{i=1}^l C_i$ segments in total.
As a result, the effective communication graph for each segment at each time would be different.
For simplicity, we only change the topology $G^{(t)}_{i,c_i}$ at the start of each aggregation, and assume that all following iterations would use the same topology.
The following describes this procedure for client $j$:

\begin{algorithm}[H]
\caption{Polling algorithm for client $j$ with directed, different segments and topology}
\SetAlgoLined
 $polledSegmentCount \gets [0,...,0]$\;
 $segmentGraphLambda \gets [0,...,0]$\;
 \While{True}{
  \eIf{segment $(i,c_i)$ from client $k$ is received}{
  $polledSegmentCount[(i,c_i)] += 1$\;
  Update $segmentGraphLambda[(i,c_i)]$ from the received message\;
  }{}
  \eIf{$|\{(i,c_i) : \mathrm{polledSegmentCount}[(i,c_i)] > 0 $ and $ segmentGraphLambda[(i,c_i)] > 0\}| \geq PSS\cdot |W^j|$}{break\;}{}
 }
\end{algorithm}
We assume that each message regarding $W^j_{L_i,out,c_i}$ would also contain a list of existing edges for this segment, so that the client could estimate if the communication topology has already become a complete graph. 
The time it takes for each segment's message to reach client $j$ is fully stochastic, so the set of segments that are aggregated for $j$, and the set of clients whose segment weights are used during the aggregation for each segment is also fully random.
In simulation, however, we could simplify the procedure as follows:

\begin{algorithm}[H]
\caption{Simplified polling algorithm for client $j$}
\SetAlgoLined
\For{each epoch}{
 Randomly generate $SegInd^j$ with size $PSS\cdot|W^j|$\;
 \For{$(i,c_i)$ in $SegInd^j$}{
  Randomly generate a small-world topology $G_{i,c_i}$\;
  Poll the segment from all its neighbors, and perform aggregation\;
 }{}}
\end{algorithm}
Notice that, unlike in the other settings, this procedure does not have guarantee of an undirected (bidirectional) update graph for each segment. Thus, it would be hard to guarantee consensus convergence in this setting. This could be tackled with future research efforts.

In Figure \ref{fig:cucross}, we compare the performances of all 4 settings. 
In Figure \ref{fig:cucross}(a-b), the general trend is that "Different segments" setting has the fastest rate to reach certain accuracy thresholds; the settings with segment-wise topology graphs often has similar performances with each other. 
On the other hand, Figure \ref{fig:cucross}(c) reveals that all settings have very similar performances in terms of peak accuracy, and (d) shows that the model distance history for each setting is also similar.

According to Figure \ref{fig:cuno}(d) and Figure \ref{fig:cuw}(d), as $E$ increases, the model distance would increase as well. 
Thus, if Figure \ref{fig:cucross}(d) are showing the results with $E=0.05$ from all settings, then the distance ranges from settings with client-wise topologies would have been even smaller than the settings with segment-wise topologies. 
Thus, the segment-wise sharing settings achieve similar performances, proving that they are viable.
However, once again, not all segments are aggregated when $PSS < 1$.
This explains why the model distances are still nonzero.

\subsubsection{Practicality Analysis and Conclusions}
\begin{figure}[!htbp]
    \centering
    \includegraphics[width=0.87\textwidth]{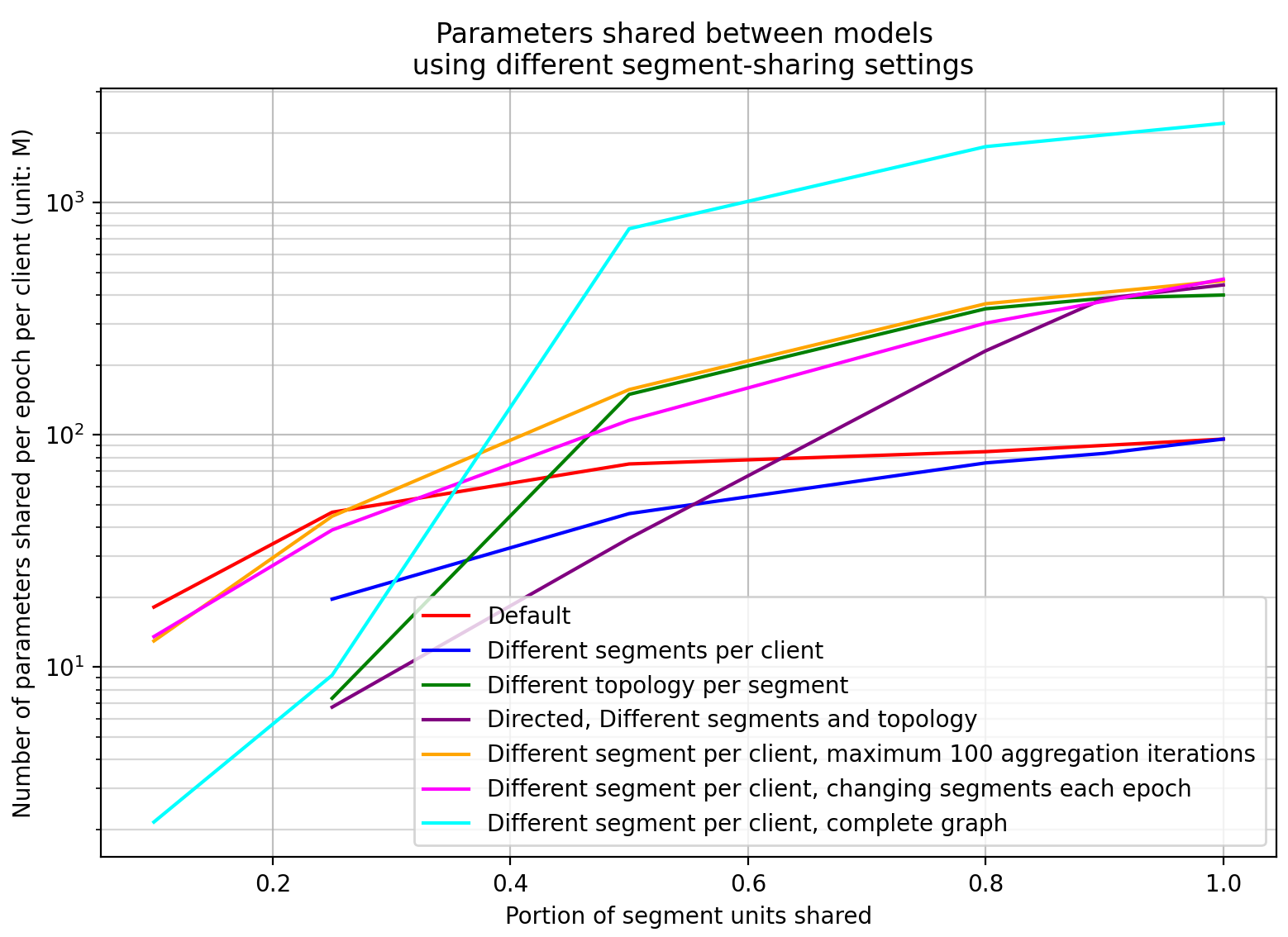}
    \caption{Average number of parameters shared by each client during each epoch's aggregation, compared between different sharing settings described in the legend. The last 3 labels had 100 as the maximum number of averaging iterations, while the others had 20.}
    \label{fig:cucommuncost}
\end{figure}

To evaluate the effect of the different segmented federated learning approaches mentioned above, 
we have documented the communication costs of the procedure, represented simply as the combined sizes of shared segments at each epoch for now. 
The results are shown in Figure \ref{fig:cucommuncost}, where the y-axis unit is in millions of floats. 
The figure compares the 4 settings defined in previous sections (with at most 20 aggregation iterations). In addition, it included the three test scenarios from Figure \ref{fig:cuPre} (with at most 100 aggregation iterations). 

As expected, larger $PSS$ values resulted in larger communication costs, but more factors come into play.
Between the first four settings, at each $PSS$ value, the amount of segments being shared are the same.
In addition, the number of edges in communication graphs should also be the same, because they followed the same Small-World topology generation procedure. 
Thus, the factor that caused different communication costs should be the amount of iterations required to reach consensus.

The communication cost difference between the complete graph setting and the rest in Figure \ref{fig:cucommuncost} demonstrates the trade-off between denser network and higher communication cost well.
The complete graph network requires relatively fewer iterations to converge than small-world networks, thanks to its high density.
However, as $PSS$ increases, the amount of messages passed within a complete graph grows faster than the amount for Small-World graphs.
As a result, the complete graph communication cost exceeds the other settings before $PSS=0.5$.
The additional cost due to higher density exceeded the savings by having a faster convergence rate.

The settings using fixed topology end up with 3 times less communication cost than the ones using segment-wise topology,
even when not considering the extra overhead generated when determining the communication topology for each segment. 
In real applications, the segment-wise small-world option would require more communication to verify the connectivity of the graph, so the communication costs would be even larger. 
In addition, from a programming perspective, the simulation runtime of segmented approaches is significantly longer than the default setting.
While this could be a result of inefficient implementation, we still conclude that segmented federated learning with channels as the basic unit is currently impractical.

\section{Conclusion and Future Works}\label{sec:conclusion}
In conclusion, our experiments show that DFL demonstrates robustness in training, even with different adjustments to the training scheme. In this section, we summarize the takeaways of each section, and list future directions.
\subsection{Different Dataset Settings}
We have observed that clients could share a small portion of training data with each other to stabilize training, especially if the training data is skewed between clients. 
In addition, models could still fit the training data well with varying normalization values and/or loss functions, provided that they are not too off from the true value.

Future work could verify if the same behavior can be observed from other benchmark datasets. In addition, one could apply the learning scheme to real-world distributed systems where clients have heterogeneous beliefs based on local observations, such as having different dataset statistics.

One may also consider testing the DFL robustness by combining different normalization parameters with skewed dataset partitions. 
In a preliminary experiment, we calculated the normalization parameters $\mu_i$ based on the partitioned dataset $D_i$ for each client $i$.
After training with Adam optimizer with learning rate at 0.001 for 50 epochs, the global model accuracy maxed out at 98\% with $S = 5\%$.
Meanwhile, the accuracy was at most 86\% when $S = 0$.
This indicates that DFL might be robust in this scenario, but more tests are needed.

\subsection{Different Optimization Methods}
In our experiments, we have explored different combinations of learning rates and optimizers.
While we have observed suboptimal performances from both scenarios, we have also seen that the DFL scheme is robust enough to arrive at working models with conservatively modified learning rates. 
In addition, we observe a slightly better performance when different optimizers are simultaneously applied to skewed datasets.

We observe that the skewed dataset experiment draws certain similarities with the notion of ensemble learning.
In ensemble learning, multiple models are trained with varying objectives, and the final prediction is determined through a weighted vote from all models \citep{emsemblelearning}. 
In our skewed dataset experiment, each client holds a model specialized with a certain training data distribution,
and models are trained with different optimizing methods.
However, instead of combining their outputs through a weighted sum, the DFL training scheme directly takes the (weighted) average of the model parameters. 
Both methods feature a distributed set of models, and a combination of predictions from all those models, albeit in different implementation methods.
Future work could study their connections and differences in more detail, as well as verify if the observations would still hold in other benchmark datasets and training tasks.

\subsection{ Segmented Federated Learning}
First and foremost, works in the future should check if the observations described above would still hold true for a task that is harder to train on. For example, using a simpler model with fewer parameters to learn MNIST classification, or using the same model in this paper for CIFAR-10 classification, would both prolong the learning process, and allow us to "zoom in" the graphs over epochs further. In addition, more complex benchmark tests would allow for less frequent model aggregations - as we have observed, less frequent aggregation would yield better and faster performances, but the current model cannot afford to do so if the accuracy reaches 99\% within the first epoch.

In addition, it is important to derive and analyze theoretical results of the methods above, if feasible. By finding theoretical bounds of the training procedure, or finding equivalencies between experiments listed above, the testing could become less complicated. 

The segmentation procedure requires optimizing the current implementation, as well. The smaller the segments, the more of them needs to be shared, and this results in larger overhead costs. Instead of reducing segment sizes, perhaps a better way is to reduce the resolution of the shared parameters, so that each client's shared parameters could be represented with much less bits - for example, \cite{fedpaq} proposed a variation algorithm, FedPAQ, that aimed at reducing communication cost by reducing the update frequency and quantizing the update process. 

Alternatively, the communication cost could be reduced by limiting the parameter structures. For example, if we approach the image classification task not with a CNN model, but with a graphical model such as a Markov Random Field, then the number of shared parameters could be reduced to the graph topology and the edge potentials. Past work regarding consensus with Bayesian Networks also exist in \cite{pena2011finding}. With the rise of interest in Graphical Neural Networks (GNN), there is much potential for this idea.

Possible future steps include sharing simplified models with similar performances, using ideas of transfer learning in decentralized ways, training Spars Convolutional Neural Networks (SCNN, developed by~\cite{liu2015sparse}) to reduce cost, etc. Some researchers have also discovered that models could grow further apart (measured by mutual information) while still increasing their parameters' correlations, thus showing us a different explanation to why the whole model distances are growing with each epoch~\citep{xiao2020averaging}. 

\subsection{Other topics}
If we want it to work on a directed graph, then we might have to use training schemes like push-sum, as mentioned by \cite{nedicpushsum,pushsumdl} to guarantee convergence. One work that combines Federated Learning with push sum scheme is \cite{pushsumfl}. 

One other idea is to have each client solve one particular classification problem (i.e. hold 10 clients, and each client $i$ is responsible for training a set of parameters $w_i$ that predicts the probability that the given image belongs to class $i$), and see if this still fulfills the decentralized optimization problem definition. One may also consider using the ADMM procedure, as outlined by~\cite{ADMM}, for federated learning~\citep{FGADMM}.

Another topic of interest is checking the effect of periodically applying random noises to some of the models, as a way to "nudge" models off from possible local minima. With multiple clients sharing their parameters in the federated learning framework, it could be possible that this model would encourage some of the clients to find a successful alternate direction. This idea could also be implemented by giving different momentum values for different clients, where clients with higher momentum values during optimization would tend to explore more, while clients with lower momentum would hold them back at each update step to prevent from possible divergence. 

Federated Learning could also apply in other settings. For example, consider Federated Reinforcement Learning~\citep{DRFL} - naturally, it would suit a multi-agent setup. Agents could communicate each other's findings through shared parameters, and converge to certain behavioral models. One other way to adapt this is to set it as a parallel training procedure, similar to ensemble learning, where multiple agents learn in the same environment with different parameters and hyper-parameters, and we try to locate one good model from them. 

\section{Acknowledgement}
This research was supported in part through research cyber-infrastructure resources and services provided by the Partnership for an Advanced Computing Environment (PACE) at the Georgia Institute of Technology, Atlanta, Georgia, USA.

\bibliographystyle{abbrvnat}
\bibliography{fl.bib}  

\end{document}